\documentclass{article}

\usepackage[preprint]{neurips_2026}

\usepackage[utf8]{inputenc} % allow utf-8 input
\usepackage[T1]{fontenc}    % use 8-bit T1 fonts
\usepackage{hyperref}       % hyperlinks
\usepackage{url}            % simple URL typesetting
\usepackage{booktabs}       % professional-quality tables
\usepackage{amsfonts}       % blackboard math symbols
\usepackage{nicefrac}       % compact symbols for 1/2, etc.
\usepackage{microtype}      % microtypography
\usepackage{xcolor}         % colors
\usepackage{amssymb,amsmath,amsthm}
\usepackage{cleveref}
\usepackage{multirow}
\usepackage{subcaption}
\usepackage{placeins}
\usepackage{longtable}

\newtheorem{ass}{Assumption}

\usepackage{algorithm}
\usepackage{algpseudocode}

\usepackage[most]{tcolorbox}
\newtcolorbox{promptbox}[1][]{
  breakable,
  colframe=black!40,         % Frame color
  colback=black!5,           % Background color
  coltitle=black,            % Color of the title text
  title=#1,                  % Optional title
  rounded corners,           % Corner style
  boxrule=0.5mm,             % Frame thickness
  boxsep=5pt,                % Space between content and box
  toptitle=1mm,              % Space above the title
  bottomtitle=1mm,           % Space below the title
  left=10pt,                 % Left padding
  right=10pt,                % Right padding
  top=5pt,                   % Top padding
  bottom=5pt,                % Bottom padding
  fonttitle=\bfseries        % Font style for the title
}

\usepackage{capt-of}

\title{Probing Perceptual Priors of MLLMs via Gibbs Sampling with Interpretable Generative Controls}

\AddToHook{cmd/appendix/before}{%
    \crefalias{section}{appendix}%
    \crefalias{subsection}{appendix}
}

\newcommand*\captiontype[1]{\def\@captype{#1}}

\author{Manuel Cherep$^1$, Pattie Maes$^1$, Nikhil Singh$^2$ \\
$^1$MIT, $^2$Dartmouth College}

\begin{document}

\maketitle

\begin{center}
\vspace{-2em}
\captiontype{figure}
\includegraphics[width=\linewidth,trim={0 10.5cm 0 0},clip]{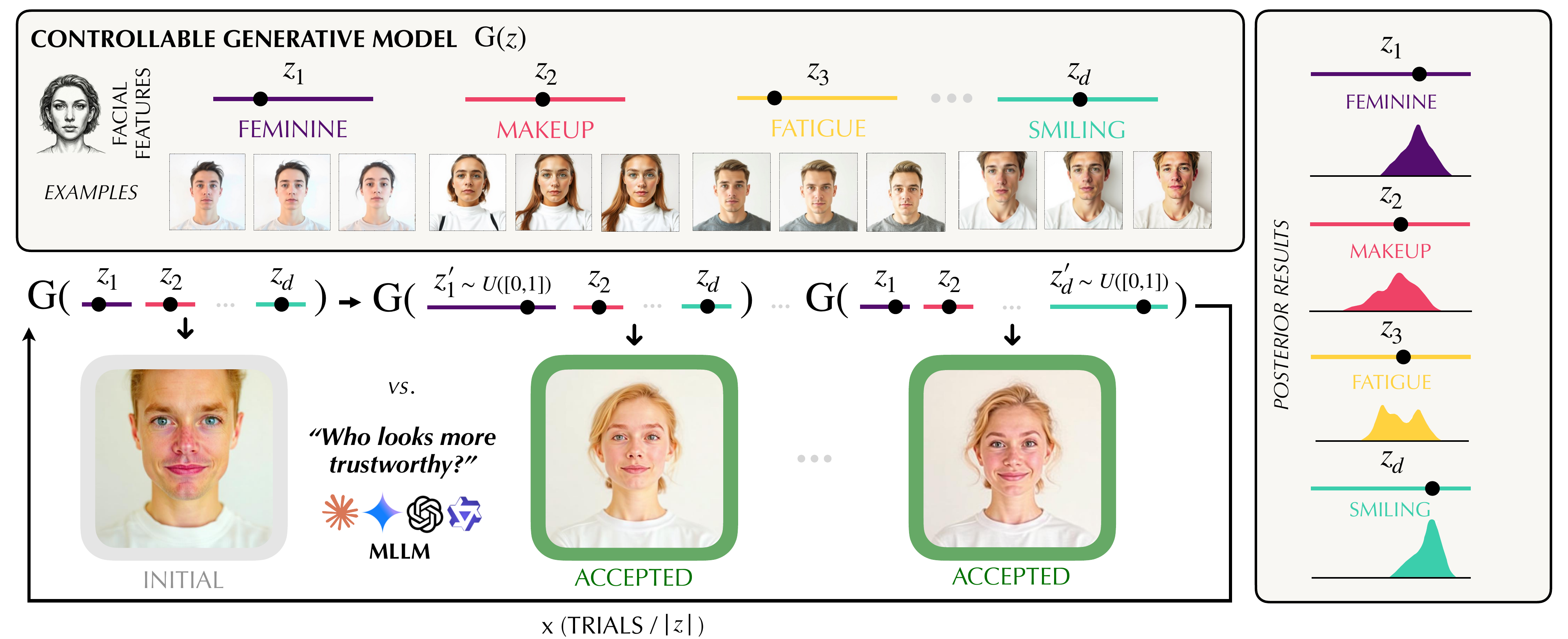}
\captionof{figure}{We probe the perceptual priors of multimodal language models by running a Metropolis--Hastings within Gibbs sampler over the interpretable latent space of a generative model. At each trial, the model is shown the current and a proposed stimulus (differing along one slider) and asked to choose the one that matches the target concept closely. Its binary choice realizes a Barker acceptance step, while iterating across slider dimensions yields Markov-chain samples from the model's implicit prior over the target. We apply the method across four domains (see examples in \Cref{app:examples}) where model priors shape how these models perceive us and our world.}
\label{fig:banner}
\end{center}

\begin{abstract}
A model's behavior on a task is jointly determined by the input it receives and the prior it brings in, i.e. the distribution over stimuli it implicitly expects. Interpretability research has traditionally studied models by holding inputs fixed and examining model responses either mechanistically, probing how internal structure represents inputs, or behaviorally, measuring how variation in inputs leads to variation in outputs. Neither reconstructs the prior distribution itself, since internal structure shows what a model can represent, not what it expects, and any fixed stimulus set leaves most of the possible input space unseen. In particular, such an input space in real-world settings, such as images seen by VLMs, is extremely high-dimensional and diverse. These priors thus remain a poorly understood component of models that nonetheless influence real-world behavior. We propose a method to sample from models' perceptual prior distributions directly, by steering a generative model to produce stimuli along controllable axes and running Gibbs sampling over that space with the model under study as the judge. We apply this to a variety of categories and target variables (such as trustworthiness in faces and cheapness in art images) and recover both canonical biases and surprising novel priors invisible to direct prompting, warranting further investigation of their downstream effects.
\end{abstract}

\section{Introduction}

When humans make decisions, we draw on implicit representations of the world. Our priors shape what we notice, what we prefer, and ultimately how we behave. Thus, behaving intelligently depends fundamentally on the quality of these representations. Models are no different, and their outputs are similarly shaped by representations learned from data. Yet while decades of work in cognitive science have developed techniques for characterizing human perceptual priors, we have comparatively few tools for probing the analogous representations of these models. This is particularly important for safety and alignment, as delegating to a model implicitly assumes its priors align with those of the humans it serves, an assumption that is rarely tested and, when it fails, can propagate biases invisibly through downstream systems.

The classical approach is to present participants with stimuli that span the space of interest and ask them to evaluate each one. In low-dimensional spaces, this works well. A small number of stimuli can densely cover the space, and the resulting evaluations directly map out the structure of the representation. But the number of stimuli needed grows combinatorially with the number of controllable dimensions, and any fixed set is implicitly a bet that the regions of interest lie within it. In high dimensions, this bet rarely pays off. Markov Chain Monte Carlo with People (MCMCP) \citep{sanborn2007markov,sanborn2010uncovering,sanborn2015exploring} and its extension Gibbs Sampling with People (GSP) \citep{harrison2020gibbs} address this problem by letting people's own judgments determine which stimuli are presented next. In MCMCP, participants make a binary choice at each step, and these choices act as the acceptance step of a Metropolis-Hastings sampler over the parameter space. GSP replaces the binary choice with a continuous slider and updates one dimension at a time, effectively implementing a Gibbs sampler that converges faster in high-dimensional spaces. Both methods were developed to extract human semantic representations, but the same logic can apply to models.

In this paper, we apply Gibbs sampling to study models rather than people. This requires a generative model with controllable, interpretable dimensions, allowing each one to be sampled in turn. Although the method is modality-agnostic, we validate it in the image domain and train sliders via SliderSpace \citep{gandikota2025sliderspace}, which provides LoRA-based sliders for the text-to-image model FLUX \citep{labs2025flux}. We apply the method across four domains: faces, affordances, aesthetics, and authenticity. Each one probes priors that shape consequential decisions. Faces capture how models see us, the lens through which they increasingly screen applicants, defendants, or partners. Affordances test whether models share the physical intuitions required for robotics and embodied reasoning. Aesthetics expose the priors that govern what generative and recommendation systems produce. Finally, authenticity reveals the priors that determine what models accept as real or altered, with implications for misinformation and content moderation. Overall, this work contributes:

\begin{enumerate}
    \item A Bayesian method for sampling the perceptual priors of multimodal language models over interpretable generative latent spaces. Unlike prior MCMC work that operates over simple words or closed-form numerical inputs \citep{zhu2024recovering,zhu2024eliciting,marjieh2024large}, our setup is validated with images as inputs over a complex text-to-image generator (FLUX + SliderSpace).
    \item An open-source modality-agnostic framework: any controllable, interpretable generator paired with a binary preference judge plugs into the same scheme, so the method extends naturally to other modalities.
    \item Experiments across four high-stakes image domains (faces, affordances, aesthetics, and authenticity), using four frontier VLMs (\textit{Claude Sonnet 4.6}, \textit{Gemini 3 Flash}, \textit{GPT-5.4}, \textit{Qwen3-VL 235B}), recovering priors as full posteriors that direct prompting fails to surface.
\end{enumerate}

\section{Related Work}

\subsection{Eliciting Human Perceptual Priors}
Cognitive scientists have a long tradition of developing methods for recovering human perceptual priors. MCMCP \citep{sanborn2007markov,sanborn2010uncovering,sanborn2015exploring} elicits these priors through adaptive Metropolis-Hastings sampling \citep{metropolis1953equation,hastings1970monte}. It casts the participant's binary choice as Barker's acceptance rule \citep{barker1965monte}, equivalent to a Bradley-Terry model of paired comparison \citep{bradley1952rank,christiano2017deep}, valid under the assumption that humans probability-match their choices to the underlying preference \citep{vulkan2000economist,shanks2002re}. GSP \citep{harrison2020gibbs} extends this with continuous slider responses and dimension-wise updates, effectively implementing a Gibbs sampler \citep{geman1984stochastic} that converges faster in high-dimensional spaces. GSP has since been applied across diverse perceptual domains, including musical consonance \citep{marjieh2022reshaping}, singing \citep{anglada2023large}, emotional prosody \citep{van2021exploring,van2022voiceme}, visual patterns \citep{kumar2022using}, musical chords \citep{marjieh2024timbral}, and multidimensional visual aesthetics \citep{van2024using}. We adapt this family of methods to multimodal language models (MLLMs) rather than people, taking the Gibbs structure from GSP and the Barker-rule binary acceptance \citep{barker1965monte} from MCMCP, leveraging categorical rather than continuous responses. The combination is a Metropolis-Hastings within Gibbs sampler \citep{tierney1994markov}, with the MLLM playing the role of the acceptance function. By updating one dimension at a time, our method's efficiency is closer to GSP than MCMCP in high-dimensional spaces.

\subsection{Probing Models}

Existing model evaluations are largely functional, asking whether the model produced the right output. But behavioral systems, such as agents powered by these models, can have different decision processes that produce identical outputs. Therefore, behavioral tests are needed to surface why a model behaves the way it does, what conditions change its behavior, and how it differs from humans \citep{cherep2025behavioral,cherep2025llm,cherep2024superficial}. Such tests require new tools, methods, and frameworks across a range of domains. From counterfactual environments that enable the causal exploration of agent behavior \citep{cherep2025framework}, to multi-agent environments \citep{vezhnevets2023generative}, to visual prompt optimization to study what visual features drive agentic decisions \citep{cherep2026visual}. Together, these efforts probe agents at multiple levels both internal and external. Our work adds another tool, recovering the perceptual priors that shape how models behave.

A growing body of work probes models in Bayesian terms. \citet{zhu2024recovering} use MCMC with LLMs to recover mental representations over text. \citet{zhu2024eliciting} elicit LLM priors over abstract numerical and causal quantities through iterated in-context learning. \citet{marjieh2024large} compare LLM judgments to human sensory judgments across six modalities. Unlike these, our work probes priors directly within a controllable and multimodal generative space rather than over text. Because the same Gibbs protocol applies to both humans and models, the recovered posteriors are directly commensurable in the same latent space.

\subsection{Controllable Generative Models}
Our method requires a generative model with independently controllable dimensions. The original GSP experiments used StyleGAN \citep{karras2019style,karras2020analyzing} controlled via GANSpace \citep{harkonen2020ganspace}, which discovers directions through PCA in the latent space. These directions are sometimes recognizable, but they are not designed to be interpretable, and many remain entangled. Interpretability matters for our purposes because it is what makes the recovered priors legible. Without interpretable axes, we cannot link a posterior over the latent space to meaningful concepts. Instead, we use SliderSpace \citep{gandikota2025sliderspace}, which extends Concept Sliders \citep{gandikota2024concept} (LoRA-based controls for diffusion models) by automatically decomposing a model's visual capabilities into named, interpretable axes. SliderSpace also can operate over FLUX \citep{labs2025flux}, which produces substantially more realistic and diverse images than StyleGAN. Other approaches to disentangled control are compatible with our method, including dedicated semantic latent spaces in diffusion models \citep{kwon2022diffusion,haas2024discovering} and sparse autoencoders \citep{surkov2025unpacking,singh2025discovering}.

\section{Methods}
We now turn to the mechanics of running Gibbs sampling with a multimodal LM as the judge. The natural design is to give the model a slider and ask it to pick a position, but this turns out to be impractical. The cost of every small slider move is high in this regime, and the perception-action loop is turn-based for VLMs which means they can expend an unbounded number of rounds making small moves in slider space by trial and error. We also observed VLMs' tendency to oscillate, explore the bounds, and take arbitrarily small steps back and forth. To mitigate this, we propose to use Metropolis-Hastings within Gibbs with the Barker acceptance function as a tractable replacement.

\paragraph{Setup.}
Let $G : \mathcal{Z} \to \mathcal{X}$ be a controllable text-to-image generator, where $\mathcal{Z} = \prod_{k=1}^d \mathcal{Z}_k$ is a product of $d$ interpretable, mutually orthogonal slider axes (for example, as in our case, obtained via training SliderSpace~\citep{gandikota2025sliderspace} on top of FLUX.1-schnell~\citep{labs2025flux}). Each $\mathcal{Z}_k$ we first reparameterize to $[0,1]$. We use SliderSpace to learn $n = 64$ candidate axes per domain; the human interpreter then selects a subset of $d \le n$ axes whose effects are semantically interpretable and potentially useful, which preserves the orthogonality property of the parent set.

Throughout, we fix a VLM judge and a textual criterion $C$ (e.g. ``more attractive''). The judge receives two stimuli and returns a binary choice. We assume choices are governed by an implicit, unobserved utility $\ell : \mathcal{Z} \to \mathbb{R}$ via the Bradley-Terry model~\citep{bradley1952rank}:
\begin{equation}
\Pr[\textrm{judge picks } z' \mid z, z'] = \sigma\big(\ell(z') - \ell(z)\big), \quad \textrm{where}\ \sigma(u) = (1+e^{-u})^{-1}.
\label{eq:bt}
\end{equation}

This is the same primitive that underlies much of the RLAIF literature~\citep{bai2022constitutional,lee2023rlaif}. The induced target distribution is then:
\begin{equation}
\pi(z) \propto \exp\big(\ell(z)\big),
\label{eq:target}
\end{equation}

which is the GSP target distribution~\citep{harrison2020gibbs} when the VLM's noise scale is absorbed into $\ell$. Modes of $\pi$ are accordingly the ``prototypes'' the VLM endorses for $C$, and the spread of $\pi$ describes the variation it tolerates. Here, two methods from the GSP/MCMCP family are most directly useful: \textit{joint MCMCP} (binary comparison with full-state proposals) and \textit{direct GSP} (asking the agent to pick a slider position). Both have been adapted to language models on low-dimensional domains~\citep{zhu2024recovering}, but (1) our combinatorial dimensionality is much higher here and queries are much costlier, and (2) our generator is much more complex than the number--to--color mapping, and thus closed-form numbers from LLMs would be uninterpretable as generation parameters.

As we noted earlier, the natural design (giving the VLM a slider and asking it to pick a position) is implausible in this regime. \citet{zhu2024recovering}~avoid such issues in a 3-D HSL space by asking GPT-4 to emit a value for the missing dimension, which is effectively a closed-form numerical answer to a closed-form numerical question. Our setting has no such closed form, since the slider is over generated images, and the ``answer'' is thus a position whose meaning the VLM only reliably knows \textit{after} another generation. This complicates the purported GSP information advantage in our setting.

A natural alternative is joint MCMCP, which proposes the entire $z$ at once and still uses a single binary VLM comparison per step. The single-query primitive is intuitively appealing, but the joint proposal is not scalable. \citet{roberts1997weak}~show that for random-walk Metropolis on a $d$-dimensional target, the proposal scale must shrink as $\sigma_d \propto d^{-1/2}$ to maintain a non-degenerate acceptance rate, after which the convergence time grows proportionally. This becomes a high-query design, which is ruled out by the high cost of stimulus generation. Our scan must therefore amortize stimulus cost over as much information as possible.

\subsection{MH-within-Gibbs with Barker acceptance}
\label{sec:algorithm}
We propose to instead use a Metropolis-Hastings within Gibbs sampler~\citep{tierney1994markov} that uses the binary VLM comparison as its acceptance step (see \Cref{alg:mh_gibbs_barker}). This construction has three pieces:

\begin{enumerate}
	\item We update one coordinate at a time, conditional on the others. This is the structural device of GSP~\citep{harrison2020gibbs}.
	\item Within each coordinate update, we use the Barker acceptance rule~\citep{barker1965monte}:
		\begin{equation}
		\alpha_B(z, z') = \frac{\pi(z')}{\pi(z) + \pi(z')}
		\label{eq:barker}
		\end{equation}
		Barker is the acceptance rule used by MCMCP~\citep{sanborn2007markov}. Substituting \eqref{eq:target} into \eqref{eq:barker} gives
		\begin{equation}
		\alpha_B(z, z') = \dfrac{e^{\ell(z')}}{e^{\ell(z)} + e^{\ell(z')}} = \sigma\big(\ell(z') - \ell(z)\big)
		\label{eq:identity}
		\end{equation}
		which is exactly the BT comparison probability (\Cref{eq:bt}). Every binary VLM query thus realizes a single Barker accept/reject step at no additional cost, the same property exploited by~\citep{sanborn2007markov,zhu2024recovering} in the joint-proposal regime.
	\item Given current state $z = (z_1, \ldots, z_d) \in \mathcal{Z}$, one full scan iterates $k = 1, \ldots, d$:
		\begin{enumerate}
			\item Propose $z_k' \sim q_k(\cdot \mid z_k)$ with $q_k$ a symmetric kernel on $\mathcal{Z}_k$; let $z' = (z_1, \ldots, z_k', \ldots, z_d)$
			\item Generate $G(z)$ and $G(z')$. Present them to the VLM with criterion $C$ in randomized order; receive choice $Y \in \{z, z'\}$
			\item Set $z \leftarrow Y$
		\end{enumerate}
		We use $q_k \sim \mathrm{Uniform}([0,1])$\footnote{This type of construction is variously known as independence chains/sampling~\citep{tierney1994markov,roberts1998markov}, Metropolized independent sampling~\citep{liu1996metropolized}, or independent Metropolis--Hastings (IMH)~\citep{holden2009adaptive}. These denote proposals independent of the current state. We use the latter (IMH), and study the special case $q_k \sim U(0, 1)$.}, drawing each proposal independently of the current state (in \Cref{app:random_walk_mh} we use a random walk to generate each proposal with the Robbins--Monro rule \citep{robbins1951stochastic}). This is Independent Metropolis-Hastings (IMH) within Gibbs. One scan across dimensions thus costs $2d$ image generations and $d$ VLM queries. The Gibbs decomposition then reuses one of the two generations from the previous step in each scan, reducing the amortized cost to $d+1$ generations per scan.
\end{enumerate}

\begin{algorithm}[h]
\caption{Metropolis--Hastings within Gibbs with Barker acceptance}
\begin{algorithmic}[1]
\Require Initial slider state $z^{(0)} \in [0,1]^d$, criterion $C$, initial proposal scale $\sigma_0$, target acceptance rate $\alpha^*$, decay $\eta$, burn-in fraction $b$, number of trials $T$.
\State $\sigma \gets \sigma_0$
\For{$\tau = 1, \ldots, T$}
    \State $k_\tau \gets ((\tau-1) \bmod d) + 1$ \Comment{Gibbs sweep over coordinates}
    \If{Proposal is uniform}
        \State Sample $z^* \sim \mathcal{U}(0, 1)$
    \ElsIf{Proposal is Gaussian}
        \State Sample $z^* \sim \mathcal{N}(z_{k_\tau}^{(\tau-1)}, \sigma^2)$ reflected at $[0,1]$
    \EndIf
    \State Set proposed state $z' \gets (z_1^{(\tau-1)}, \ldots, z^*, \ldots, z_d^{(\tau-1)})$, replacing coordinate $k_\tau$
    \State Generate current stimulus $G(z^{(\tau-1)})$
    \State Generate proposed stimulus $G(z')$
    \State Randomize image order \& query VLM under criterion $C$ \Comment{Barker rule from VLM choice}
    \If{the VLM chooses $G(z')$}
        \State $z^{(\tau)} \gets z'$
        \State $a \gets 1$
    \Else
        \State $z^{(\tau)} \gets z^{(\tau-1)}$
        \State $a \gets 0$
    \EndIf
    \If{Proposal is Gaussian and $\tau \leq bT$}
        \State $\log \sigma \gets \log \sigma + \tau^{-\eta}(a - \alpha^*)$ \Comment{Robbins--Monro rule \citep{robbins1951stochastic}}
    \EndIf
\EndFor
\label{alg:mh_gibbs_barker}
\end{algorithmic}
\end{algorithm}

Importantly, we observe that the BT assumption (\Cref{eq:bt}) need only hold along single SliderSpace axes for our chain to be valid:

\begin{ass}[Coordinate-wise Bradley-Terry]
\label{ass:cwbt}
For every coordinate $k \in \{1, \ldots, d\}$, every $z_{-k} \in \prod_{j\ne k} \mathcal{Z}_j$, and every pair $z_k, z_k' \in \mathcal{Z}_k$:
	\begin{equation*}
		\Pr[\text{VLM picks } (z_k', z_{-k}) \mid (z_k, z_{-k}), (z_k', z_{-k})] \;=\; \sigma\!\big(\ell(z_k', z_{-k}) - \ell(z_k, z_{-k})\big)
	\end{equation*}
\end{ass}

Assumption~\ref{ass:cwbt} is weaker than the full BT assumption, since it only requires BT to hold for comparisons that vary on a single axis. In our regime, Assumption~\ref{ass:cwbt} is more plausible than full BT for a few reasons. First, SliderSpace explicitly seeks ``semantic orthogonality'' among the principal directions~\citep{gandikota2025sliderspace} (axes are initialized as principal components of the CLIP-embedding distribution induced by the prompt, and the LoRA adapter for each axis is trained to align $\Delta\phi_k = \phi(G(z+\delta e_k)) - \phi(G(z))$ with the corresponding principal direction $v_k$). Single-axis perturbations therefore produce perceptual changes confined to one orthogonal CLIP direction by construction. Second, SliderSpace also enforces that each axis induces consistent transformations across seeds and prompt variations~\citep{gandikota2025sliderspace}. The variation along $z_k$ at a fixed $z_{-k}$ thus has stable semantic meaning, which is what allows $\ell(\cdot, z_{-k})$ to be a coherent function of $z_k$. Finally, if we human-select a subset of $d$ axes from the $n$ that SliderSpace produces, the orthogonality and consistency properties survive regardless. Orthogonality of $\{v_1,\ldots,v_n\}$ in CLIP space implies orthogonality of any subset, and per-axis consistency is independent of the other axes retained. The complementary assumption we make is that SliderSpace orthogonality maps to a regime where VLM judges are reliable. Empirically, prior work has found that LLM/VLM judgements can degrade in multi-attribute settings~\citep{stureborg2024large}. Though not directly comparable to our setting, it further supports the value of reducing VLM responses to coordinate-wise comparisons. We therefore expect Assumption~\ref{ass:cwbt} to hold to a closer approximation than its full-BT counterpart would.

\subsection{Controllable Generative Models}
\label{sec:sliderspace}
Our sampler requires a generator with two properties: (1) each controllable dimension corresponds to a single, semantically interpretable feature; and (2) different dimensions are at least approximately orthogonal, so the coordinate-wise updates licensed by \Cref{ass:cwbt} do not couple unrelated factors.

We instantiate this with FLUX.1-schnell~\citep{labs2025flux} as the generator and SliderSpace~\citep{gandikota2025sliderspace} to train the orthogonal interpretable controls. We train a rank-$1$ LoRA for $n=64$ candidate sliders per domain (faces, aesthetics, authenticity, and affordances) with 50,000 PCA samples over the CLIP-embedding distribution induced by domain-specific diverse prompts; compute requirements are reported in \Cref{app:compute}. Effectively, perturbing the slider value translates the generated image's CLIP embedding along the corresponding principal axis.

Because the $n$ directions are PCA-orthogonal in CLIP space by construction, any subset is also orthogonal. We manually select $d=10$ sliders per domain whose effects are recognizable and useful for the target probing question, discarding the rest (see \Cref{app:sliders} for examples). This also keeps the per-scan cost tractable. Slider training details are in \Cref{app:sliderspace_training}, and validation that each selected axis behaves as a usable Gibbs coordinate, and that joint configurations actually steer the generator across distinct image distributions, is in \Cref{app:sliderspace_validation}. We automatically label the sliders using an auto-interpretability procedure described in \Cref{app:autointerp}.

\subsection{Probing Experiments}
\label{sec:exps}
We probe the priors of four frontier multimodal language models: \textit{Claude Sonnet 4.6}, \textit{Gemini 3 Flash}, \textit{GPT-5.4}, and \textit{Qwen3-VL 235B}. All models are queried with default temperature and sampling parameters. For each binary comparison, the order in which the current and proposed images are presented to the judge is independently randomized per query, so that the judge's preference is invariant to position effects \citep{pezeshkpour2024large}. For each (model, target) configuration, we run 10 independent chains of 2,000 trials each, with $d = 10$ sliders per domain. Seeds differ per chain and per target, so no two chains or targets share initial images.

We probe four domains, each with its own slider set and target list: \textbf{Faces} (8 targets): \textit{attractive}, \textit{criminal}, \textit{fun}, \textit{hardworking}, \textit{intelligent}, \textit{serious}, \textit{trustworthy}, \textit{youthful}; \textbf{Aesthetics} (5 targets): \textit{amateur}, \textit{beautiful}, \textit{cheap}, \textit{expensive}, \textit{experimental}; \textbf{Authenticity} (2 targets): \textit{authentic}, \textit{manipulated}; \textbf{Affordances} (8 targets): \textit{sittable}, \textit{carryable}, \textit{throwable}, \textit{breakable}, \textit{bounceable}, \textit{resonant}, \textit{rough}, \textit{squeezable}. This gives 23 targets total. The full task and prompts per domain are listed in \Cref{app:prompts}, and compute requirements are in \Cref{app:compute}. Before running the main experiments, we sanity-check the kernel on a controlled low-dimensional task with clear ground truth (HSL colors, \Cref{app:colors}).

\section{Results}
Across all domains and VLMs, our experiments span 2.5M+ trials and 1B+ tokens. Qualitative examples of the recovered priors across the four domains are shown in \Cref{app:examples}. Unless otherwise stated, posterior summaries are computed over post-burn-in samples from the IMH-within-Gibbs chains. Because the sampled quantities are slider coordinates in a controllable generative model, the reported values should be interpreted as model-specific association signals within the reachable SliderSpace--FLUX stimulus manifold (i.e. not as direct measurements of real-world attributes implied by the labels). For reference, \Cref{app:additional_analysis} shows the top-3 and bottom-3 sliders per target across the four models, and \Cref{app:posterior_summaries} shows posterior summaries for all (target, parameter, model) conditions in the experiment.

\subsection{Strongest and Weakest Recovered Priors}

\begin{figure}[!htb]
    \centering
    \includegraphics[width=\linewidth]{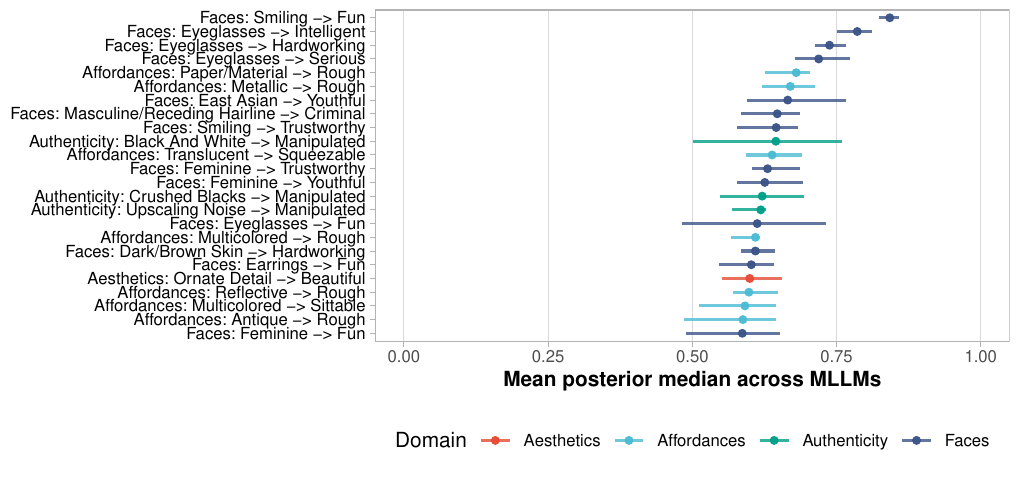}
    \caption{Strongest recovered priors across the $23$ targets, pooled across the four MLLM judges. Each row is a (target, slider) condition ordered by mean across models; the point is the cross-model mean and the bar shows the cross-model interquartile range. Slider values are in $[0,1]$; a value of $0.5$ corresponds to no preference along that axis.}
    \label{fig:headline_priors}
\end{figure}

\Cref{fig:headline_priors} reports the strongest recovered priors across the $23$ targets, ordered by cross-model mean. For face traits with cognitive-positive valence (\textit{Intelligent}, \textit{Hardworking}, \textit{Serious}) the top-recovered slider is the same: \textit{Eyeglasses}. For face traits with affective-positive valence (\textit{Fun}, \textit{Trustworthy}) the top-recovered slider is \textit{Smiling}. In the authenticity domain, models converge on visual edit-signatures (\textit{Black and White}, \textit{Crushed Blacks}, \textit{Upscaling Noise}) as markers of \textit{manipulated} images. For affordances, the top-recovered slider for \textit{Rough} is \textit{Paper/Material}. For art/aesthetics, the top slider is \textit{Ornate Detail} for \textit{Beautiful}.

\begin{figure}[!htb]
    \centering
    \includegraphics[width=\linewidth]{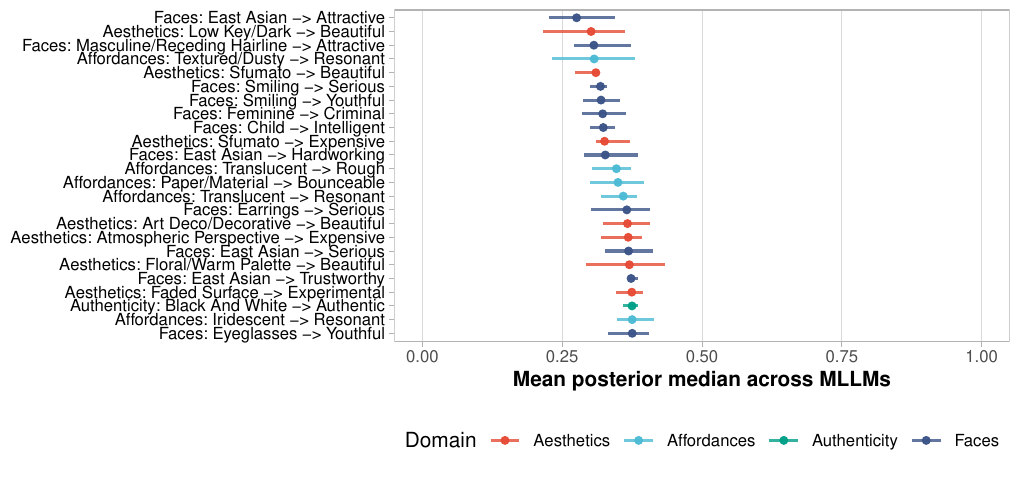}
    \caption{Weakest recovered priors across the $23$ targets, pooled across the four MLLM judges. Each row is a (target, slider) condition ordered by mean across models; the point is the cross-model mean and the bar shows the cross-model interquartile range. Slider values are in $[0,1]$; a value of $0.5$ corresponds to no preference along that axis.}
    \label{fig:headline_bottom_priors}
\end{figure}

\Cref{fig:headline_bottom_priors} parallels the previous plot in the opposite direction, surfacing the most negative (below the 0.5 baseline) associations models appear to encode. For the non-identity targets, several recovered priors place below baseline: \textit{Smiling} for \textit{Serious} and \textit{Youthful}, \textit{Child} for \textit{Intelligent}, \textit{Sfumato} for \textit{Beautiful} and \textit{Expensive}, \textit{Low Key/Dark} for \textit{Beautiful}, and \textit{Textured/Dusty} for \textit{Resonant}. We next look at identity-related priors.

\subsection{Demographic Associations}
\label{sec:results_demographic}
A central purpose of this proposed method is to make visible associations that direct prompting may not, so that they can be audited. As such models are increasingly used to screen, rank, or moderate content involving people, a directional preference linking a target concept to a demographic-linked visual axis is consequential even when a model might not state it when asked, as has been shown in prior work on covert biases~\citep{hofmann2024ai}. Among the recovered priors, several such conditions appear. We report them as candidate implicit biases the method exposes, deserving of further scrutiny. However, we note that the chain only recovers which direction along an axis a judge prefers for a given criterion. Whether that preference originates in pre-training, post-training alignment, or interaction with the generator, and whether it propagates to downstream decisions, is outside the scope of what these experiments can establish (\Cref{sec:limitations}). All values reported are medians over the full slider posteriors, where $0.5$ is the no-preference point.

Concerningly, for \textit{Attractive}, \textit{Trustworthy}, \textit{Intelligent}, and \textit{Hardworking}, all 4 judges place the \textit{East Asian} and \textit{Masculine/Receding Hairline} auto-labeled axes below the baseline. This association is target-dependent and not a uniform prior over the slider: for \textit{Youthful} the same \textit{East Asian} slider sits well above baseline for 3/4 judges, as does \textit{Masculine/Receding Hairline} for \textit{Criminal}. Conversely, the \textit{Feminine} axis places below baseline for \textit{Criminal} across all 4 judges and, more weakly, for \textit{Intelligent} across all four, while being above baseline for \textit{Attractive} and \textit{Trustworthy} in 3/4. Taken together across judges, the axes auto-interpreted as masculinity raise perceived criminality and lower perceived attractiveness, trustworthiness, and intelligence, and femininity inverts much of that ordering (though not on intelligence).

We emphasize that these are properties of the (FLUX + SliderSpace + judge) system as probed, and that the consistency we report is consistency of direction across models.

\subsection{Top \& Bottom Sliders per Model}
\label{sec:results_by_model}
\begin{figure}[!htb]
    \centering
    \includegraphics[width=\linewidth]{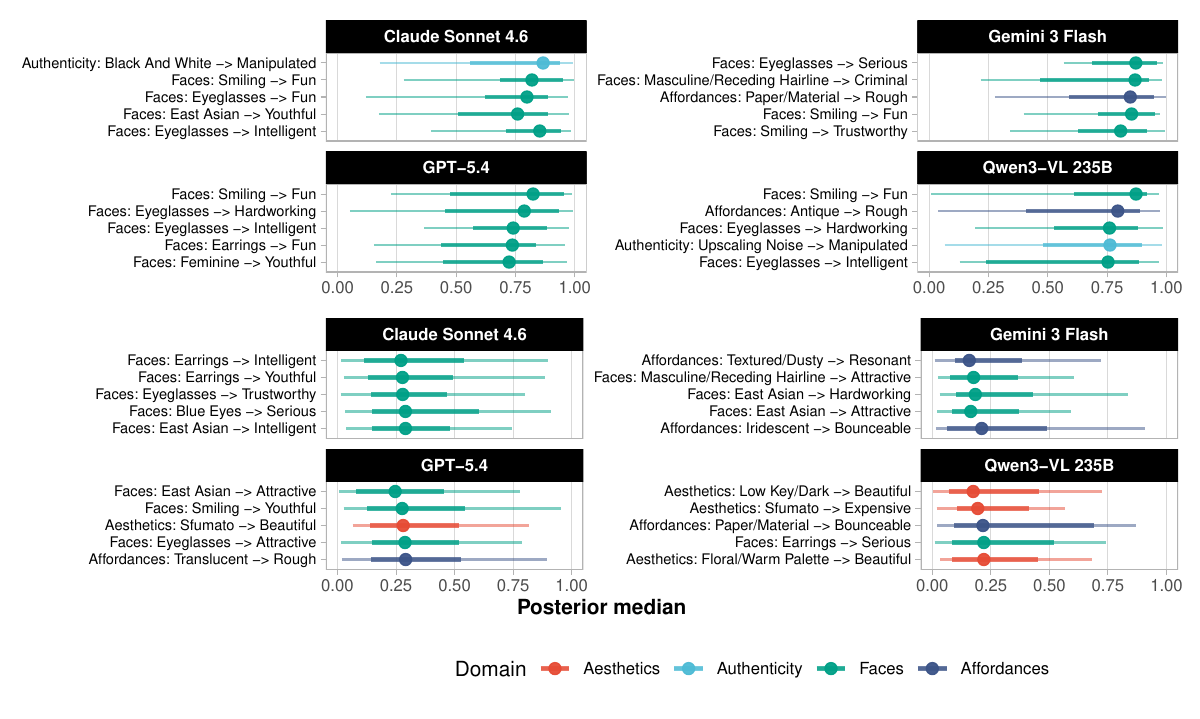}
    \caption{Top-5 (\textbf{top 2 rows}) and bottom-5 (\textbf{bottom 2 rows}) (target, slider) conditions per model. Points are per-model posterior medians, bars the $50\%$ credible intervals.}
    \label{fig:by_model}
\end{figure}

\Cref{fig:by_model} unpacks the pooled signal into per-model top (first 2 rows) and bottom (last 2 rows) rankings. \textit{Gemini 3 Flash} produces the strongest extremes in both directions: for instance, it places \textit{Masculine/Receding Hairline} at $0.87$ for \textit{Criminal} (vs.\ pooled $0.65$) but pushes \textit{East Asian} for \textit{Attractive} down to $0.16$, the most extreme version of the aforementioned bias. \textit{Claude Sonnet 4.6} and \textit{GPT-5.4} are comparatively less extreme. Claude is most often out of step with the others: it puts low utility on \textit{Glasses} for \textit{Trustworthy} ($0.28$) and its low magnitudes are uniformly gentler ($0.27$--$0.29$) than Gemini's ($0.16$--$0.23$). The agreement structure quantified next is partly a consequence of these per-model styles.

\subsection{Cross-model Rank Agreement}
\label{sec:results_agreement}
\begin{figure}[!htb]
    \centering
    \includegraphics[width=\linewidth]{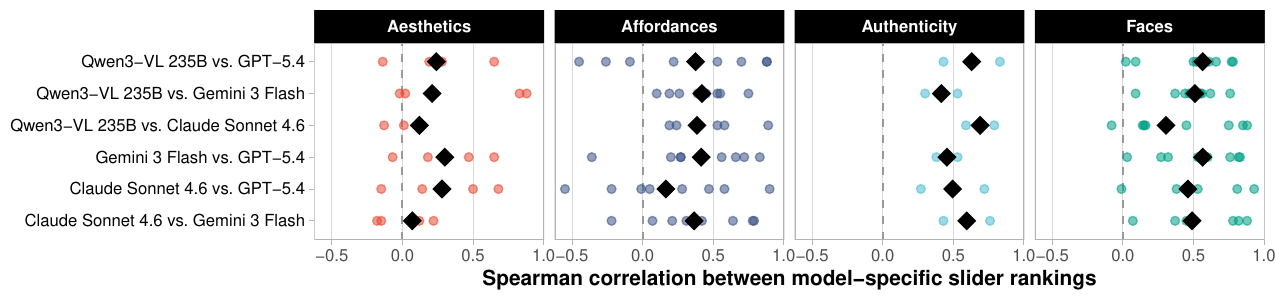}
    \caption{Cross-model rank agreement on recovered slider values, per domain, across the six model pairs. Higher $\rho$ means more consensus on which sliders matter for the targets in that domain.}
    \label{fig:model_agreement}
\end{figure}

\Cref{fig:model_agreement} reports cross-model Spearman rank correlations on the recovered slider values, by domain. \textit{Authenticity} is the highest-consensus domain (median $\rho \in [0.42, 0.69]$ across the six model pairs); \textit{Faces} and \textit{Affordances} are intermediate ($[0.30, 0.56]$); \textit{Aesthetics} is the lowest ($[0.07, 0.30]$). When the concept has a sharp visual signature (e.g., an edited photograph), models agree on the slider it cues; when the concept is diffuse (e.g., \textit{cheap} or \textit{experimental} aesthetics), they pick up different correlates. \textit{Claude Sonnet 4.6} is in the lowest-$\rho$ pair for both \textit{Faces} (Claude--Qwen, $\rho = 0.30$) and \textit{Aesthetics} (Claude--Gemini, $0.07$), consistent with \Cref{sec:results_by_model}, where Claude was the model most often out of step.

\subsection{Top-slider Selectivity per Target}
\label{sec:results_selectivity}
\begin{figure}[!htb]
    \centering
    \includegraphics[width=\linewidth]{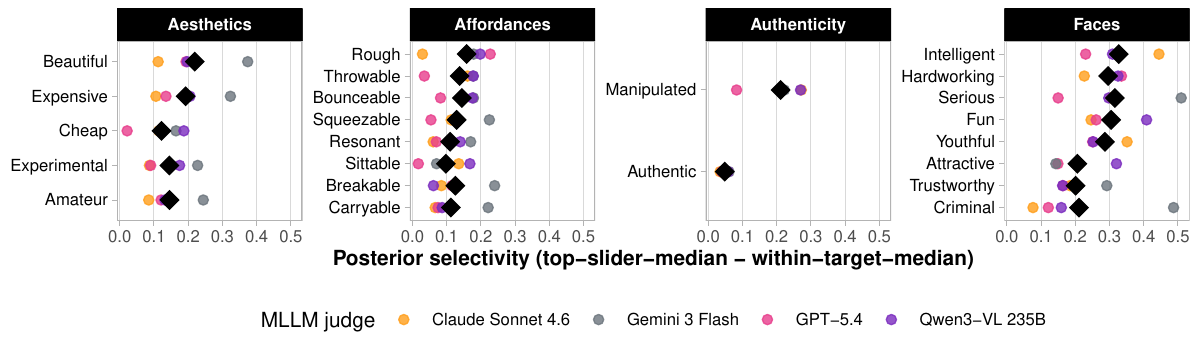}
    \caption{Per-target selectivity of the top-recovered slider, defined as the gap between the top slider's posterior median and the median posterior median across all 10 sliders for that target (per model, then averaged across models).}
    \label{fig:selectivity}
\end{figure}

\Cref{fig:selectivity} reports selectivity, the gap between the top slider and the rest of that target's sliders. Face sliders: \textit{Intelligent}/\textit{Eyeglasses} ($0.33$), \textit{Serious}/\textit{Eyeglasses} ($0.32$), \textit{Fun}/\textit{Smiling} ($0.30$), \textit{Hardworking}/\textit{Eyeglasses} ($0.30$). \textit{Authentic}---with \textit{Natural Edit} as its nominal top slider---has selectivity $0.05$ across models, close to flat. \textit{Cheap} is also diffuse (Floral Warm/Palette = $0.12$). These results corroborate the absence of \textit{Authentic} and \textit{Cheap} from \Cref{fig:headline_priors}.

\subsection{Explicit-Prior Baselines}
\label{sec:explicit_priors}

Our sampler is designed to recover a joint implicit prior over the selected slider space. A natural baseline is to ask whether simpler, explicit elicitation procedures recover the same structure. We therefore compare IMH against two direct baselines that query the model on fixed slider sweeps: a scalar rating task (how well does each sample match the target?) and a paired forced-choice task (between two samples, which better matches the target? This is the same binary choice primitive as our primary IMH method). The full set of experiments is described in \Cref{app:explicit_priors}, which shows that these recover more uniform-like (less informative) priors, and that their misalignment from the IMH-Gibbs approach correlates with inter-slider correlations (which explicit methods assume are effectively zero).

Finally, we give additional convergence, predictive, and interaction diagnostics in \Cref{app:diagnostics}.

\section{Limitations}
\label{sec:limitations}
Our method inherits its expressiveness from the underlying generative model: the recovered prior can only span what SliderSpace can learn and FLUX can express, and any concept absent from or poorly represented by the latent space is invisible to the chain. While SliderSpace is trained to produce orthogonal sliders, multi-attribute interactions could in principle violate this. Finally, while the recovered priors expose biases that direct prompting hides, our method does not by itself attribute them; downstream causal analysis is needed to determine whether a recovered preference reflects pre-training data, post-training alignment, or in-context confounds.

\section{Conclusion}
\label{sec:conclusion}

Cognitive scientists have spent decades building methods to extract what humans cannot say, the priors their perceptual systems carry into every decision. With recent AI models, our traditional practice has been to prompt them. However, as these models start adjudicating consequential decisions, and direct prompting reveals little of what they actually perceive, we need methods that let the models' own choices, rather than their introspective reports, drive the search. Characterizing what models bring to a task, before they act on it, is increasingly a prerequisite for designing and governing them responsibly. We show that the same sampling tradition, a Gibbs sampler whose acceptance step is the model's own pairwise preference, transfers when run over inputs in the model's native perceptual modality, recovering full posteriors. Finally, while surfacing the priors of a model has positive impacts for auditing them, it also makes implicit priors legible to potential bad actors who may amplify or exploit them in downstream pipelines.

\begin{ack}
We received funding from SK Telecom with MIT's Generative AI Impact Consortium (MGAIC). Research reported in this publication was supported by an Amazon Research Award, Fall 2024. Experiments conducted in this paper were generously supported via API credits provided by OpenAI, Anthropic, and Google. MC is supported by a fellowship from ``la Caixa'' Foundation (ID 100010434) with code LCF/BQ/EU23/12010079. The authors acknowledge the MIT Office of Research Computing and Data for providing high performance computing resources that have contributed to the research results reported within this paper.
\end{ack}

\bibliography{refs}
\bibliographystyle{apalike2}

%%%%%%%%%%%%%%%%%%%%%%%%%%%%%%%%%%%%%%%%%%%%%%%%%%%%%%%%%%%%

\appendix

\section{Examples}
\label{app:examples}

\begin{figure}[h]
    \centering
    \includegraphics[trim={8cm 5cm 8cm 4cm},clip,width=0.8\linewidth]{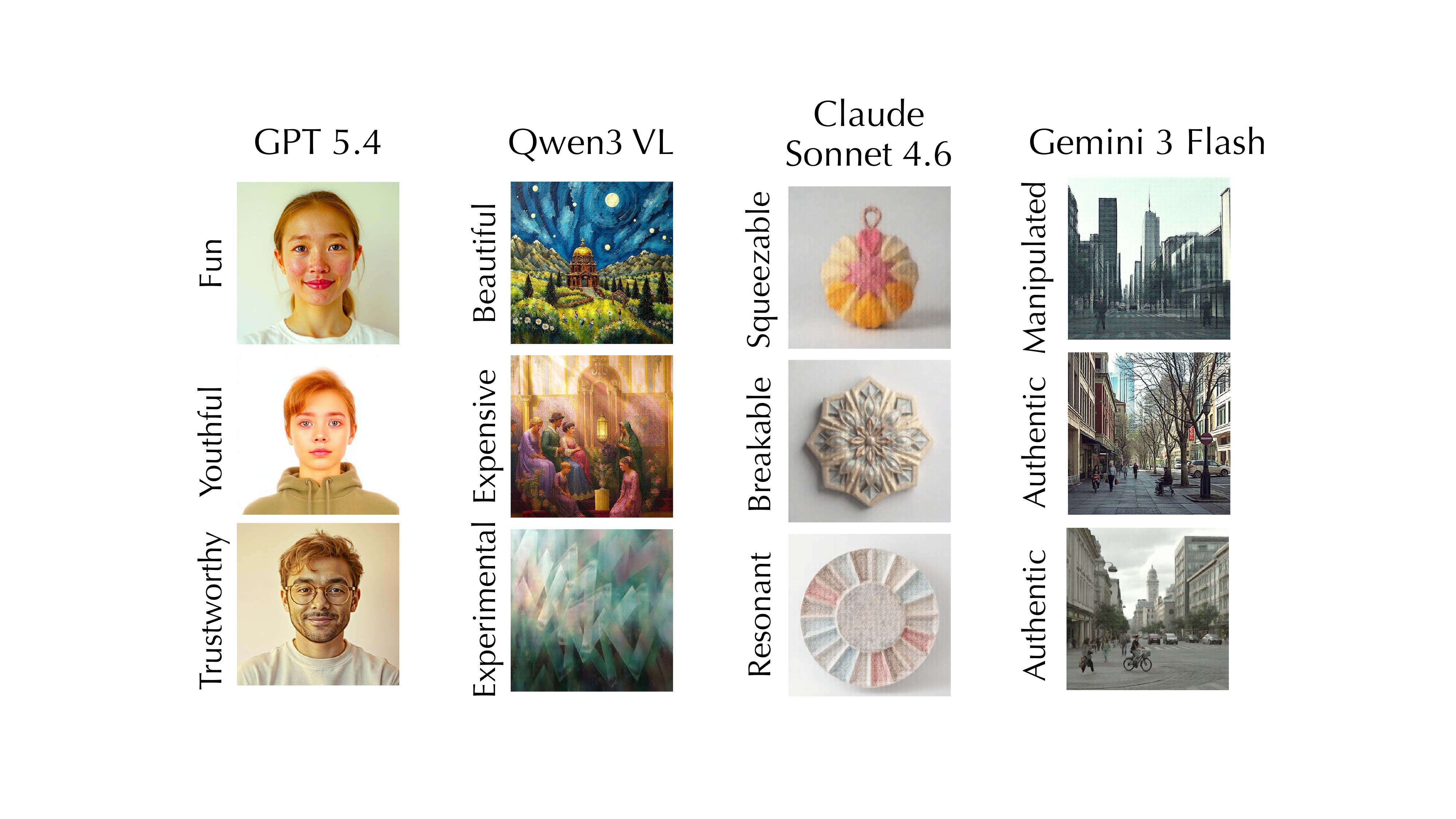}
    \caption{Example of recovered priors across the four domains (faces, affordances, aesthetics, authenticity), with one domain per model.}
    \label{fig:examples}
\end{figure}

We show qualitative examples of the recovered priors across the four domains in \Cref{fig:examples}. Each column corresponds to a (target, model) pair.

\section{Sliders}
\label{app:sliders}

\begin{figure}[!htb]
    \centering
    \includegraphics[width=\linewidth]{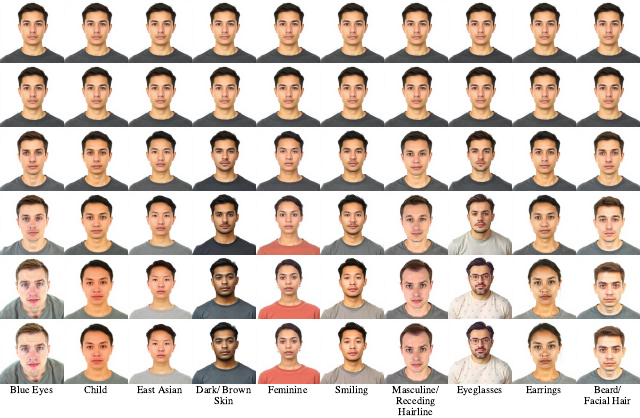}
    \caption{Examples for the face sliders trained with SliderSpace. The top image is the original, progressively steered towards the bottom. Labels are automatically generated (see \Cref{app:autointerp}).}
    \label{fig:sliders_faces}
\end{figure}

\begin{figure}[!htb]
    \centering
    \includegraphics[width=\linewidth]{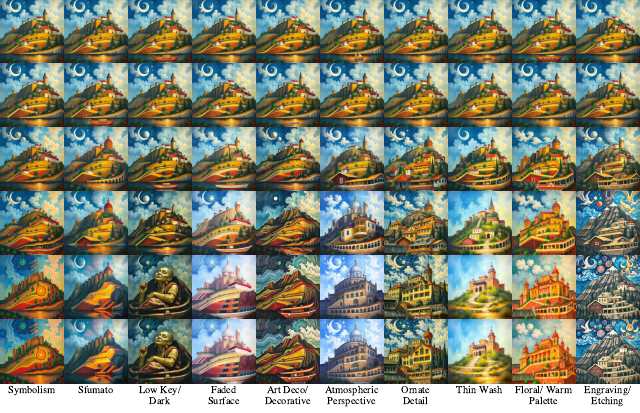}
    \caption{Examples for the aesthetics sliders trained with SliderSpace. The top image is the original, progressively steered towards the bottom. Labels are automatically generated (see \Cref{app:autointerp}).}
    \label{fig:sliders_art}
\end{figure}

\begin{figure}[!htb]
    \centering
    \includegraphics[width=\linewidth]{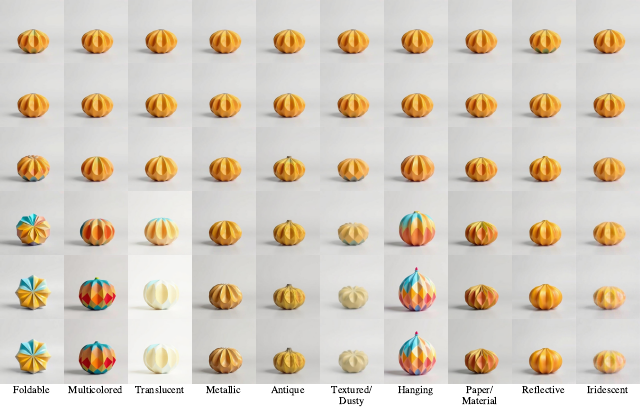}
    \caption{Examples for the affordances sliders trained with SliderSpace. The top image is the original, progressively steered towards the bottom. Labels are automatically generated (see \Cref{app:autointerp}).}
    \label{fig:sliders_affordances}
\end{figure}

\begin{figure}[!htb]
    \centering
    \includegraphics[width=\linewidth]{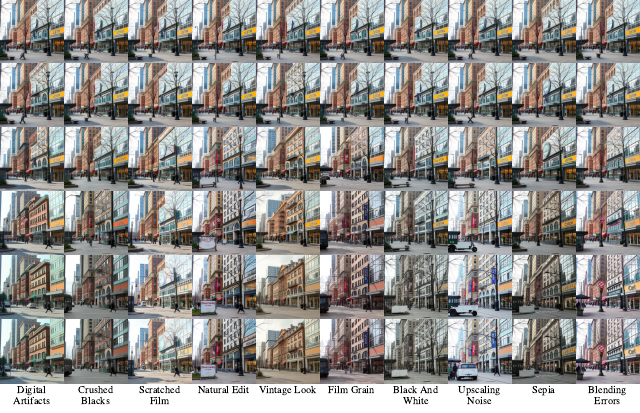}
    \caption{Examples for the authenticity sliders trained with SliderSpace. The top image is the original, progressively steered towards the bottom. Labels are automatically generated (see \Cref{app:autointerp}).}
    \label{fig:sliders_authenticity}
\end{figure}

For each domain, \Cref{fig:sliders_faces,fig:sliders_art,fig:sliders_affordances,fig:sliders_authenticity} show qualitative sweeps of the $d=10$ sliders selected from the $n=64$ SliderSpace candidates: each starts from the same seed image and progressively increases the slider's value while holding the others fixed. The sweeps illustrate what each axis controls and that the selected sliders produce coherent, monotone semantic changes when varied in isolation.

\section{Comparison to Explicit Priors}
\label{app:explicit_priors}
The IMH sampler is designed to recover a joint implicit prior, i.e. it observes which images a model prefers during a sequence of adaptive pairwise comparisons and turns those choices into samples from each model's target-conditioned distribution over the slider space. A simpler alternative is to query the model explicitly on a fixed set of slider sweeps. We implement and compare two such baselines, described below.

First, for each domain, we construct direct slider sweeps. For each chain, we first sample a base slider vector and a fixed generation seed. For each slider, we then vary only it while holding the other nine slider values and the generation seed fixed. Thus, within a sweep, visible changes are causally attributable to one slider. This design gives the explicit priors baseline the cleanest possible one-dimensional evidence about each control.

\paragraph{Rating baseline.}
In the rating baseline, the model sees one image at a time and assigns a score from 1 to 10 indicating how well the image matches the target. For a given target and slider, each image contributes its slider value and its rating. We normalize the ratings into weights and compute weighted posterior summaries over the slider coordinate. This produces a one-dimensional explicit posterior for every target--slider pair and is a direct analog of asking the model ``how much does this isolated slider setting match the target concept?''

\paragraph{Pairwise Bradley--Terry baseline.}
Here the model instead gets two images from the same slider sweep and chooses which image better matches the target (with randomized pair order). For each target and slider, we fit a Bradley--Terry model in which the latent utility of an image is linear in the value of the varied slider:
\begin{equation}
    \Pr(a \succ b) = \dfrac{\exp(\beta z_a)}{\exp(\beta z_a) + \exp(\beta z_b)}
\end{equation}
The fitted utility $\beta z$ defines a set of importance weights over the swept slider values, from which we compute weighted posterior medians and intervals.

\begin{figure*}[!htb]
    \centering
    \includegraphics[width=0.9\linewidth]{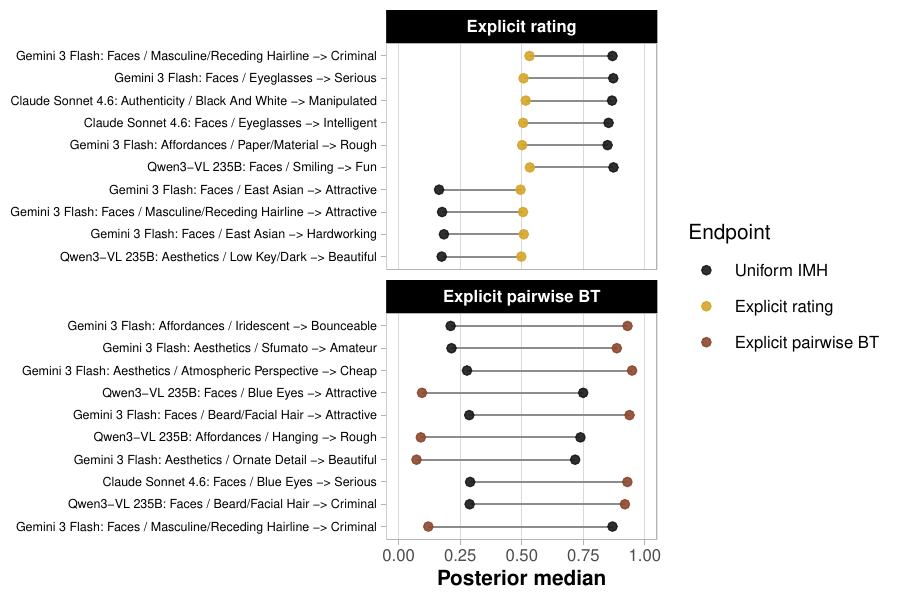}
    \caption{Magnitude of misalignment between explicit-prior and implicit Gibbs representations. Bars summarize how strongly each model--domain setting departs from the explicit slider coordinate system, aggregating across sliders and chains.}
    \label{fig:explicit-implicit-misalignment-leaderboard}
\end{figure*}

\begin{figure*}[!htb]
    \centering
    \includegraphics[width=0.9\linewidth]{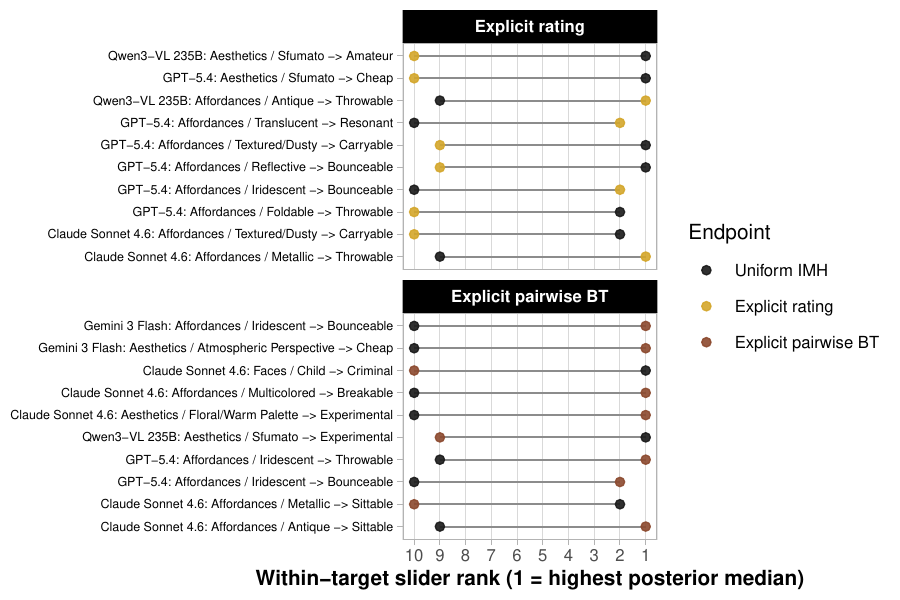}
    \caption{Rank misalignment between explicit-prior sliders and implicit Gibbs sliders across domains and models. For each setting, we compare the ordering induced by explicit slider values to the ordering recovered from implicit samples, with higher values indicating larger disagreement.}
    \label{fig:explicit-implicit-rank-misalignment-leaderboard}
\end{figure*}

\paragraph{Magnitude and rank disagreement.}
Then, for each target, model, domain, and slider, we compare the explicit posterior median to the IMH posterior median. The signed difference gives the direction and magnitude of disagreement. Because the procedures can differ in calibration, we also compare centered profiles, subtracting the mean posterior median across the sliders before computing differences. Finally, we rank sliders within each target by posterior median and compute rank displacement between explicit and IMH rankings, to study whether these methods identify different sliders as important. Results are shown in \Cref{fig:explicit-implicit-misalignment-leaderboard,fig:explicit-implicit-rank-misalignment-leaderboard} for the top-10 disagreements by each method.

\begin{figure*}[!htb]
    \centering
    \includegraphics[width=\linewidth]{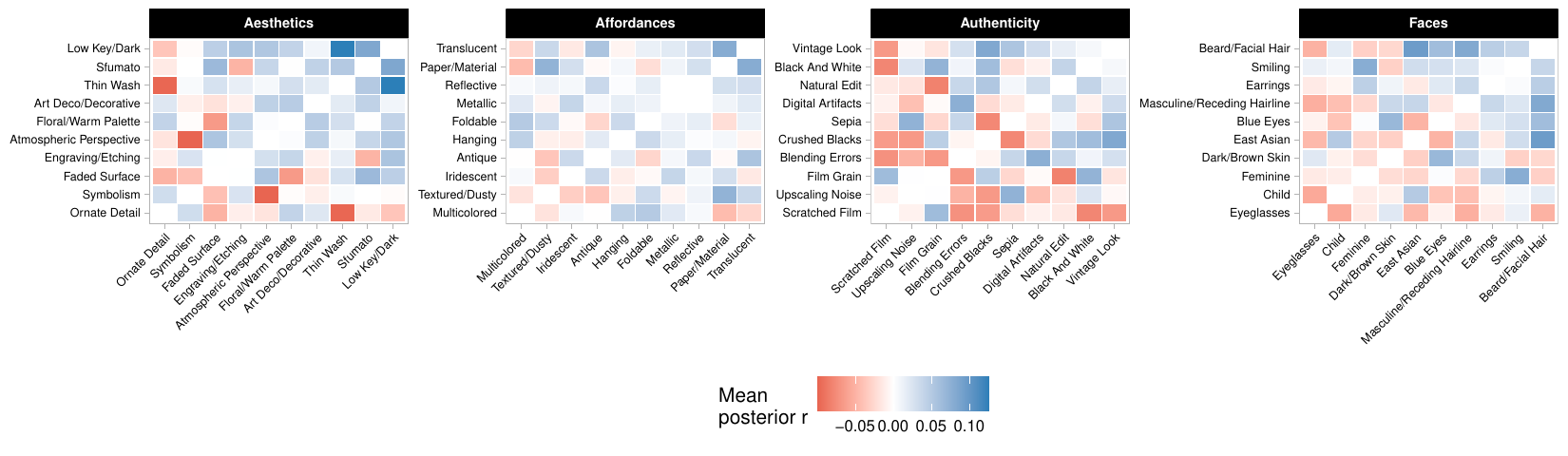}
    \caption{Correlation structure among learned slider dimensions. Each condition reports the empirical correlation between a pair of slider coordinates across sampled images, revealing whether sliders vary independently or are entangled. Explicit methods assume all off-diagonal entries are zero.}
    \label{fig:slider-corr-heatmap}
\end{figure*}

\begin{figure*}[!htb]
    \centering
    \includegraphics[width=0.5\linewidth]{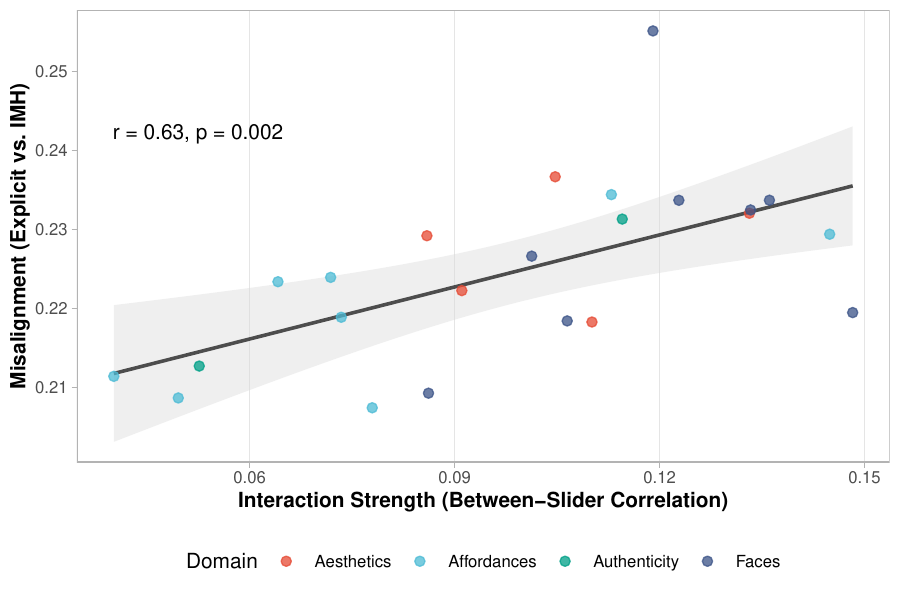}
    \caption{Relationship between slider interaction strength (X-axis) and explicit-implicit misalignment (Y-axis), showing that misalignments are larger as sliders are more correlated.}
    \label{fig:interaction-strength-vs-misalignment-global}
\end{figure*}

The explicit baselines are by construction insensitive to posterior dependence among sliders. IMH, by contrast, samples the joint distribution and can show that a model prefers combinations of features beyond independent marginal changes. To quantify this difference, we estimate slider--slider correlations from IMH draws and summarize each target by the mean absolute off-diagonal correlation. We then ask whether targets with stronger slider interactions are also the targets where explicit and implicit priors disagree most. \Cref{fig:slider-corr-heatmap} shows the off-diagonal correlations ignored by the explicit methods, and \Cref{fig:interaction-strength-vs-misalignment-global} shows that misalignment and interaction strength are correlated with Spearman $\rho=0.63, p=0.02$.

\paragraph{Informativeness.}
Finally, we quantify how much each elicitation procedure moves beyond the uninformative baseline of a uniform slider prior. For this, we compute the KL divergence from the recovered slider posterior to the uniform prior. For IMH, this is estimated from binned post-burn-in posterior draws. For the explicit baselines, it is computed as the discrete KL divergence of the induced weights over their fixed sampled design. This quantity should be interpreted as posterior concentration (however, note that a concentrated posterior may still be wrong if the elicitation procedure is biased or misspecified). Nevertheless, it estimates the value added over assuming a blank uniform posterior.

\begin{figure*}[!htb]
    \centering
    \includegraphics[width=\linewidth]{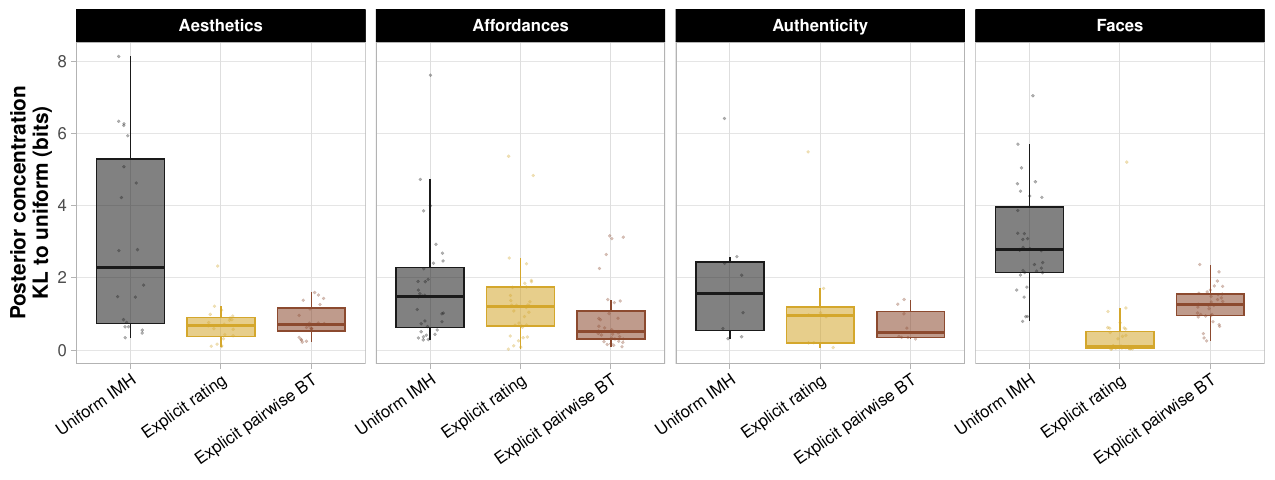}
    \caption{Posterior concentration relative to a uniform slider prior. Each point is a model--target combination, and the y-axis reports KL divergence from the recovered posterior to the corresponding uniform baseline. Larger values indicate that the elicitation procedure recovered a posterior that departs more strongly from an uninformative uniform prior.}
    \label{fig:posterior-concentration-kl-by-analysis}
\end{figure*}

\section{Auto-Interpretability Pipeline}
\label{app:autointerp}
We interpret each latent slider by matching image trajectories to a vocabulary of visual descriptors. The base candidate label set was generated by Claude 5 Fable, given high-level descriptions of the task and application domains. This resulted in a total of 867 candidate labels, across  Affordances (227), Art/Aesthetics (206), Authenticity (173), and Faces (261). Candidate labels and sampled images are embedded into a shared multimodal embedding space using \texttt{Gemini Embedding 2}.

\paragraph{Image sampling.}
We use a variant of the explicit-prior baseline images (described in \Cref{app:explicit_priors}) where instead all sliders are set to 0 except the focal slider for a given chain, and we run 5 chains (here, different seeds/starting images) for each slider. We sort images within each $(\text{domain}, \text{chain}, \text{slider})$ by the corresponding slider value.

\paragraph{Slider scoring.}
For each slider $j$, chain $c$, and candidate label $\ell$, we compute image--text similarities:
\begin{equation}
    s_{ic\ell} = \langle f_{\mathrm{img}}(x_{ic}), f_{\mathrm{text}}(\ell) \rangle
\end{equation}

To reduce generic labels, similarities are residualized by subtracting the per-image mean similarity across candidate labels:
\begin{equation}
    \tilde{s}_{ic\ell} = s_{ic\ell} - \frac{1}{|\mathcal{L}|} \sum_{\ell' \in \mathcal{L}} s_{ic\ell'}
\end{equation}

We score labels using endpoint contrast. Let $L$ and $H$ denote the low- and high-slider endpoint sets, respectively. By default, each endpoint contains the outer $25\%$ of samples along the slider. The per-chain endpoint score is:
\begin{equation}
    S_{cj\ell} =
    \max\left(0,\ \frac{1}{|H|} \sum_{i \in H} \tilde{s}_{ic\ell} - \frac{1}{|L|} \sum_{i \in L} \tilde{s}_{ic\ell}\right)
\end{equation}

Thus, labels are rewarded when their image-text similarity increases from the low end to the high end of the slider, and negative contrasts are zeroed out.

\paragraph{Chain weighting.}
Some chains produce more visually salient slider sweeps than others, because the generation seed and the nuisance slider settings can make the same control more or less visible. We therefore weight chains by the amount of image-embedding movement induced by the current slider. However, we need a way to estimate the overall movement in a coherent direction.

For each chain $c$ and slider $j$, let $\mathbf{e}_{ic}=f_{\mathrm{img}}(x_{ic})$ be the image embedding for sample $i$, and let $z_{icj}$ be its value on the current slider. We fit:
\begin{equation}
    \mathbf{e}_{ic} = \boldsymbol{\alpha}_{cj} + z_{icj}\boldsymbol{\beta}_{cj} + \boldsymbol{\varepsilon}_{ic}
\end{equation}
and use the slope norm
\begin{equation}
    m_{cj} = \left\|\hat{\boldsymbol{\beta}}_{cj}\right\|_2
\end{equation}
as a chain-level movement score. These movement scores are normalized within slider and converted into softmax-squared aggregation weights, so chains with stronger coherent embedding movement contribute more to the final label score.

\paragraph{Global label assignment.}
To avoid duplicate final labels within a domain, we solve a domain-level assignment problem. For each domain, we build a slider $\times$ label score matrix using the top candidate labels per slider, using the Hungarian algorithm to select one label per slider and uses each canonical label at most once within the domain.

\paragraph{LM refinement.}
Finally, we perform a post-hoc label refinement step over the assigned label and top runner-up labels, with five candidate labels per slider by default. The refinement prompt includes the endpoint scores. The refinement is performed in a single call to \texttt{Gemini 3.1 Flash Lite}. The model is instructed to summarize compatible candidate labels, resolve conflicts using scores and visual plausibility, and enforce semantic uniqueness within each domain. The refined label is used in downstream outputs. \Cref{tab:autointerp-labels} shows the results across all 40 sliders.

\begin{promptbox}
You are summarizing slider labels for an image analysis project.

For each slider, write one concise, visually plausible final label that summarizes the common visual concept across the candidate labels.

Candidate scores are nonnegative evidence strengths from the selected image-embedding scoring method; use them to resolve conflicts.

Do not simply copy the top-ranked label when several high-ranked labels express the same broader idea.

Combine compatible labels into a natural summary, for example \{sad, unhappy, frowning, grumpy\} -> Sad.

When candidate labels conflict, defer to labels that are more visually reasonable, general enough to use as a slider interpretation, and supported by stronger scores.

Before returning labels, review all sliders within each domain together and enforce semantic uniqueness across that domain.

Final labels must be semantically unique within each domain, not just textually different.

Do not assign near-duplicate concepts to separate sliders, for example Woman/Feminine/Female, Man/Masculine/Male, Happy/Smiling, Old/Aged, Young/Youthful, Dark/Black, or Large/Big.

If two sliders point to overlapping concepts, keep the concept only for the slider where it is most strongly and reasonably supported, then choose a distinct runner-up interpretation for the other slider.

Avoid overly narrow/specific or awkward labels unless the candidate set clearly supports them.

Use a brief noun phrase or adjective phrase, not a sentence.

Return only JSON with this schema:
\begin{verbatim}
{
    "labels": [
        {
            "domain": "faces",
            "slider_idx": 0,
            "final_label": "Sad/Frowning"
        }
    ]
}
\end{verbatim}
\end{promptbox}

{\scriptsize
\begin{longtable}{llp{0.14\linewidth}|p{0.1\linewidth} p{0.1\linewidth} p{0.1\linewidth} p{0.1\linewidth} p{0.1\linewidth}}
\caption{Top candidate labels and post-hoc refined labels used for slider interpretation.}\\
\label{tab:autointerp-labels}\\
\toprule
 & \S & Final & Top 1 & Top 2 & Top 3 & Top 4 & Top 5 \\
\midrule
\multirow{10}{*}{\rotatebox[origin=c]{90}{\textbf{Affordances}}} & v1 & Foldable & Folded & Foldable & Metallic & Pristine & Bronze \\
 & v2 & Multicolored & Multicolored & Stained Glass & Red & Frayed & Hanging \\
 & v3 & Translucent & White & Cream & Frosted Glass & Translucent & Beige \\
 & v4 & Metallic & Bronze & Gold Colored & Copper Colored & Antique & Copper \\
 & v5 & Antique & Antique & Dusty & Tarnished & Old & Hammered \\
 & v6 & Textured/Dusty & Gray & Dusty & Granite & Plaster & Silver Colored \\
 & v7 & Hanging & Hangable & Multicolored & Stained Glass & Hanging & Pink \\
 & v8 & Paper/Material & Foldable & Folded & Paper & Cardboard & Antique \\
 & v9 & Reflective & Glossy & Bronze & Reflective & Chrome & Copper Colored \\
 & v10 & Iridescent & Translucent & Iridescent & Pink & Frosted Glass & Transparent \\

\midrule
\multirow{10}{*}{\rotatebox[origin=c]{90}{\textbf{Aesthetics}}} & v1 & Symbolism & Pre-Raphaelite & Symbolic Imagery & Symbolism & Allegorical & Analogous Colors \\
 & v2 & Sfumato & Sfumato & Digital Painting & Mythological Scene & Dramatic Mood & Drybrush \\
 & v3 & Low Key/Dark & Low Key & Dark Tonality & Gothic & Mythological Scene & Tenebrism \\
 & v4 & Faded Surface & Faded Surface & Architectural Subject & Atmospheric Perspective & Thin Wash & Futurism \\
 & v5 & Art Deco/Decorative & Art Deco & Decorative Pattern & Futurism & Distorted Forms & Linocut \\
 & v6 & Atmospheric Perspective & Atmospheric Perspective & Tonalism & Seascape & High Key & Luminous \\
 & v7 & Ornate Detail & Ornate Detail & Dense Detail & Encaustic & Art Nouveau & Gold Leaf \\
 & v8 & Thin Wash & Thin Wash & Atmospheric Perspective & Tonalism & Architectural Subject & Underpainting Visible \\
 & v9 & Floral/Warm Palette & Floral Subject & Warm Palette & Architectural Subject & Encaustic & Pre-Raphaelite \\
 & v10 & Engraving/Etching & Dense Detail & Engraving & Etching & Ornate Detail & Pen And Ink \\

\midrule
\multirow{10}{*}{\rotatebox[origin=c]{90}{\textbf{Authenticity}}} & v1 & Digital Artifacts & AI Generated Look & Face Swap Artifacts & Photorealistic Render & Scratched Film & Green Tint \\
 & v2 & Crushed Blacks & Crushed Blacks & Scratched Film & Black And White & Smoky & Teal And Orange Grade \\
 & v3 & Scratched Film & Scratched Film & Sepia & Reflections Present & Faded Film & Edge Halos \\
 & v4 & Natural Edit & Natural Edit & Vintage Look & Black And White & Dramatic Lighting & Cool White Balance \\
 & v5 & Vintage Look & Vintage Look & Sepia & Object Removal Traces & Incorrect Scale & Faded Film \\
 & v6 & Film Grain & Film Grain & Faded Film & Rephotographed Screen & Scratched Film & Scanned Print \\
 & v7 & Black And White & Black And White & Sepia & Rainy & Faded Film & Dodging And Burning \\
 & v8 & Upscaling Noise & Upscaling Artifacts & Digital Noise & JPEG Artifacts & Edge Halos & Deepfake Blending Seams \\
 & v9 & Sepia & Sepia & Black And White & Crushed Blacks & Vintage Look & Dust Spots \\
 & v10 & Blending Errors & Hair Blending Errors & Black And White & Purple Fringing & Selective Color Edit & Oversharpened \\

\midrule
\multirow{10}{*}{\rotatebox[origin=c]{90}{\textbf{Faces}}} & v1 & Blue Eyes & Blue Eyes & Dry Skin & Green Eyes & Dyed Hair & Crow's Feet \\
 & v2 & Child & Child & East Asian Ethnicity & Unnatural Hair Color & Dreadlocks & Dyed Hair \\
 & v3 & East Asian & East Asian Ethnicity & Bloodshot Eyes & Freckled Skin & Monolid Eyes & Southeast Asian Ethnicity \\
 & v4 & Dark/Brown Skin & Dark Brown Skin & Brown Skin & South Asian Ethnicity & Tan Skin & Black Ethnicity \\
 & v5 & Feminine & Earrings & Female Presenting & Feminine & No Makeup & Ponytail \\
 & v6 & Smiling & Smiling & East Asian Ethnicity & Southeast Asian Ethnicity & Grinning & Smirking \\
 & v7 & Masculine/Receding Hairline & Male Presenting & Receding Hairline & Blue Eyes & Stubble & Crow's Feet \\
 & v8 & Eyeglasses & Eyeglasses & Full Beard & Long Beard & Trimmed Beard & Stubble \\
 & v9 & Earrings & Female Presenting & Feminine & Earrings & Southeast Asian Ethnicity & Pacific Islander Ethnicity \\
 & v10 & Beard/Facial Hair & Stubble & Male Presenting & Long Beard & Receding Hairline & Full Beard \\
\bottomrule
\end{longtable}
}

\FloatBarrier

\section{Additional Diagnostics}
\label{app:diagnostics}

The main posterior summaries are marginal summaries of sampled slider values. We add two additional predictive diagnostics that use the accept/reject decisions to test whether the sampled marginal density has explanatory power for future proposals. For each proposal, we extract the current value and proposed value along the dimension being updated. We then estimate the marginal density of post-burn-in chain states along that same dimension using a kernel density estimate on the logit-transformed slider value. A proposal is predicted to be accepted when the estimated density at the proposed value is at least as large as the estimated density at the current value (chance = 0.5).

\begin{figure*}[!htb]
    \centering
    \includegraphics[width=\linewidth]{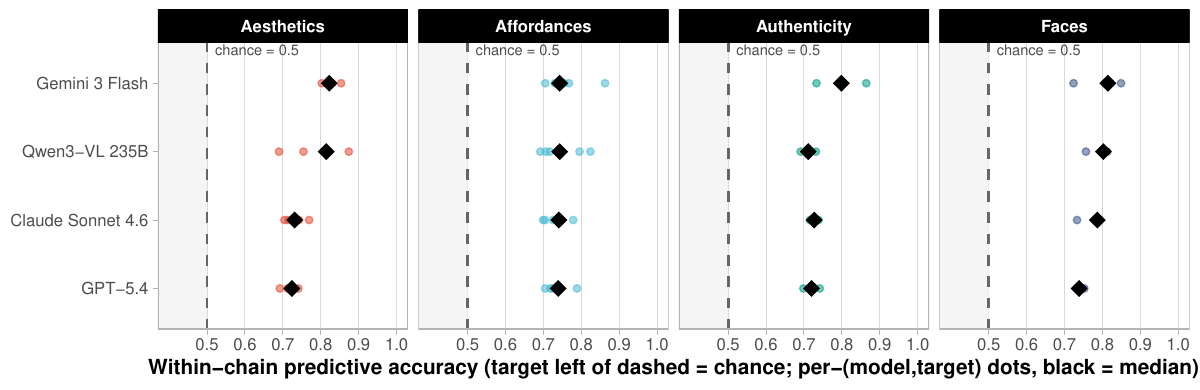}
    \caption{Within-chain predictive accuracy for slider-based models.}
    \label{fig:within-chain-predictive-accuracy}
\end{figure*}

\paragraph{Within-chain predictive check.}
First, we ask whether one part of a chain predicts held-out decisions from the same chain. For each chain and updated dimension, we split proposals into contiguous folds, fit the marginal density on the training folds, and evaluate accept/reject prediction on the held-out fold. High accuracy means the chain's own posterior mass is aligned with the model's later choices. Results, shown in \Cref{fig:within-chain-predictive-accuracy}, point to consistent 

\begin{figure*}[!htb]
    \centering
    \includegraphics[width=\linewidth]{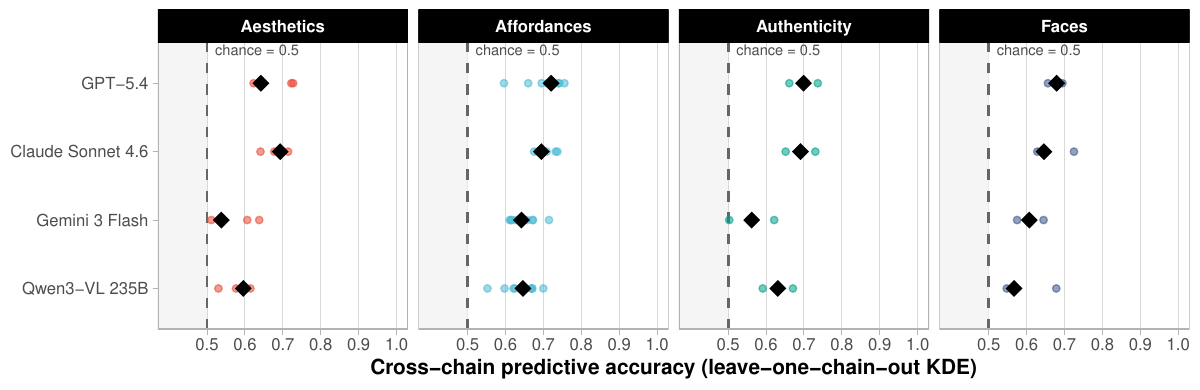}
    \caption{Cross-chain predictive accuracy for slider-based models.}
    \label{fig:cross-chain-predictive-accuracy}
\end{figure*}

\paragraph{Cross-chain predictive check.}
We then apply a stricter cross-chain check. For each target and updated dimension, we fit the marginal density using all chains except one and evaluate proposals from the held-out chain to test whether different seeds recover compatible posterior structure. Results are shown in \Cref{fig:cross-chain-predictive-accuracy}.

\subsection{Convergence}
\label{sec:results_convergence}
\Cref{tab:diagnostic_overview} reports per-(domain, model) convergence diagnostics aggregated over slider parameters, and \Cref{tab:diagnostic-flags} breaks the same chains down into the fraction of parameters that cross standard $\hat{R}$ and acceptance thresholds. \textit{GPT-5.4} and \textit{Claude Sonnet 4.6} produce the most-converged chains across all four domains (median split-$\hat{R} \in [1.02, 1.20]$); \textit{Qwen3-VL 235B} and \textit{Gemini 3 Flash} are systematically harder to mix, with median $\hat{R}$ reaching $1.6$ on \textit{Aesthetics}. \textit{Affordances} mixes comparably well for all models: even \textit{Qwen3-VL 235B} and \textit{Gemini 3 Flash} sit near median $\hat{R} = 1.15$ there, well below their \textit{Faces} and \textit{Aesthetics} values. Parameter-level flags in \Cref{tab:diagnostic-flags} show that for \textit{GPT-5.4} and \textit{Claude Sonnet 4.6}, almost no parameters exceed $\hat{R} = 1.5$ in any domain ($\le 1\%$), and none fall below $5\%$ acceptance. \textit{Qwen3-VL 235B} and \textit{Gemini 3 Flash} concentrate their poorly-mixing parameters in \textit{Aesthetics} ($50$--$58\%$ above $\hat{R} = 1.5$ and $16$--$18\%$ below $5\%$ acceptance) and to a lesser extent \textit{Faces} ($22$--$32\%$ above $\hat{R} = 1.5$); their \textit{Affordances} chains, by contrast, rarely cross either threshold despite a high share above the looser $\hat{R} = 1.1$ cutoff. Acceptance rates align with these observations: \textit{GPT-5.4} and \textit{Claude Sonnet 4.6} accept the proposal $20$--$40\%$ of the time on average while \textit{Qwen3-VL 235B} and \textit{Gemini 3 Flash} drop to $7$--$15\%$ on the harder domains. This suggests that the latter two may have more decisive implicit priors that make a random proposal harder to satisfy.

\begin{table}[!htb]
\centering
\caption{Per-(domain, model) convergence diagnostics, aggregated across the slider parameters of every target in the domain. \texttt{med.\ max\ $\hat{R}$} is the median across targets of the worst-mixing slider's $\hat{R}$; \texttt{med.\ $\hat{R}$} is the median across all (target, slider) pairs; \texttt{lag-1} is the mean lag-1 autocorrelation post-burn-in.}
\label{tab:diagnostic_overview}
\small
\begin{tabular}{llrrrr}
\toprule
Domain & Model & med.\ max $\hat{R}$ & worst $\hat{R}$ & med.\ $\hat{R}$ & lag-1 \\
\midrule
\multirow{4}{*}{Aesthetics} & Claude Sonnet 4.6 & 1.08 & 1.22 & 1.03 & 0.63 \\
& GPT-5.4 & 1.09 & 1.29 & 1.06 & 0.69 \\
& Qwen3-VL 235B & 2.10 & 2.74 & 1.60 & 0.86 \\
& Gemini 3 Flash & 1.96 & 3.20 & 1.56 & 0.88 \\
\addlinespace

\multirow{4}{*}{Authenticity} & Claude Sonnet 4.6 & 1.11 & 1.20 & 1.06 & 0.65 \\
& GPT-5.4 & 1.04 & 1.05 & 1.02 & 0.58 \\
& Qwen3-VL 235B & 1.28 & 1.45 & 1.08 & 0.76 \\
& Gemini 3 Flash & 1.66 & 2.07 & 1.39 & 0.81 \\
\addlinespace

\multirow{4}{*}{Affordances} & Claude Sonnet 4.6 & 1.08 & 1.20 & 1.03 & 0.60 \\
& GPT-5.4 & 1.04 & 1.55 & 1.02 & 0.61 \\
& Qwen3-VL 235B & 1.31 & 2.10 & 1.15 & 0.77 \\
& Gemini 3 Flash & 1.33 & 2.44 & 1.15 & 0.79 \\
\addlinespace

\multirow{4}{*}{Faces} & Claude Sonnet 4.6 & 1.32 & 1.78 & 1.20 & 0.77 \\
& GPT-5.4 & 1.25 & 1.42 & 1.11 & 0.77 \\
& Qwen3-VL 235B & 1.59 & 2.31 & 1.33 & 0.81 \\
& Gemini 3 Flash & 1.68 & 2.79 & 1.33 & 0.85 \\
\bottomrule
\end{tabular}
\end{table}

\begin{table}[!htb]
\centering
\caption{Parameter-level diagnostic checks by perceptual domain and MLLM judge (Acc. = acceptance rate $\in [0, 1]$).}
\label{tab:diagnostic-flags}
\small
\begin{tabular}{llrrrr}
\toprule
Domain & Model & $\hat{R} > 1.1$ & $\hat{R} >1.5$ & Acc. $< 5\%$ & med. Acc. \\
\midrule
Aesthetics & Claude Sonnet 4.6 & 16\% & 0\% & 0\% & 0.36 \\
Aesthetics & GPT-5.4 & 14\% & 0\% & 0\% & 0.28 \\
Aesthetics & Qwen3-VL 235B & 94\% & 50\% & 16\% & 0.08 \\
Aesthetics & Gemini 3 Flash & 100\% & 58\% & 18\% & 0.07 \\
\addlinespace

Authenticity & Claude Sonnet 4.6 & 15\% & 0\% & 0\% & 0.33 \\
Authenticity & GPT-5.4 & 0\% & 0\% & 0\% & 0.41 \\
Authenticity & Qwen3-VL 235B & 35\% & 0\% & 0\% & 0.21 \\
Authenticity & Gemini 3 Flash & 85\% & 40\% & 5\% & 0.14 \\
\addlinespace

Affordances & Claude Sonnet 4.6 & 5\% & 0\% & 0\% & 0.37 \\
Affordances & GPT-5.4 & 16\% & 1\% & 0\% & 0.42 \\
Affordances & Qwen3-VL 235B & 61\% & 12\% & 0\% & 0.21 \\
Affordances & Gemini 3 Flash & 61\% & 10\% & 0\% & 0.20 \\
\addlinespace

Faces & Claude Sonnet 4.6 & 62\% & 1\% & 0\% & 0.20 \\
Faces & GPT-5.4 & 49\% & 0\% & 0\% & 0.23 \\
Faces & Qwen3-VL 235B & 81\% & 22\% & 0\% & 0.14 \\
Faces & Gemini 3 Flash & 90\% & 32\% & 0\% & 0.15 \\
\bottomrule
\end{tabular}
\end{table}

\section{Random Walk Metropolis-Hastings}
\label{app:random_walk_mh}

In the main experiments, we drew the proposals independently from a $\mathrm{Uniform}([0,1])$ distribution. However, this can also be done with a random walk by using a symmetric Gaussian proposal centered on the current value, reflected at the boundaries. Symmetry is required because the MLLM only provides a binary choice. The initial proposal standard deviation is $\sigma_0 = 0.2$. We adapt $\sigma$ during the burn-in fraction of each chain (30\% by default) via Robbins--Monro \citep{robbins1951stochastic} to target a desired acceptance rate (default $0.27$ \citep{agrawal2023optimal}); once burn-in ends, $\sigma$ is frozen for the remainder of the chain.

In practice, the random-walk variant underperforms the uniform proposal of the main experiments, as the median $\hat{R}$ (see \Cref{tab:walk_diagnostics}) frequently exceeds $1.5$. The Robbins--Monro update often collapses $\sigma$ to $\lesssim 0.1$ in search of the $0.27$ target rate, but no $\sigma$ achieves it. Large moves leave the high-$\pi$ region and small moves still cross the local gradient sharply enough to be rejected, so acceptance plateaus at $0.05$--$0.15$ and the chain barely moves. It is possible to adapt one scale per slider per chain, which might yield better results, but at the cost of a longer burn-in and thus more trials.

\begin{table}[!htb]
\centering
\caption{Diagnostics for the random-walk variant on the faces experiment, aggregated across the $d=10$ sliders per (agent, target). Each condition reports the chain-mean post-burn-in proposal scale $\sigma$ followed by the median split-$\hat{R}$. Collapsed $\sigma \lesssim 0.1$ generally coincides with $\hat{R} > 2$, while stable $\sigma \in [0.4, 0.95]$ coincides with $\hat{R} \le 1.5$.}
\label{tab:walk_diagnostics}
\begin{tabular}{lcccc}
\toprule
Target & Qwen3-VL 235B & Claude Sonnet 4.6 & Gemini 3 Flash & GPT-5.4 \\
\midrule
attractive  & $0.05 \mid 2.95$ & $0.30 \mid 1.65$ & $0.13 \mid 2.19$ & $0.67 \mid 1.31$ \\
fun         & $0.03 \mid 4.67$ & $0.18 \mid 1.87$ & $0.15 \mid 2.11$ & $0.27 \mid 1.43$ \\
intelligent & $0.36 \mid 1.77$ & $0.25 \mid 1.56$ & $0.25 \mid 1.70$ & $0.32 \mid 1.69$ \\
serious     & $0.02 \mid 4.36$ & $0.46 \mid 1.60$ & $0.19 \mid 2.24$ & $0.43 \mid 1.38$ \\
trustworthy & $0.48 \mid 1.17$ & $0.06 \mid 3.44$ & $0.08 \mid 3.02$ & $0.52 \mid 1.35$ \\
youthful    & $0.08 \mid 2.59$ & $0.10 \mid 2.04$ & $0.02 \mid 7.28$ & $0.34 \mid 1.44$ \\
criminal    & $0.13 \mid 2.56$ & $0.67 \mid 1.19$ & $0.07 \mid 3.38$ & $0.85 \mid 1.13$ \\
hardworking & $0.46 \mid 1.47$ & $0.94 \mid 1.06$ & $0.33 \mid 1.37$ & $0.61 \mid 1.46$ \\
\bottomrule
\end{tabular}
\end{table}

\section{Validation with Color Experiment}
\label{app:colors}
To verify that our Gibbs+Barker kernel correctly samples from the model's induced prior, we run a validation experiment in a low-dimensional space where ground truth is unambiguous: HSL colors. The state $z \in [0,1]^3$ corresponds directly to the (hue, saturation, lightness) coordinates of a flat color tile, so the stimulus is fully determined by $z$ without any generative model in between. We pick the same eight color-name targets as \cite{harrison2020gibbs,zhu2024recovering}---\textit{sunset}, \textit{eggshell}, \textit{lavender}, \textit{chocolate}, \textit{lemon}, \textit{cloud}, \textit{strawberry}, and \textit{grass}---to compare against their human data. We run 10 independent chains of 2,000 trials each, using only \textit{Qwen3-VL 235B}. Because the color associated with each target is human-recognizable, deviations between the recovered posterior mode and the canonical color flag kernel-level errors that would be hard to spot in higher-dimensional settings.

\begin{figure}[!htb]
    \centering
    \begin{subfigure}{\linewidth}
        \centering
        \includegraphics[width=\linewidth]{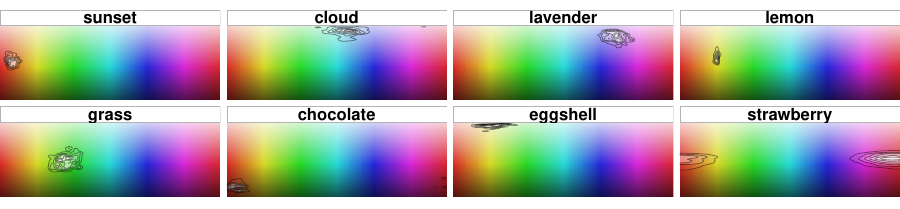}
        \caption{2D density overlays of the posterior over HSL coordinates for each target.}
        \label{fig:colors-hsl-density}
    \end{subfigure}

    \vspace{2pt}

    \begin{subfigure}{\linewidth}
        \centering
        \includegraphics[width=\linewidth]{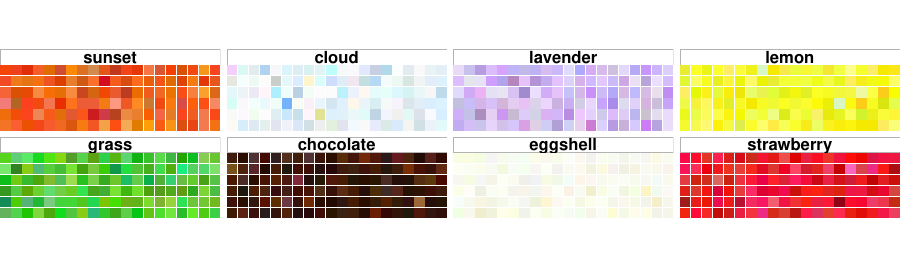}
        \caption{Posterior samples rendered as color swatches for each named target.}
        \label{fig:colors-posterior-tiles}
    \end{subfigure}
    \caption{Posterior color recovery for all eight named targets.}
    \label{fig:colors-joint}
\end{figure}

The recovered posteriors are visualized as 2D density overlays in \Cref{fig:colors-hsl-density}. \Cref{fig:colors-posterior-tiles} shows posterior samples for each named color. For every target, the central tendency falls within the canonical color associated with the name: saturated red-orange for \textit{sunset}, saturated yellow for \textit{lemon}, green for \textit{grass}, light purple for \textit{lavender}, dark red-brown for \textit{chocolate}, near-white with high lightness for \textit{cloud} and \textit{eggshell}, and saturated red for \textit{strawberry}. The kernel therefore recovers the correct region of the color space for every target, purely from binary preferences.

\section{Compute Resources}
\label{app:compute}

SliderSpace sliders were trained (see \Cref{app:sliderspace_training} for details) on a single NVIDIA H200 GPU (141\,GB VRAM). Per domain, data generation and PCA over the $50{,}000$ CLIP-encoded samples take approximately $3$ hours, and slider training takes approximately $6$ hours; total ${\approx}9$ hours/domain.

For the sampling experiments reported in the paper, we run FLUX.1-schnell inference on $40$ parallel Amazon SageMaker endpoints, each backed by a single NVIDIA L40S GPU ($48$\,GB VRAM). Each endpoint serves the same FLUX.1-schnell pipeline with the trained SliderSpace LoRAs, compiled and highly optimized to run inference at ${\sim}0.7$s per image.

\section{Posterior Summaries}
\label{app:posterior_summaries}

\Cref{tab:posteriors_affordances,tab:posteriors_art,tab:posteriors_authenticity,tab:posteriors_faces} show posterior summaries (medians with 50\% credible intervals) for all conditions across the primary IMH experiments reported in the main paper.

{
\footnotesize
\begin{longtable}{p{1.4cm}p{2cm}rrrr}
\caption{Posterior medians with 50\% credible intervals for Affordances.} \label{tab:posteriors_affordances} \\
\toprule
Target & Parameter & Claude Sonnet 4.6 & GPT-5.4 & Gemini 3 Flash & Qwen3-VL 235B \\
\midrule
\endfirsthead
\caption[]{Posterior medians with 50\% credible intervals for Affordances.} \\
\toprule
 & agent & Claude Sonnet 4.6 & GPT-5.4 & Gemini 3 Flash & Qwen3-VL 235B \\
target & parameter &  &  &  &  \\
\midrule
\endhead
\midrule
\multicolumn{6}{r}{Continued on next page} \\
\midrule
\endfoot
\bottomrule
\endlastfoot
\multirow[t]{10}{*}{Bounceable} & Antique & 0.40 {\tiny \color{gray} [0.19, 0.68]} & 0.52 {\tiny \color{gray} [0.27, 0.85]} & 0.39 {\tiny \color{gray} [0.18, 0.66]} & 0.50 {\tiny \color{gray} [0.24, 0.83]} \\
 & Foldable & 0.53 {\tiny \color{gray} [0.31, 0.77]} & 0.53 {\tiny \color{gray} [0.24, 0.67]} & 0.48 {\tiny \color{gray} [0.16, 0.86]} & 0.49 {\tiny \color{gray} [0.31, 0.77]} \\
 & Hanging & 0.54 {\tiny \color{gray} [0.29, 0.71]} & 0.54 {\tiny \color{gray} [0.29, 0.74]} & 0.40 {\tiny \color{gray} [0.20, 0.79]} & 0.40 {\tiny \color{gray} [0.31, 0.78]} \\
 & Iridescent & 0.47 {\tiny \color{gray} [0.25, 0.74]} & 0.41 {\tiny \color{gray} [0.15, 0.68]} & 0.21 {\tiny \color{gray} [0.06, 0.49]} & 0.46 {\tiny \color{gray} [0.14, 0.75]} \\
 & Metallic & 0.47 {\tiny \color{gray} [0.20, 0.70]} & 0.48 {\tiny \color{gray} [0.23, 0.68]} & 0.34 {\tiny \color{gray} [0.13, 0.53]} & 0.42 {\tiny \color{gray} [0.16, 0.56]} \\
 & Multicolored & 0.43 {\tiny \color{gray} [0.25, 0.76]} & 0.50 {\tiny \color{gray} [0.29, 0.76]} & 0.36 {\tiny \color{gray} [0.11, 0.71]} & 0.45 {\tiny \color{gray} [0.26, 0.78]} \\
 & Paper/Material & 0.37 {\tiny \color{gray} [0.19, 0.67]} & 0.49 {\tiny \color{gray} [0.26, 0.67]} & 0.33 {\tiny \color{gray} [0.10, 0.63]} & 0.22 {\tiny \color{gray} [0.09, 0.69]} \\
 & Reflective & 0.53 {\tiny \color{gray} [0.26, 0.75]} & 0.58 {\tiny \color{gray} [0.30, 0.82]} & 0.49 {\tiny \color{gray} [0.23, 0.76]} & 0.31 {\tiny \color{gray} [0.15, 0.65]} \\
 & Textured/Dusty & 0.56 {\tiny \color{gray} [0.25, 0.75]} & 0.46 {\tiny \color{gray} [0.18, 0.73]} & 0.58 {\tiny \color{gray} [0.37, 0.79]} & 0.46 {\tiny \color{gray} [0.28, 0.74]} \\
 & Translucent & 0.64 {\tiny \color{gray} [0.34, 0.85]} & 0.49 {\tiny \color{gray} [0.26, 0.73]} & 0.53 {\tiny \color{gray} [0.22, 0.72]} & 0.63 {\tiny \color{gray} [0.32, 0.88]} \\
\cline{1-6}
\multirow[t]{10}{*}{Breakable} & Antique & 0.43 {\tiny \color{gray} [0.21, 0.68]} & 0.45 {\tiny \color{gray} [0.23, 0.70]} & 0.49 {\tiny \color{gray} [0.27, 0.76]} & 0.43 {\tiny \color{gray} [0.22, 0.75]} \\
 & Foldable & 0.42 {\tiny \color{gray} [0.21, 0.65]} & 0.42 {\tiny \color{gray} [0.23, 0.67]} & 0.35 {\tiny \color{gray} [0.22, 0.60]} & 0.35 {\tiny \color{gray} [0.19, 0.62]} \\
 & Hanging & 0.51 {\tiny \color{gray} [0.27, 0.73]} & 0.60 {\tiny \color{gray} [0.30, 0.82]} & 0.60 {\tiny \color{gray} [0.29, 0.81]} & 0.50 {\tiny \color{gray} [0.28, 0.77]} \\
 & Iridescent & 0.44 {\tiny \color{gray} [0.24, 0.63]} & 0.44 {\tiny \color{gray} [0.23, 0.69]} & 0.37 {\tiny \color{gray} [0.21, 0.55]} & 0.47 {\tiny \color{gray} [0.24, 0.68]} \\
 & Metallic & 0.47 {\tiny \color{gray} [0.20, 0.73]} & 0.53 {\tiny \color{gray} [0.27, 0.76]} & 0.70 {\tiny \color{gray} [0.42, 0.84]} & 0.45 {\tiny \color{gray} [0.24, 0.71]} \\
 & Multicolored & 0.38 {\tiny \color{gray} [0.16, 0.64]} & 0.53 {\tiny \color{gray} [0.28, 0.77]} & 0.48 {\tiny \color{gray} [0.27, 0.76]} & 0.45 {\tiny \color{gray} [0.26, 0.71]} \\
 & Paper/Material & 0.52 {\tiny \color{gray} [0.24, 0.76]} & 0.53 {\tiny \color{gray} [0.25, 0.79]} & 0.68 {\tiny \color{gray} [0.37, 0.84]} & 0.51 {\tiny \color{gray} [0.23, 0.75]} \\
 & Reflective & 0.46 {\tiny \color{gray} [0.23, 0.71]} & 0.51 {\tiny \color{gray} [0.30, 0.73]} & 0.39 {\tiny \color{gray} [0.19, 0.65]} & 0.52 {\tiny \color{gray} [0.32, 0.74]} \\
 & Textured/Dusty & 0.44 {\tiny \color{gray} [0.21, 0.72]} & 0.45 {\tiny \color{gray} [0.22, 0.67]} & 0.33 {\tiny \color{gray} [0.16, 0.55]} & 0.49 {\tiny \color{gray} [0.24, 0.81]} \\
 & Translucent & 0.54 {\tiny \color{gray} [0.28, 0.80]} & 0.42 {\tiny \color{gray} [0.21, 0.68]} & 0.44 {\tiny \color{gray} [0.19, 0.69]} & 0.45 {\tiny \color{gray} [0.23, 0.67]} \\
\cline{1-6}
\multirow[t]{10}{*}{Carryable} & Antique & 0.50 {\tiny \color{gray} [0.23, 0.75]} & 0.51 {\tiny \color{gray} [0.24, 0.75]} & 0.55 {\tiny \color{gray} [0.30, 0.72]} & 0.51 {\tiny \color{gray} [0.25, 0.78]} \\
 & Foldable & 0.48 {\tiny \color{gray} [0.21, 0.73]} & 0.41 {\tiny \color{gray} [0.20, 0.70]} & 0.47 {\tiny \color{gray} [0.21, 0.80]} & 0.27 {\tiny \color{gray} [0.15, 0.56]} \\
 & Hanging & 0.45 {\tiny \color{gray} [0.19, 0.72]} & 0.41 {\tiny \color{gray} [0.20, 0.71]} & 0.36 {\tiny \color{gray} [0.20, 0.66]} & 0.29 {\tiny \color{gray} [0.13, 0.57]} \\
 & Iridescent & 0.50 {\tiny \color{gray} [0.28, 0.75]} & 0.45 {\tiny \color{gray} [0.24, 0.68]} & 0.27 {\tiny \color{gray} [0.12, 0.44]} & 0.49 {\tiny \color{gray} [0.27, 0.67]} \\
 & Metallic & 0.45 {\tiny \color{gray} [0.21, 0.70]} & 0.46 {\tiny \color{gray} [0.23, 0.69]} & 0.38 {\tiny \color{gray} [0.16, 0.67]} & 0.40 {\tiny \color{gray} [0.22, 0.66]} \\
 & Multicolored & 0.50 {\tiny \color{gray} [0.25, 0.78]} & 0.45 {\tiny \color{gray} [0.21, 0.68]} & 0.39 {\tiny \color{gray} [0.17, 0.69]} & 0.44 {\tiny \color{gray} [0.22, 0.75]} \\
 & Paper/Material & 0.45 {\tiny \color{gray} [0.23, 0.71]} & 0.45 {\tiny \color{gray} [0.21, 0.74]} & 0.64 {\tiny \color{gray} [0.35, 0.86]} & 0.44 {\tiny \color{gray} [0.16, 0.65]} \\
 & Reflective & 0.50 {\tiny \color{gray} [0.24, 0.75]} & 0.52 {\tiny \color{gray} [0.23, 0.78]} & 0.44 {\tiny \color{gray} [0.21, 0.74]} & 0.42 {\tiny \color{gray} [0.21, 0.70]} \\
 & Textured/Dusty & 0.54 {\tiny \color{gray} [0.29, 0.80]} & 0.53 {\tiny \color{gray} [0.28, 0.79]} & 0.44 {\tiny \color{gray} [0.19, 0.63]} & 0.45 {\tiny \color{gray} [0.23, 0.78]} \\
 & Translucent & 0.57 {\tiny \color{gray} [0.33, 0.79]} & 0.44 {\tiny \color{gray} [0.21, 0.67]} & 0.29 {\tiny \color{gray} [0.16, 0.46]} & 0.34 {\tiny \color{gray} [0.16, 0.49]} \\
\cline{1-6}
\multirow[t]{10}{*}{Resonant} & Antique & 0.50 {\tiny \color{gray} [0.24, 0.74]} & 0.49 {\tiny \color{gray} [0.23, 0.72]} & 0.37 {\tiny \color{gray} [0.16, 0.60]} & 0.43 {\tiny \color{gray} [0.17, 0.72]} \\
 & Foldable & 0.56 {\tiny \color{gray} [0.28, 0.78]} & 0.51 {\tiny \color{gray} [0.28, 0.73]} & 0.52 {\tiny \color{gray} [0.22, 0.80]} & 0.48 {\tiny \color{gray} [0.22, 0.71]} \\
 & Hanging & 0.53 {\tiny \color{gray} [0.28, 0.75]} & 0.55 {\tiny \color{gray} [0.30, 0.76]} & 0.61 {\tiny \color{gray} [0.46, 0.82]} & 0.59 {\tiny \color{gray} [0.34, 0.81]} \\
 & Iridescent & 0.47 {\tiny \color{gray} [0.24, 0.72]} & 0.39 {\tiny \color{gray} [0.19, 0.64]} & 0.25 {\tiny \color{gray} [0.10, 0.52]} & 0.38 {\tiny \color{gray} [0.16, 0.66]} \\
 & Metallic & 0.50 {\tiny \color{gray} [0.28, 0.78]} & 0.51 {\tiny \color{gray} [0.25, 0.75]} & 0.57 {\tiny \color{gray} [0.34, 0.81]} & 0.45 {\tiny \color{gray} [0.29, 0.72]} \\
 & Multicolored & 0.47 {\tiny \color{gray} [0.23, 0.75]} & 0.55 {\tiny \color{gray} [0.29, 0.83]} & 0.40 {\tiny \color{gray} [0.17, 0.80]} & 0.58 {\tiny \color{gray} [0.29, 0.75]} \\
 & Paper/Material & 0.52 {\tiny \color{gray} [0.28, 0.74]} & 0.58 {\tiny \color{gray} [0.32, 0.82]} & 0.60 {\tiny \color{gray} [0.35, 0.83]} & 0.52 {\tiny \color{gray} [0.26, 0.76]} \\
 & Reflective & 0.50 {\tiny \color{gray} [0.25, 0.75]} & 0.51 {\tiny \color{gray} [0.23, 0.77]} & 0.48 {\tiny \color{gray} [0.35, 0.75]} & 0.60 {\tiny \color{gray} [0.32, 0.73]} \\
 & Textured/Dusty & 0.46 {\tiny \color{gray} [0.20, 0.74]} & 0.36 {\tiny \color{gray} [0.16, 0.61]} & 0.16 {\tiny \color{gray} [0.10, 0.38]} & 0.26 {\tiny \color{gray} [0.13, 0.51]} \\
 & Translucent & 0.49 {\tiny \color{gray} [0.24, 0.71]} & 0.34 {\tiny \color{gray} [0.16, 0.58]} & 0.26 {\tiny \color{gray} [0.11, 0.46]} & 0.35 {\tiny \color{gray} [0.14, 0.62]} \\
\cline{1-6}
\multirow[t]{10}{*}{Rough} & Antique & 0.60 {\tiny \color{gray} [0.36, 0.84]} & 0.47 {\tiny \color{gray} [0.27, 0.75]} & 0.49 {\tiny \color{gray} [0.30, 0.82]} & 0.80 {\tiny \color{gray} [0.41, 0.89]} \\
 & Foldable & 0.46 {\tiny \color{gray} [0.23, 0.77]} & 0.49 {\tiny \color{gray} [0.30, 0.78]} & 0.66 {\tiny \color{gray} [0.32, 0.84]} & 0.35 {\tiny \color{gray} [0.14, 0.70]} \\
 & Hanging & 0.48 {\tiny \color{gray} [0.24, 0.75]} & 0.36 {\tiny \color{gray} [0.24, 0.72]} & 0.69 {\tiny \color{gray} [0.55, 0.91]} & 0.74 {\tiny \color{gray} [0.24, 0.83]} \\
 & Iridescent & 0.58 {\tiny \color{gray} [0.32, 0.80]} & 0.38 {\tiny \color{gray} [0.20, 0.60]} & 0.60 {\tiny \color{gray} [0.18, 0.83]} & 0.53 {\tiny \color{gray} [0.35, 0.74]} \\
 & Metallic & 0.60 {\tiny \color{gray} [0.38, 0.84]} & 0.70 {\tiny \color{gray} [0.46, 0.85]} & 0.76 {\tiny \color{gray} [0.48, 0.88]} & 0.63 {\tiny \color{gray} [0.42, 0.78]} \\
 & Multicolored & 0.58 {\tiny \color{gray} [0.32, 0.83]} & 0.57 {\tiny \color{gray} [0.32, 0.83]} & 0.72 {\tiny \color{gray} [0.55, 0.94]} & 0.57 {\tiny \color{gray} [0.32, 0.90]} \\
 & Paper/Material & 0.57 {\tiny \color{gray} [0.28, 0.79]} & 0.66 {\tiny \color{gray} [0.20, 0.86]} & 0.85 {\tiny \color{gray} [0.59, 0.95]} & 0.65 {\tiny \color{gray} [0.43, 0.84]} \\
 & Reflective & 0.60 {\tiny \color{gray} [0.31, 0.79]} & 0.47 {\tiny \color{gray} [0.22, 0.77]} & 0.68 {\tiny \color{gray} [0.30, 0.81]} & 0.64 {\tiny \color{gray} [0.47, 0.85]} \\
 & Textured/Dusty & 0.53 {\tiny \color{gray} [0.26, 0.77]} & 0.40 {\tiny \color{gray} [0.22, 0.60]} & 0.28 {\tiny \color{gray} [0.08, 0.44]} & 0.47 {\tiny \color{gray} [0.23, 0.58]} \\
 & Translucent & 0.44 {\tiny \color{gray} [0.20, 0.65]} & 0.29 {\tiny \color{gray} [0.14, 0.53]} & 0.35 {\tiny \color{gray} [0.16, 0.42]} & 0.31 {\tiny \color{gray} [0.13, 0.54]} \\
\cline{1-6}
\multirow[t]{10}{*}{Sittable} & Antique & 0.44 {\tiny \color{gray} [0.17, 0.74]} & 0.53 {\tiny \color{gray} [0.30, 0.77]} & 0.52 {\tiny \color{gray} [0.28, 0.76]} & 0.43 {\tiny \color{gray} [0.21, 0.74]} \\
 & Foldable & 0.50 {\tiny \color{gray} [0.21, 0.72]} & 0.45 {\tiny \color{gray} [0.25, 0.71]} & 0.57 {\tiny \color{gray} [0.25, 0.75]} & 0.68 {\tiny \color{gray} [0.43, 0.84]} \\
 & Hanging & 0.43 {\tiny \color{gray} [0.19, 0.73]} & 0.54 {\tiny \color{gray} [0.26, 0.78]} & 0.53 {\tiny \color{gray} [0.25, 0.81]} & 0.73 {\tiny \color{gray} [0.44, 0.87]} \\
 & Iridescent & 0.48 {\tiny \color{gray} [0.32, 0.74]} & 0.50 {\tiny \color{gray} [0.24, 0.73]} & 0.55 {\tiny \color{gray} [0.25, 0.79]} & 0.40 {\tiny \color{gray} [0.20, 0.64]} \\
 & Metallic & 0.63 {\tiny \color{gray} [0.33, 0.76]} & 0.52 {\tiny \color{gray} [0.29, 0.74]} & 0.54 {\tiny \color{gray} [0.26, 0.70]} & 0.55 {\tiny \color{gray} [0.28, 0.74]} \\
 & Multicolored & 0.50 {\tiny \color{gray} [0.35, 0.73]} & 0.52 {\tiny \color{gray} [0.29, 0.75]} & 0.61 {\tiny \color{gray} [0.44, 0.86]} & 0.74 {\tiny \color{gray} [0.45, 0.88]} \\
 & Paper/Material & 0.49 {\tiny \color{gray} [0.26, 0.66]} & 0.53 {\tiny \color{gray} [0.27, 0.81]} & 0.39 {\tiny \color{gray} [0.17, 0.72]} & 0.54 {\tiny \color{gray} [0.21, 0.83]} \\
 & Reflective & 0.52 {\tiny \color{gray} [0.26, 0.68]} & 0.53 {\tiny \color{gray} [0.26, 0.77]} & 0.58 {\tiny \color{gray} [0.36, 0.81]} & 0.58 {\tiny \color{gray} [0.36, 0.75]} \\
 & Textured/Dusty & 0.57 {\tiny \color{gray} [0.25, 0.71]} & 0.51 {\tiny \color{gray} [0.26, 0.74]} & 0.62 {\tiny \color{gray} [0.43, 0.82]} & 0.56 {\tiny \color{gray} [0.30, 0.74]} \\
 & Translucent & 0.64 {\tiny \color{gray} [0.31, 0.86]} & 0.45 {\tiny \color{gray} [0.25, 0.70]} & 0.48 {\tiny \color{gray} [0.26, 0.72]} & 0.70 {\tiny \color{gray} [0.41, 0.84]} \\
\cline{1-6}
\multirow[t]{10}{*}{Squeezable} & Antique & 0.46 {\tiny \color{gray} [0.17, 0.73]} & 0.48 {\tiny \color{gray} [0.23, 0.73]} & 0.55 {\tiny \color{gray} [0.33, 0.90]} & 0.61 {\tiny \color{gray} [0.31, 0.78]} \\
 & Foldable & 0.40 {\tiny \color{gray} [0.17, 0.68]} & 0.38 {\tiny \color{gray} [0.17, 0.62]} & 0.39 {\tiny \color{gray} [0.24, 0.59]} & 0.34 {\tiny \color{gray} [0.17, 0.60]} \\
 & Hanging & 0.57 {\tiny \color{gray} [0.27, 0.76]} & 0.54 {\tiny \color{gray} [0.28, 0.82]} & 0.39 {\tiny \color{gray} [0.25, 0.74]} & 0.70 {\tiny \color{gray} [0.40, 0.89]} \\
 & Iridescent & 0.61 {\tiny \color{gray} [0.34, 0.85]} & 0.54 {\tiny \color{gray} [0.29, 0.80]} & 0.57 {\tiny \color{gray} [0.39, 0.92]} & 0.62 {\tiny \color{gray} [0.40, 0.88]} \\
 & Metallic & 0.45 {\tiny \color{gray} [0.23, 0.68]} & 0.45 {\tiny \color{gray} [0.23, 0.70]} & 0.38 {\tiny \color{gray} [0.20, 0.60]} & 0.35 {\tiny \color{gray} [0.16, 0.59]} \\
 & Multicolored & 0.56 {\tiny \color{gray} [0.25, 0.78]} & 0.50 {\tiny \color{gray} [0.28, 0.76]} & 0.52 {\tiny \color{gray} [0.32, 0.68]} & 0.66 {\tiny \color{gray} [0.32, 0.89]} \\
 & Paper/Material & 0.41 {\tiny \color{gray} [0.24, 0.67]} & 0.48 {\tiny \color{gray} [0.26, 0.75]} & 0.51 {\tiny \color{gray} [0.14, 0.75]} & 0.40 {\tiny \color{gray} [0.21, 0.73]} \\
 & Reflective & 0.44 {\tiny \color{gray} [0.18, 0.72]} & 0.44 {\tiny \color{gray} [0.25, 0.74]} & 0.46 {\tiny \color{gray} [0.31, 0.79]} & 0.47 {\tiny \color{gray} [0.20, 0.75]} \\
 & Textured/Dusty & 0.55 {\tiny \color{gray} [0.35, 0.77]} & 0.51 {\tiny \color{gray} [0.27, 0.73]} & 0.63 {\tiny \color{gray} [0.40, 0.83]} & 0.53 {\tiny \color{gray} [0.31, 0.79]} \\
 & Translucent & 0.62 {\tiny \color{gray} [0.38, 0.87]} & 0.52 {\tiny \color{gray} [0.25, 0.74]} & 0.74 {\tiny \color{gray} [0.42, 0.86]} & 0.67 {\tiny \color{gray} [0.37, 0.82]} \\
\cline{1-6}
\multirow[t]{10}{*}{Throwable} & Antique & 0.48 {\tiny \color{gray} [0.22, 0.77]} & 0.52 {\tiny \color{gray} [0.22, 0.76]} & 0.47 {\tiny \color{gray} [0.26, 0.70]} & 0.38 {\tiny \color{gray} [0.23, 0.71]} \\
 & Foldable & 0.37 {\tiny \color{gray} [0.16, 0.61]} & 0.51 {\tiny \color{gray} [0.23, 0.73]} & 0.32 {\tiny \color{gray} [0.12, 0.54]} & 0.32 {\tiny \color{gray} [0.17, 0.42]} \\
 & Hanging & 0.51 {\tiny \color{gray} [0.26, 0.74]} & 0.50 {\tiny \color{gray} [0.27, 0.74]} & 0.44 {\tiny \color{gray} [0.20, 0.81]} & 0.49 {\tiny \color{gray} [0.17, 0.82]} \\
 & Iridescent & 0.46 {\tiny \color{gray} [0.23, 0.75]} & 0.45 {\tiny \color{gray} [0.21, 0.68]} & 0.32 {\tiny \color{gray} [0.14, 0.60]} & 0.48 {\tiny \color{gray} [0.17, 0.61]} \\
 & Metallic & 0.41 {\tiny \color{gray} [0.21, 0.69]} & 0.46 {\tiny \color{gray} [0.21, 0.68]} & 0.36 {\tiny \color{gray} [0.12, 0.53]} & 0.38 {\tiny \color{gray} [0.17, 0.65]} \\
 & Multicolored & 0.53 {\tiny \color{gray} [0.26, 0.74]} & 0.48 {\tiny \color{gray} [0.23, 0.76]} & 0.54 {\tiny \color{gray} [0.30, 0.83]} & 0.57 {\tiny \color{gray} [0.27, 0.86]} \\
 & Paper/Material & 0.42 {\tiny \color{gray} [0.17, 0.72]} & 0.48 {\tiny \color{gray} [0.24, 0.70]} & 0.60 {\tiny \color{gray} [0.35, 0.76]} & 0.66 {\tiny \color{gray} [0.37, 0.90]} \\
 & Reflective & 0.48 {\tiny \color{gray} [0.27, 0.73]} & 0.48 {\tiny \color{gray} [0.22, 0.74]} & 0.50 {\tiny \color{gray} [0.26, 0.78]} & 0.66 {\tiny \color{gray} [0.38, 0.88]} \\
 & Textured/Dusty & 0.49 {\tiny \color{gray} [0.25, 0.77]} & 0.51 {\tiny \color{gray} [0.27, 0.74]} & 0.40 {\tiny \color{gray} [0.20, 0.61]} & 0.40 {\tiny \color{gray} [0.18, 0.64]} \\
 & Translucent & 0.64 {\tiny \color{gray} [0.31, 0.84]} & 0.43 {\tiny \color{gray} [0.19, 0.70]} & 0.29 {\tiny \color{gray} [0.11, 0.50]} & 0.51 {\tiny \color{gray} [0.18, 0.86]} \\
\cline{1-6}
\end{longtable}

\begin{longtable}{p{1.4cm}p{2cm}rrrr}
\caption{Posterior medians with 50\% credible intervals for Art.} \label{tab:posteriors_art} \\
\toprule
Target & Parameter & Claude Sonnet 4.6 & GPT-5.4 & Gemini 3 Flash & Qwen3-VL 235B \\
\midrule
\endfirsthead
\caption[]{Posterior medians with 50\% credible intervals for Art.} \\
\toprule
 & agent & Claude Sonnet 4.6 & GPT-5.4 & Gemini 3 Flash & Qwen3-VL 235B \\
target & parameter &  &  &  &  \\
\midrule
\endhead
\midrule
\multicolumn{6}{r}{Continued on next page} \\
\midrule
\endfoot
\bottomrule
\endlastfoot
\multirow[t]{10}{*}{Amateur} & Art Deco/Decorative & 0.50 {\tiny \color{gray} [0.23, 0.78]} & 0.45 {\tiny \color{gray} [0.21, 0.74]} & 0.75 {\tiny \color{gray} [0.25, 0.94]} & 0.56 {\tiny \color{gray} [0.24, 0.88]} \\
 & Atmospheric Perspective & 0.51 {\tiny \color{gray} [0.28, 0.79]} & 0.59 {\tiny \color{gray} [0.30, 0.84]} & 0.59 {\tiny \color{gray} [0.16, 0.84]} & 0.58 {\tiny \color{gray} [0.32, 0.83]} \\
 & Engraving/Etching & 0.49 {\tiny \color{gray} [0.21, 0.79]} & 0.42 {\tiny \color{gray} [0.20, 0.71]} & 0.44 {\tiny \color{gray} [0.22, 0.72]} & 0.48 {\tiny \color{gray} [0.23, 0.73]} \\
 & Faded Surface & 0.47 {\tiny \color{gray} [0.21, 0.72]} & 0.51 {\tiny \color{gray} [0.26, 0.77]} & 0.61 {\tiny \color{gray} [0.22, 0.86]} & 0.55 {\tiny \color{gray} [0.20, 0.71]} \\
 & Floral/Warm Palette & 0.54 {\tiny \color{gray} [0.28, 0.78]} & 0.63 {\tiny \color{gray} [0.34, 0.86]} & 0.60 {\tiny \color{gray} [0.23, 0.90]} & 0.54 {\tiny \color{gray} [0.28, 0.74]} \\
 & Low Key/Dark & 0.47 {\tiny \color{gray} [0.25, 0.70]} & 0.41 {\tiny \color{gray} [0.22, 0.73]} & 0.49 {\tiny \color{gray} [0.14, 0.69]} & 0.54 {\tiny \color{gray} [0.25, 0.79]} \\
 & Ornate Detail & 0.47 {\tiny \color{gray} [0.25, 0.70]} & 0.45 {\tiny \color{gray} [0.23, 0.70]} & 0.40 {\tiny \color{gray} [0.17, 0.68]} & 0.51 {\tiny \color{gray} [0.23, 0.81]} \\
 & Sfumato & 0.48 {\tiny \color{gray} [0.23, 0.71]} & 0.54 {\tiny \color{gray} [0.24, 0.75]} & 0.21 {\tiny \color{gray} [0.13, 0.37]} & 0.67 {\tiny \color{gray} [0.30, 0.88]} \\
 & Symbolism & 0.47 {\tiny \color{gray} [0.22, 0.77]} & 0.56 {\tiny \color{gray} [0.26, 0.78]} & 0.53 {\tiny \color{gray} [0.22, 0.75]} & 0.48 {\tiny \color{gray} [0.19, 0.71]} \\
 & Thin Wash & 0.57 {\tiny \color{gray} [0.25, 0.81]} & 0.50 {\tiny \color{gray} [0.29, 0.75]} & 0.30 {\tiny \color{gray} [0.20, 0.74]} & 0.49 {\tiny \color{gray} [0.25, 0.84]} \\
\cline{1-6}
\multirow[t]{10}{*}{Beautiful} & Art Deco/Decorative & 0.48 {\tiny \color{gray} [0.24, 0.70]} & 0.38 {\tiny \color{gray} [0.24, 0.63]} & 0.27 {\tiny \color{gray} [0.10, 0.52]} & 0.34 {\tiny \color{gray} [0.16, 0.59]} \\
 & Atmospheric Perspective & 0.59 {\tiny \color{gray} [0.28, 0.75]} & 0.39 {\tiny \color{gray} [0.21, 0.64]} & 0.42 {\tiny \color{gray} [0.31, 0.65]} & 0.53 {\tiny \color{gray} [0.19, 0.68]} \\
 & Engraving/Etching & 0.53 {\tiny \color{gray} [0.23, 0.76]} & 0.39 {\tiny \color{gray} [0.20, 0.74]} & 0.37 {\tiny \color{gray} [0.22, 0.53]} & 0.46 {\tiny \color{gray} [0.26, 0.74]} \\
 & Faded Surface & 0.48 {\tiny \color{gray} [0.14, 0.76]} & 0.38 {\tiny \color{gray} [0.19, 0.58]} & 0.42 {\tiny \color{gray} [0.10, 0.78]} & 0.45 {\tiny \color{gray} [0.32, 0.76]} \\
 & Floral/Warm Palette & 0.55 {\tiny \color{gray} [0.37, 0.82]} & 0.39 {\tiny \color{gray} [0.18, 0.63]} & 0.32 {\tiny \color{gray} [0.13, 0.57]} & 0.22 {\tiny \color{gray} [0.09, 0.45]} \\
 & Low Key/Dark & 0.48 {\tiny \color{gray} [0.32, 0.76]} & 0.32 {\tiny \color{gray} [0.17, 0.66]} & 0.23 {\tiny \color{gray} [0.06, 0.41]} & 0.17 {\tiny \color{gray} [0.07, 0.46]} \\
 & Ornate Detail & 0.47 {\tiny \color{gray} [0.31, 0.63]} & 0.58 {\tiny \color{gray} [0.30, 0.79]} & 0.72 {\tiny \color{gray} [0.46, 0.79]} & 0.63 {\tiny \color{gray} [0.40, 0.72]} \\
 & Sfumato & 0.42 {\tiny \color{gray} [0.34, 0.72]} & 0.28 {\tiny \color{gray} [0.14, 0.52]} & 0.28 {\tiny \color{gray} [0.21, 0.41]} & 0.26 {\tiny \color{gray} [0.13, 0.61]} \\
 & Symbolism & 0.43 {\tiny \color{gray} [0.27, 0.70]} & 0.40 {\tiny \color{gray} [0.21, 0.70]} & 0.31 {\tiny \color{gray} [0.17, 0.56]} & 0.42 {\tiny \color{gray} [0.19, 0.68]} \\
 & Thin Wash & 0.51 {\tiny \color{gray} [0.42, 0.69]} & 0.45 {\tiny \color{gray} [0.28, 0.67]} & 0.55 {\tiny \color{gray} [0.39, 0.72]} & 0.60 {\tiny \color{gray} [0.29, 0.76]} \\
\cline{1-6}
\multirow[t]{10}{*}{Cheap} & Art Deco/Decorative & 0.55 {\tiny \color{gray} [0.26, 0.73]} & 0.42 {\tiny \color{gray} [0.23, 0.72]} & 0.34 {\tiny \color{gray} [0.19, 0.71]} & 0.56 {\tiny \color{gray} [0.29, 0.79]} \\
 & Atmospheric Perspective & 0.61 {\tiny \color{gray} [0.36, 0.84]} & 0.52 {\tiny \color{gray} [0.23, 0.79]} & 0.28 {\tiny \color{gray} [0.11, 0.74]} & 0.67 {\tiny \color{gray} [0.22, 0.88]} \\
 & Engraving/Etching & 0.50 {\tiny \color{gray} [0.23, 0.72]} & 0.45 {\tiny \color{gray} [0.22, 0.71]} & 0.40 {\tiny \color{gray} [0.22, 0.73]} & 0.28 {\tiny \color{gray} [0.13, 0.75]} \\
 & Faded Surface & 0.54 {\tiny \color{gray} [0.30, 0.81]} & 0.53 {\tiny \color{gray} [0.32, 0.78]} & 0.39 {\tiny \color{gray} [0.14, 0.68]} & 0.46 {\tiny \color{gray} [0.18, 0.85]} \\
 & Floral/Warm Palette & 0.50 {\tiny \color{gray} [0.29, 0.74]} & 0.48 {\tiny \color{gray} [0.23, 0.75]} & 0.60 {\tiny \color{gray} [0.31, 0.87]} & 0.72 {\tiny \color{gray} [0.45, 0.91]} \\
 & Low Key/Dark & 0.39 {\tiny \color{gray} [0.23, 0.70]} & 0.51 {\tiny \color{gray} [0.23, 0.77]} & 0.32 {\tiny \color{gray} [0.11, 0.76]} & 0.57 {\tiny \color{gray} [0.16, 0.75]} \\
 & Ornate Detail & 0.49 {\tiny \color{gray} [0.28, 0.75]} & 0.47 {\tiny \color{gray} [0.23, 0.71]} & 0.58 {\tiny \color{gray} [0.41, 0.86]} & 0.62 {\tiny \color{gray} [0.24, 0.88]} \\
 & Sfumato & 0.44 {\tiny \color{gray} [0.24, 0.67]} & 0.53 {\tiny \color{gray} [0.29, 0.79]} & 0.47 {\tiny \color{gray} [0.18, 0.76]} & 0.50 {\tiny \color{gray} [0.15, 0.78]} \\
 & Symbolism & 0.45 {\tiny \color{gray} [0.25, 0.74]} & 0.52 {\tiny \color{gray} [0.28, 0.78]} & 0.47 {\tiny \color{gray} [0.14, 0.71]} & 0.47 {\tiny \color{gray} [0.14, 0.77]} \\
 & Thin Wash & 0.58 {\tiny \color{gray} [0.31, 0.84]} & 0.51 {\tiny \color{gray} [0.22, 0.77]} & 0.52 {\tiny \color{gray} [0.27, 0.74]} & 0.41 {\tiny \color{gray} [0.25, 0.77]} \\
\cline{1-6}
\multirow[t]{10}{*}{Expensive} & Art Deco/Decorative & 0.47 {\tiny \color{gray} [0.27, 0.75]} & 0.46 {\tiny \color{gray} [0.27, 0.65]} & 0.53 {\tiny \color{gray} [0.23, 0.79]} & 0.66 {\tiny \color{gray} [0.42, 0.82]} \\
 & Atmospheric Perspective & 0.49 {\tiny \color{gray} [0.22, 0.72]} & 0.33 {\tiny \color{gray} [0.16, 0.58]} & 0.30 {\tiny \color{gray} [0.15, 0.55]} & 0.36 {\tiny \color{gray} [0.13, 0.50]} \\
 & Engraving/Etching & 0.57 {\tiny \color{gray} [0.30, 0.76]} & 0.65 {\tiny \color{gray} [0.36, 0.84]} & 0.52 {\tiny \color{gray} [0.28, 0.75]} & 0.49 {\tiny \color{gray} [0.45, 0.78]} \\
 & Faded Surface & 0.53 {\tiny \color{gray} [0.35, 0.77]} & 0.54 {\tiny \color{gray} [0.34, 0.79]} & 0.33 {\tiny \color{gray} [0.15, 0.54]} & 0.43 {\tiny \color{gray} [0.15, 0.66]} \\
 & Floral/Warm Palette & 0.59 {\tiny \color{gray} [0.36, 0.81]} & 0.54 {\tiny \color{gray} [0.30, 0.76]} & 0.36 {\tiny \color{gray} [0.24, 0.62]} & 0.39 {\tiny \color{gray} [0.11, 0.54]} \\
 & Low Key/Dark & 0.48 {\tiny \color{gray} [0.32, 0.72]} & 0.54 {\tiny \color{gray} [0.30, 0.77]} & 0.45 {\tiny \color{gray} [0.30, 0.84]} & 0.61 {\tiny \color{gray} [0.33, 0.70]} \\
 & Ornate Detail & 0.43 {\tiny \color{gray} [0.23, 0.74]} & 0.48 {\tiny \color{gray} [0.26, 0.72]} & 0.75 {\tiny \color{gray} [0.42, 0.84]} & 0.68 {\tiny \color{gray} [0.42, 0.92]} \\
 & Sfumato & 0.35 {\tiny \color{gray} [0.18, 0.60]} & 0.40 {\tiny \color{gray} [0.20, 0.63]} & 0.36 {\tiny \color{gray} [0.21, 0.49]} & 0.19 {\tiny \color{gray} [0.10, 0.41]} \\
 & Symbolism & 0.58 {\tiny \color{gray} [0.30, 0.83]} & 0.59 {\tiny \color{gray} [0.32, 0.79]} & 0.58 {\tiny \color{gray} [0.46, 0.85]} & 0.48 {\tiny \color{gray} [0.22, 0.89]} \\
 & Thin Wash & 0.45 {\tiny \color{gray} [0.22, 0.72]} & 0.47 {\tiny \color{gray} [0.25, 0.68]} & 0.40 {\tiny \color{gray} [0.28, 0.63]} & 0.47 {\tiny \color{gray} [0.20, 0.78]} \\
\cline{1-6}
\multirow[t]{10}{*}{Experimental} & Art Deco/Decorative & 0.54 {\tiny \color{gray} [0.33, 0.79]} & 0.51 {\tiny \color{gray} [0.27, 0.73]} & 0.61 {\tiny \color{gray} [0.22, 0.78]} & 0.33 {\tiny \color{gray} [0.11, 0.59]} \\
 & Atmospheric Perspective & 0.58 {\tiny \color{gray} [0.30, 0.81]} & 0.51 {\tiny \color{gray} [0.29, 0.71]} & 0.36 {\tiny \color{gray} [0.19, 0.59]} & 0.52 {\tiny \color{gray} [0.29, 0.71]} \\
 & Engraving/Etching & 0.48 {\tiny \color{gray} [0.22, 0.75]} & 0.42 {\tiny \color{gray} [0.18, 0.66]} & 0.62 {\tiny \color{gray} [0.33, 0.87]} & 0.44 {\tiny \color{gray} [0.26, 0.80]} \\
 & Faded Surface & 0.48 {\tiny \color{gray} [0.23, 0.73]} & 0.37 {\tiny \color{gray} [0.19, 0.66]} & 0.29 {\tiny \color{gray} [0.06, 0.59]} & 0.37 {\tiny \color{gray} [0.17, 0.50]} \\
 & Floral/Warm Palette & 0.42 {\tiny \color{gray} [0.20, 0.71]} & 0.49 {\tiny \color{gray} [0.23, 0.78]} & 0.39 {\tiny \color{gray} [0.20, 0.78]} & 0.62 {\tiny \color{gray} [0.39, 0.79]} \\
 & Low Key/Dark & 0.49 {\tiny \color{gray} [0.27, 0.74]} & 0.49 {\tiny \color{gray} [0.26, 0.68]} & 0.42 {\tiny \color{gray} [0.25, 0.71]} & 0.35 {\tiny \color{gray} [0.26, 0.79]} \\
 & Ornate Detail & 0.53 {\tiny \color{gray} [0.22, 0.77]} & 0.51 {\tiny \color{gray} [0.24, 0.73]} & 0.51 {\tiny \color{gray} [0.20, 0.80]} & 0.46 {\tiny \color{gray} [0.27, 0.79]} \\
 & Sfumato & 0.56 {\tiny \color{gray} [0.36, 0.79]} & 0.48 {\tiny \color{gray} [0.25, 0.74]} & 0.34 {\tiny \color{gray} [0.19, 0.52]} & 0.63 {\tiny \color{gray} [0.35, 0.80]} \\
 & Symbolism & 0.50 {\tiny \color{gray} [0.23, 0.75]} & 0.58 {\tiny \color{gray} [0.33, 0.80]} & 0.63 {\tiny \color{gray} [0.38, 0.81]} & 0.61 {\tiny \color{gray} [0.43, 0.74]} \\
 & Thin Wash & 0.48 {\tiny \color{gray} [0.21, 0.76]} & 0.49 {\tiny \color{gray} [0.25, 0.75]} & 0.33 {\tiny \color{gray} [0.24, 0.52]} & 0.38 {\tiny \color{gray} [0.16, 0.74]} \\
\cline{1-6}
\end{longtable}

\begin{longtable}{p{1.4cm}p{2cm}rrrr}
\caption{Posterior medians with 50\% credible intervals for Authenticity.} \label{tab:posteriors_authenticity} \\
\toprule
Target & Parameter & Claude Sonnet 4.6 & GPT-5.4 & Gemini 3 Flash & Qwen3-VL 235B \\
\midrule
\endfirsthead
\caption[]{Posterior medians with 50\% credible intervals for Authenticity.} \\
\toprule
 & agent & Claude Sonnet 4.6 & GPT-5.4 & Gemini 3 Flash & Qwen3-VL 235B \\
target & parameter &  &  &  &  \\
\midrule
\endhead
\midrule
\multicolumn{6}{r}{Continued on next page} \\
\midrule
\endfoot
\bottomrule
\endlastfoot
\multirow[t]{10}{*}{Authentic} & Black And White & 0.37 {\tiny \color{gray} [0.18, 0.65]} & 0.42 {\tiny \color{gray} [0.23, 0.69]} & 0.34 {\tiny \color{gray} [0.12, 0.60]} & 0.36 {\tiny \color{gray} [0.20, 0.57]} \\
 & Blending Errors & 0.52 {\tiny \color{gray} [0.22, 0.76]} & 0.53 {\tiny \color{gray} [0.23, 0.75]} & 0.51 {\tiny \color{gray} [0.21, 0.65]} & 0.55 {\tiny \color{gray} [0.28, 0.73]} \\
 & Crushed Blacks & 0.49 {\tiny \color{gray} [0.22, 0.71]} & 0.46 {\tiny \color{gray} [0.20, 0.69]} & 0.48 {\tiny \color{gray} [0.22, 0.77]} & 0.30 {\tiny \color{gray} [0.13, 0.47]} \\
 & Digital Artifacts & 0.48 {\tiny \color{gray} [0.23, 0.71]} & 0.46 {\tiny \color{gray} [0.21, 0.72]} & 0.35 {\tiny \color{gray} [0.19, 0.58]} & 0.38 {\tiny \color{gray} [0.18, 0.63]} \\
 & Film Grain & 0.52 {\tiny \color{gray} [0.28, 0.75]} & 0.50 {\tiny \color{gray} [0.26, 0.73]} & 0.41 {\tiny \color{gray} [0.17, 0.67]} & 0.53 {\tiny \color{gray} [0.24, 0.71]} \\
 & Natural Edit & 0.54 {\tiny \color{gray} [0.30, 0.77]} & 0.49 {\tiny \color{gray} [0.26, 0.74]} & 0.55 {\tiny \color{gray} [0.29, 0.72]} & 0.57 {\tiny \color{gray} [0.28, 0.80]} \\
 & Scratched Film & 0.53 {\tiny \color{gray} [0.28, 0.77]} & 0.53 {\tiny \color{gray} [0.29, 0.77]} & 0.50 {\tiny \color{gray} [0.32, 0.71]} & 0.52 {\tiny \color{gray} [0.26, 0.75]} \\
 & Sepia & 0.51 {\tiny \color{gray} [0.29, 0.75]} & 0.53 {\tiny \color{gray} [0.28, 0.75]} & 0.49 {\tiny \color{gray} [0.25, 0.74]} & 0.57 {\tiny \color{gray} [0.26, 0.80]} \\
 & Upscaling Noise & 0.50 {\tiny \color{gray} [0.25, 0.75]} & 0.48 {\tiny \color{gray} [0.24, 0.73]} & 0.41 {\tiny \color{gray} [0.15, 0.73]} & 0.51 {\tiny \color{gray} [0.23, 0.74]} \\
 & Vintage Look & 0.50 {\tiny \color{gray} [0.27, 0.76]} & 0.50 {\tiny \color{gray} [0.24, 0.74]} & 0.55 {\tiny \color{gray} [0.29, 0.77]} & 0.41 {\tiny \color{gray} [0.21, 0.64]} \\
\cline{1-6}
\multirow[t]{10}{*}{Manipulated} & Black And White & 0.87 {\tiny \color{gray} [0.56, 0.94]} & 0.51 {\tiny \color{gray} [0.25, 0.75]} & 0.72 {\tiny \color{gray} [0.27, 0.87]} & 0.48 {\tiny \color{gray} [0.35, 0.80]} \\
 & Blending Errors & 0.66 {\tiny \color{gray} [0.32, 0.88]} & 0.47 {\tiny \color{gray} [0.25, 0.74]} & 0.37 {\tiny \color{gray} [0.25, 0.70]} & 0.69 {\tiny \color{gray} [0.31, 0.85]} \\
 & Crushed Blacks & 0.71 {\tiny \color{gray} [0.33, 0.93]} & 0.55 {\tiny \color{gray} [0.30, 0.80]} & 0.69 {\tiny \color{gray} [0.53, 0.88]} & 0.54 {\tiny \color{gray} [0.29, 0.76]} \\
 & Digital Artifacts & 0.41 {\tiny \color{gray} [0.23, 0.68]} & 0.44 {\tiny \color{gray} [0.25, 0.74]} & 0.60 {\tiny \color{gray} [0.42, 0.81]} & 0.45 {\tiny \color{gray} [0.31, 0.70]} \\
 & Film Grain & 0.58 {\tiny \color{gray} [0.32, 0.81]} & 0.49 {\tiny \color{gray} [0.25, 0.75]} & 0.42 {\tiny \color{gray} [0.14, 0.79]} & 0.46 {\tiny \color{gray} [0.16, 0.71]} \\
 & Natural Edit & 0.61 {\tiny \color{gray} [0.28, 0.77]} & 0.52 {\tiny \color{gray} [0.32, 0.80]} & 0.51 {\tiny \color{gray} [0.40, 0.86]} & 0.52 {\tiny \color{gray} [0.33, 0.79]} \\
 & Scratched Film & 0.40 {\tiny \color{gray} [0.14, 0.70]} & 0.49 {\tiny \color{gray} [0.24, 0.76]} & 0.37 {\tiny \color{gray} [0.15, 0.73]} & 0.33 {\tiny \color{gray} [0.22, 0.51]} \\
 & Sepia & 0.46 {\tiny \color{gray} [0.21, 0.70]} & 0.51 {\tiny \color{gray} [0.28, 0.72]} & 0.25 {\tiny \color{gray} [0.10, 0.66]} & 0.38 {\tiny \color{gray} [0.20, 0.75]} \\
 & Upscaling Noise & 0.57 {\tiny \color{gray} [0.31, 0.74]} & 0.58 {\tiny \color{gray} [0.28, 0.80]} & 0.56 {\tiny \color{gray} [0.46, 0.80]} & 0.76 {\tiny \color{gray} [0.48, 0.90]} \\
 & Vintage Look & 0.62 {\tiny \color{gray} [0.43, 0.84]} & 0.49 {\tiny \color{gray} [0.25, 0.74]} & 0.49 {\tiny \color{gray} [0.24, 0.64]} & 0.51 {\tiny \color{gray} [0.31, 0.76]} \\
\cline{1-6}
\end{longtable}

\begin{longtable}{p{1.4cm}p{2cm}rrrr}
\caption{Posterior medians with 50\% credible intervals for Faces.} \label{tab:posteriors_faces} \\
\toprule
Target & Parameter & Claude Sonnet 4.6 & GPT-5.4 & Gemini 3 Flash & Qwen3-VL 235B \\
\midrule
\endfirsthead
\caption[]{Posterior medians with 50\% credible intervals for Faces.} \\
\toprule
Target & Parameter & Claude Sonnet 4.6 & GPT-5.4 & Gemini 3 Flash & Qwen3-VL 235B \\
\midrule
\endhead
\midrule
\multicolumn{6}{r}{Continued on next page} \\
\midrule
\endfoot
\bottomrule
\endlastfoot
\multirow[t]{10}{*}{Attractive} & Beard/Facial Hair & 0.44 {\tiny \color{gray} [0.13, 0.77]} & 0.40 {\tiny \color{gray} [0.19, 0.63]} & 0.29 {\tiny \color{gray} [0.12, 0.52]} & 0.39 {\tiny \color{gray} [0.22, 0.55]} \\
 & Blue Eyes & 0.47 {\tiny \color{gray} [0.16, 0.76]} & 0.61 {\tiny \color{gray} [0.31, 0.82]} & 0.45 {\tiny \color{gray} [0.21, 0.82]} & 0.75 {\tiny \color{gray} [0.19, 0.89]} \\
 & Child & 0.40 {\tiny \color{gray} [0.18, 0.73]} & 0.53 {\tiny \color{gray} [0.29, 0.72]} & 0.49 {\tiny \color{gray} [0.24, 0.72]} & 0.47 {\tiny \color{gray} [0.31, 0.58]} \\
 & Dark/Brown Skin & 0.55 {\tiny \color{gray} [0.26, 0.79]} & 0.48 {\tiny \color{gray} [0.22, 0.74]} & 0.57 {\tiny \color{gray} [0.27, 0.80]} & 0.59 {\tiny \color{gray} [0.40, 0.74]} \\
 & Earrings & 0.48 {\tiny \color{gray} [0.16, 0.81]} & 0.51 {\tiny \color{gray} [0.30, 0.78]} & 0.51 {\tiny \color{gray} [0.34, 0.75]} & 0.39 {\tiny \color{gray} [0.17, 0.71]} \\
 & East Asian & 0.35 {\tiny \color{gray} [0.18, 0.44]} & 0.25 {\tiny \color{gray} [0.08, 0.46]} & 0.16 {\tiny \color{gray} [0.08, 0.37]} & 0.34 {\tiny \color{gray} [0.15, 0.54]} \\
 & Eyeglasses & 0.44 {\tiny \color{gray} [0.24, 0.67]} & 0.29 {\tiny \color{gray} [0.15, 0.52]} & 0.46 {\tiny \color{gray} [0.29, 0.64]} & 0.36 {\tiny \color{gray} [0.19, 0.71]} \\
 & Feminine & 0.65 {\tiny \color{gray} [0.36, 0.80]} & 0.49 {\tiny \color{gray} [0.22, 0.74]} & 0.59 {\tiny \color{gray} [0.35, 0.84]} & 0.59 {\tiny \color{gray} [0.21, 0.91]} \\
 & Masculine/Receding Hairline & 0.30 {\tiny \color{gray} [0.12, 0.52]} & 0.38 {\tiny \color{gray} [0.14, 0.59]} & 0.18 {\tiny \color{gray} [0.08, 0.36]} & 0.37 {\tiny \color{gray} [0.26, 0.69]} \\
 & Smiling & 0.38 {\tiny \color{gray} [0.14, 0.68]} & 0.45 {\tiny \color{gray} [0.17, 0.69]} & 0.33 {\tiny \color{gray} [0.21, 0.62]} & 0.64 {\tiny \color{gray} [0.34, 0.81]} \\
\cline{1-6}
\multirow[t]{10}{*}{Criminal} & Beard/Facial Hair & 0.53 {\tiny \color{gray} [0.33, 0.81]} & 0.41 {\tiny \color{gray} [0.21, 0.58]} & 0.32 {\tiny \color{gray} [0.14, 0.66]} & 0.29 {\tiny \color{gray} [0.20, 0.49]} \\
 & Blue Eyes & 0.55 {\tiny \color{gray} [0.21, 0.83]} & 0.52 {\tiny \color{gray} [0.32, 0.77]} & 0.34 {\tiny \color{gray} [0.22, 0.55]} & 0.57 {\tiny \color{gray} [0.33, 0.81]} \\
 & Child & 0.58 {\tiny \color{gray} [0.22, 0.77]} & 0.50 {\tiny \color{gray} [0.27, 0.73]} & 0.44 {\tiny \color{gray} [0.15, 0.55]} & 0.36 {\tiny \color{gray} [0.24, 0.76]} \\
 & Dark/Brown Skin & 0.45 {\tiny \color{gray} [0.33, 0.68]} & 0.51 {\tiny \color{gray} [0.30, 0.79]} & 0.72 {\tiny \color{gray} [0.47, 0.91]} & 0.64 {\tiny \color{gray} [0.20, 0.84]} \\
 & Earrings & 0.52 {\tiny \color{gray} [0.19, 0.79]} & 0.59 {\tiny \color{gray} [0.31, 0.82]} & 0.58 {\tiny \color{gray} [0.23, 0.88]} & 0.34 {\tiny \color{gray} [0.15, 0.73]} \\
 & East Asian & 0.55 {\tiny \color{gray} [0.28, 0.82]} & 0.39 {\tiny \color{gray} [0.15, 0.65]} & 0.23 {\tiny \color{gray} [0.08, 0.52]} & 0.55 {\tiny \color{gray} [0.24, 0.83]} \\
 & Eyeglasses & 0.48 {\tiny \color{gray} [0.19, 0.77]} & 0.62 {\tiny \color{gray} [0.33, 0.83]} & 0.43 {\tiny \color{gray} [0.16, 0.80]} & 0.63 {\tiny \color{gray} [0.45, 0.86]} \\
 & Feminine & 0.41 {\tiny \color{gray} [0.21, 0.69]} & 0.35 {\tiny \color{gray} [0.15, 0.57]} & 0.30 {\tiny \color{gray} [0.13, 0.44]} & 0.23 {\tiny \color{gray} [0.10, 0.39]} \\
 & Masculine/Receding Hairline & 0.47 {\tiny \color{gray} [0.31, 0.75]} & 0.63 {\tiny \color{gray} [0.33, 0.79]} & 0.87 {\tiny \color{gray} [0.47, 0.93]} & 0.62 {\tiny \color{gray} [0.44, 0.89]} \\
 & Smiling & 0.40 {\tiny \color{gray} [0.27, 0.67]} & 0.46 {\tiny \color{gray} [0.23, 0.71]} & 0.23 {\tiny \color{gray} [0.06, 0.49]} & 0.40 {\tiny \color{gray} [0.20, 0.69]} \\
\cline{1-6}
\multirow[t]{10}{*}{Fun} & Beard/Facial Hair & 0.46 {\tiny \color{gray} [0.20, 0.70]} & 0.36 {\tiny \color{gray} [0.20, 0.53]} & 0.35 {\tiny \color{gray} [0.22, 0.68]} & 0.38 {\tiny \color{gray} [0.21, 0.72]} \\
 & Blue Eyes & 0.62 {\tiny \color{gray} [0.31, 0.78]} & 0.50 {\tiny \color{gray} [0.20, 0.71]} & 0.54 {\tiny \color{gray} [0.28, 0.71]} & 0.41 {\tiny \color{gray} [0.21, 0.75]} \\
 & Child & 0.49 {\tiny \color{gray} [0.29, 0.63]} & 0.55 {\tiny \color{gray} [0.30, 0.80]} & 0.43 {\tiny \color{gray} [0.13, 0.60]} & 0.62 {\tiny \color{gray} [0.33, 0.69]} \\
 & Dark/Brown Skin & 0.59 {\tiny \color{gray} [0.24, 0.75]} & 0.58 {\tiny \color{gray} [0.29, 0.81]} & 0.56 {\tiny \color{gray} [0.37, 0.72]} & 0.56 {\tiny \color{gray} [0.22, 0.69]} \\
 & Earrings & 0.56 {\tiny \color{gray} [0.24, 0.82]} & 0.74 {\tiny \color{gray} [0.44, 0.84]} & 0.61 {\tiny \color{gray} [0.39, 0.82]} & 0.50 {\tiny \color{gray} [0.31, 0.74]} \\
 & East Asian & 0.66 {\tiny \color{gray} [0.39, 0.87]} & 0.67 {\tiny \color{gray} [0.43, 0.84]} & 0.41 {\tiny \color{gray} [0.22, 0.74]} & 0.37 {\tiny \color{gray} [0.21, 0.79]} \\
 & Eyeglasses & 0.80 {\tiny \color{gray} [0.62, 0.89]} & 0.49 {\tiny \color{gray} [0.25, 0.71]} & 0.71 {\tiny \color{gray} [0.23, 0.88]} & 0.45 {\tiny \color{gray} [0.12, 0.83]} \\
 & Feminine & 0.49 {\tiny \color{gray} [0.14, 0.63]} & 0.62 {\tiny \color{gray} [0.30, 0.72]} & 0.76 {\tiny \color{gray} [0.42, 0.95]} & 0.48 {\tiny \color{gray} [0.22, 0.80]} \\
 & Masculine/Receding Hairline & 0.54 {\tiny \color{gray} [0.35, 0.71]} & 0.37 {\tiny \color{gray} [0.22, 0.58]} & 0.53 {\tiny \color{gray} [0.32, 0.69]} & 0.38 {\tiny \color{gray} [0.30, 0.80]} \\
 & Smiling & 0.82 {\tiny \color{gray} [0.68, 0.95]} & 0.82 {\tiny \color{gray} [0.47, 0.95]} & 0.85 {\tiny \color{gray} [0.71, 0.95]} & 0.87 {\tiny \color{gray} [0.61, 0.92]} \\
\cline{1-6}
\multirow[t]{10}{*}{Hardworking} & Beard/Facial Hair & 0.41 {\tiny \color{gray} [0.22, 0.65]} & 0.40 {\tiny \color{gray} [0.22, 0.66]} & 0.39 {\tiny \color{gray} [0.17, 0.48]} & 0.48 {\tiny \color{gray} [0.25, 0.73]} \\
 & Blue Eyes & 0.38 {\tiny \color{gray} [0.19, 0.66]} & 0.48 {\tiny \color{gray} [0.23, 0.68]} & 0.49 {\tiny \color{gray} [0.29, 0.62]} & 0.36 {\tiny \color{gray} [0.18, 0.62]} \\
 & Child & 0.47 {\tiny \color{gray} [0.18, 0.76]} & 0.35 {\tiny \color{gray} [0.11, 0.69]} & 0.25 {\tiny \color{gray} [0.12, 0.56]} & 0.44 {\tiny \color{gray} [0.22, 0.70]} \\
 & Dark/Brown Skin & 0.60 {\tiny \color{gray} [0.30, 0.81]} & 0.64 {\tiny \color{gray} [0.35, 0.81]} & 0.65 {\tiny \color{gray} [0.40, 0.89]} & 0.55 {\tiny \color{gray} [0.27, 0.81]} \\
 & Earrings & 0.48 {\tiny \color{gray} [0.23, 0.69]} & 0.43 {\tiny \color{gray} [0.18, 0.64]} & 0.46 {\tiny \color{gray} [0.28, 0.76]} & 0.43 {\tiny \color{gray} [0.29, 0.65]} \\
 & East Asian & 0.43 {\tiny \color{gray} [0.18, 0.71]} & 0.32 {\tiny \color{gray} [0.15, 0.60]} & 0.18 {\tiny \color{gray} [0.10, 0.43]} & 0.37 {\tiny \color{gray} [0.17, 0.63]} \\
 & Eyeglasses & 0.68 {\tiny \color{gray} [0.36, 0.87]} & 0.79 {\tiny \color{gray} [0.45, 0.93]} & 0.72 {\tiny \color{gray} [0.56, 0.86]} & 0.76 {\tiny \color{gray} [0.53, 0.88]} \\
 & Feminine & 0.50 {\tiny \color{gray} [0.25, 0.72]} & 0.47 {\tiny \color{gray} [0.24, 0.72]} & 0.41 {\tiny \color{gray} [0.15, 0.69]} & 0.52 {\tiny \color{gray} [0.27, 0.70]} \\
 & Masculine/Receding Hairline & 0.41 {\tiny \color{gray} [0.24, 0.65]} & 0.35 {\tiny \color{gray} [0.23, 0.62]} & 0.44 {\tiny \color{gray} [0.29, 0.60]} & 0.33 {\tiny \color{gray} [0.15, 0.67]} \\
 & Smiling & 0.44 {\tiny \color{gray} [0.28, 0.67]} & 0.48 {\tiny \color{gray} [0.27, 0.79]} & 0.31 {\tiny \color{gray} [0.21, 0.60]} & 0.39 {\tiny \color{gray} [0.20, 0.65]} \\
\cline{1-6}
\multirow[t]{10}{*}{Intelligent} & Beard/Facial Hair & 0.59 {\tiny \color{gray} [0.27, 0.78]} & 0.54 {\tiny \color{gray} [0.33, 0.76]} & 0.50 {\tiny \color{gray} [0.30, 0.78]} & 0.52 {\tiny \color{gray} [0.32, 0.81]} \\
 & Blue Eyes & 0.39 {\tiny \color{gray} [0.17, 0.73]} & 0.52 {\tiny \color{gray} [0.27, 0.74]} & 0.57 {\tiny \color{gray} [0.23, 0.76]} & 0.40 {\tiny \color{gray} [0.17, 0.63]} \\
 & Child & 0.34 {\tiny \color{gray} [0.17, 0.55]} & 0.30 {\tiny \color{gray} [0.15, 0.49]} & 0.30 {\tiny \color{gray} [0.16, 0.44]} & 0.36 {\tiny \color{gray} [0.21, 0.65]} \\
 & Dark/Brown Skin & 0.64 {\tiny \color{gray} [0.42, 0.84]} & 0.64 {\tiny \color{gray} [0.39, 0.84]} & 0.59 {\tiny \color{gray} [0.35, 0.81]} & 0.46 {\tiny \color{gray} [0.31, 0.80]} \\
 & Earrings & 0.27 {\tiny \color{gray} [0.11, 0.54]} & 0.50 {\tiny \color{gray} [0.26, 0.75]} & 0.33 {\tiny \color{gray} [0.15, 0.59]} & 0.41 {\tiny \color{gray} [0.21, 0.80]} \\
 & East Asian & 0.29 {\tiny \color{gray} [0.15, 0.48]} & 0.43 {\tiny \color{gray} [0.15, 0.66]} & 0.41 {\tiny \color{gray} [0.17, 0.56]} & 0.43 {\tiny \color{gray} [0.26, 0.70]} \\
 & Eyeglasses & 0.85 {\tiny \color{gray} [0.71, 0.94]} & 0.74 {\tiny \color{gray} [0.57, 0.88]} & 0.80 {\tiny \color{gray} [0.68, 0.90]} & 0.75 {\tiny \color{gray} [0.24, 0.88]} \\
 & Feminine & 0.42 {\tiny \color{gray} [0.20, 0.76]} & 0.46 {\tiny \color{gray} [0.26, 0.70]} & 0.34 {\tiny \color{gray} [0.10, 0.59]} & 0.49 {\tiny \color{gray} [0.31, 0.82]} \\
 & Masculine/Receding Hairline & 0.39 {\tiny \color{gray} [0.24, 0.56]} & 0.35 {\tiny \color{gray} [0.18, 0.64]} & 0.46 {\tiny \color{gray} [0.26, 0.76]} & 0.39 {\tiny \color{gray} [0.18, 0.65]} \\
 & Smiling & 0.44 {\tiny \color{gray} [0.19, 0.73]} & 0.56 {\tiny \color{gray} [0.24, 0.81]} & 0.49 {\tiny \color{gray} [0.24, 0.70]} & 0.51 {\tiny \color{gray} [0.29, 0.84]} \\
\cline{1-6}
\multirow[t]{10}{*}{Serious} & Beard/Facial Hair & 0.45 {\tiny \color{gray} [0.27, 0.70]} & 0.55 {\tiny \color{gray} [0.32, 0.72]} & 0.37 {\tiny \color{gray} [0.16, 0.54]} & 0.49 {\tiny \color{gray} [0.16, 0.63]} \\
 & Blue Eyes & 0.29 {\tiny \color{gray} [0.15, 0.61]} & 0.50 {\tiny \color{gray} [0.24, 0.74]} & 0.48 {\tiny \color{gray} [0.25, 0.75]} & 0.32 {\tiny \color{gray} [0.11, 0.72]} \\
 & Child & 0.42 {\tiny \color{gray} [0.21, 0.65]} & 0.54 {\tiny \color{gray} [0.27, 0.71]} & 0.28 {\tiny \color{gray} [0.14, 0.67]} & 0.41 {\tiny \color{gray} [0.17, 0.60]} \\
 & Dark/Brown Skin & 0.58 {\tiny \color{gray} [0.24, 0.83]} & 0.64 {\tiny \color{gray} [0.39, 0.84]} & 0.66 {\tiny \color{gray} [0.24, 0.86]} & 0.47 {\tiny \color{gray} [0.30, 0.80]} \\
 & Earrings & 0.33 {\tiny \color{gray} [0.14, 0.60]} & 0.56 {\tiny \color{gray} [0.24, 0.67]} & 0.36 {\tiny \color{gray} [0.15, 0.63]} & 0.22 {\tiny \color{gray} [0.08, 0.52]} \\
 & East Asian & 0.40 {\tiny \color{gray} [0.24, 0.60]} & 0.34 {\tiny \color{gray} [0.15, 0.54]} & 0.28 {\tiny \color{gray} [0.12, 0.59]} & 0.46 {\tiny \color{gray} [0.15, 0.72]} \\
 & Eyeglasses & 0.72 {\tiny \color{gray} [0.36, 0.89]} & 0.54 {\tiny \color{gray} [0.32, 0.82]} & 0.87 {\tiny \color{gray} [0.69, 0.96]} & 0.74 {\tiny \color{gray} [0.56, 0.90]} \\
 & Feminine & 0.43 {\tiny \color{gray} [0.20, 0.79]} & 0.37 {\tiny \color{gray} [0.15, 0.54]} & 0.60 {\tiny \color{gray} [0.42, 0.78]} & 0.56 {\tiny \color{gray} [0.26, 0.74]} \\
 & Masculine/Receding Hairline & 0.42 {\tiny \color{gray} [0.17, 0.76]} & 0.69 {\tiny \color{gray} [0.40, 0.83]} & 0.25 {\tiny \color{gray} [0.10, 0.51]} & 0.43 {\tiny \color{gray} [0.14, 0.64]} \\
 & Smiling & 0.35 {\tiny \color{gray} [0.14, 0.60]} & 0.32 {\tiny \color{gray} [0.14, 0.59]} & 0.30 {\tiny \color{gray} [0.14, 0.49]} & 0.30 {\tiny \color{gray} [0.11, 0.57]} \\
\cline{1-6}
\multirow[t]{10}{*}{Trustworthy} & Beard/Facial Hair & 0.51 {\tiny \color{gray} [0.16, 0.80]} & 0.52 {\tiny \color{gray} [0.30, 0.71]} & 0.57 {\tiny \color{gray} [0.30, 0.81]} & 0.48 {\tiny \color{gray} [0.29, 0.75]} \\
 & Blue Eyes & 0.36 {\tiny \color{gray} [0.11, 0.65]} & 0.37 {\tiny \color{gray} [0.12, 0.66]} & 0.36 {\tiny \color{gray} [0.18, 0.69]} & 0.42 {\tiny \color{gray} [0.20, 0.67]} \\
 & Child & 0.51 {\tiny \color{gray} [0.24, 0.75]} & 0.65 {\tiny \color{gray} [0.40, 0.76]} & 0.40 {\tiny \color{gray} [0.18, 0.68]} & 0.40 {\tiny \color{gray} [0.17, 0.68]} \\
 & Dark/Brown Skin & 0.48 {\tiny \color{gray} [0.29, 0.76]} & 0.45 {\tiny \color{gray} [0.20, 0.72]} & 0.50 {\tiny \color{gray} [0.21, 0.78]} & 0.61 {\tiny \color{gray} [0.33, 0.77]} \\
 & Earrings & 0.49 {\tiny \color{gray} [0.28, 0.71]} & 0.59 {\tiny \color{gray} [0.37, 0.85]} & 0.53 {\tiny \color{gray} [0.32, 0.80]} & 0.52 {\tiny \color{gray} [0.27, 0.79]} \\
 & East Asian & 0.34 {\tiny \color{gray} [0.19, 0.54]} & 0.39 {\tiny \color{gray} [0.17, 0.58]} & 0.39 {\tiny \color{gray} [0.18, 0.62]} & 0.38 {\tiny \color{gray} [0.15, 0.64]} \\
 & Eyeglasses & 0.28 {\tiny \color{gray} [0.14, 0.47]} & 0.34 {\tiny \color{gray} [0.13, 0.63]} & 0.65 {\tiny \color{gray} [0.25, 0.83]} & 0.56 {\tiny \color{gray} [0.25, 0.82]} \\
 & Feminine & 0.67 {\tiny \color{gray} [0.41, 0.91]} & 0.65 {\tiny \color{gray} [0.29, 0.85]} & 0.73 {\tiny \color{gray} [0.56, 0.90]} & 0.48 {\tiny \color{gray} [0.23, 0.68]} \\
 & Masculine/Receding Hairline & 0.32 {\tiny \color{gray} [0.13, 0.73]} & 0.34 {\tiny \color{gray} [0.15, 0.70]} & 0.41 {\tiny \color{gray} [0.16, 0.64]} & 0.44 {\tiny \color{gray} [0.21, 0.64]} \\
 & Smiling & 0.54 {\tiny \color{gray} [0.27, 0.88]} & 0.59 {\tiny \color{gray} [0.33, 0.84]} & 0.81 {\tiny \color{gray} [0.63, 0.92]} & 0.64 {\tiny \color{gray} [0.36, 0.84]} \\
\cline{1-6}
\multirow[t]{10}{*}{Youthful} & Beard/Facial Hair & 0.46 {\tiny \color{gray} [0.14, 0.76]} & 0.46 {\tiny \color{gray} [0.18, 0.72]} & 0.74 {\tiny \color{gray} [0.39, 0.93]} & 0.44 {\tiny \color{gray} [0.22, 0.72]} \\
 & Blue Eyes & 0.38 {\tiny \color{gray} [0.21, 0.72]} & 0.49 {\tiny \color{gray} [0.23, 0.73]} & 0.46 {\tiny \color{gray} [0.35, 0.71]} & 0.41 {\tiny \color{gray} [0.08, 0.64]} \\
 & Child & 0.42 {\tiny \color{gray} [0.24, 0.63]} & 0.48 {\tiny \color{gray} [0.25, 0.73]} & 0.56 {\tiny \color{gray} [0.30, 0.79]} & 0.35 {\tiny \color{gray} [0.23, 0.63]} \\
 & Dark/Brown Skin & 0.40 {\tiny \color{gray} [0.23, 0.68]} & 0.42 {\tiny \color{gray} [0.29, 0.67]} & 0.53 {\tiny \color{gray} [0.17, 0.80]} & 0.58 {\tiny \color{gray} [0.32, 0.79]} \\
 & Earrings & 0.28 {\tiny \color{gray} [0.13, 0.50]} & 0.50 {\tiny \color{gray} [0.24, 0.68]} & 0.43 {\tiny \color{gray} [0.12, 0.75]} & 0.52 {\tiny \color{gray} [0.28, 0.69]} \\
 & East Asian & 0.76 {\tiny \color{gray} [0.51, 0.89]} & 0.63 {\tiny \color{gray} [0.44, 0.80]} & 0.79 {\tiny \color{gray} [0.65, 0.85]} & 0.49 {\tiny \color{gray} [0.22, 0.78]} \\
 & Eyeglasses & 0.38 {\tiny \color{gray} [0.21, 0.56]} & 0.30 {\tiny \color{gray} [0.16, 0.52]} & 0.48 {\tiny \color{gray} [0.24, 0.67]} & 0.34 {\tiny \color{gray} [0.26, 0.68]} \\
 & Feminine & 0.61 {\tiny \color{gray} [0.46, 0.89]} & 0.72 {\tiny \color{gray} [0.44, 0.87]} & 0.49 {\tiny \color{gray} [0.43, 0.61]} & 0.68 {\tiny \color{gray} [0.46, 0.90]} \\
 & Masculine/Receding Hairline & 0.49 {\tiny \color{gray} [0.21, 0.81]} & 0.31 {\tiny \color{gray} [0.18, 0.60]} & 0.49 {\tiny \color{gray} [0.26, 0.69]} & 0.30 {\tiny \color{gray} [0.23, 0.83]} \\
 & Smiling & 0.36 {\tiny \color{gray} [0.13, 0.67]} & 0.28 {\tiny \color{gray} [0.13, 0.54]} & 0.29 {\tiny \color{gray} [0.23, 0.46]} & 0.35 {\tiny \color{gray} [0.15, 0.58]} \\
\cline{1-6}
\end{longtable}

}

\section{Prompts}
\label{app:prompts}

The prompts used for the task instructions are the ones sent to the MLLMs to complete the task (i.e. match the target). The prompts passed to the generative image model FLUX.1-schnell are fixed within each domain, and constrain the visual format so that the slider perturbations are the dominant source of variation across stimuli.

\begin{promptbox}[Tasks]
\textbf{Faces}: \textit{Which face better matches this adjective: \{target\}?}

\textbf{Affordances}: \textit{Which object better matches this property: \{target\}?}

\textbf{Aesthetics}: \textit{Which artwork better matches this description: \{target\}?}

\textbf{Authenticity}: \textit{Which image looks more \{target\}?}
\end{promptbox}

\begin{promptbox}[Image Generation]
\textbf{Faces}: \textit{id photo of a person from the shoulders up, plain white background,
looking straight at camera, even lighting, centered, eye level camera angle,
unobstructed face}

\textbf{Affordances}: \textit{a single decorative colored object with a unique shape, isolated photograph, large in the frame, light gray seamless backdrop, soft studio lighting, centered, soft shadow, contrast, photorealistic}

\textbf{Aesthetics}: \textit{a painting}

\textbf{Authenticity}: \textit{a photograph of an outdoor city scene}
\end{promptbox}

\section{Slider Validation}
\label{app:sliderspace_validation}
Beyond per-slider behavior, we need to verify that joint configurations of the sliders can actually steer the generative model toward diverse image distributions. This is a fundamental constraint our method inherits from the underlying generator (as indicated in \Cref{sec:limitations}), and the reason why we trained with diverse prompts (details in \Cref{app:sliderspace_training}). As in \cite{harrison2020gibbs} and their StyleGAN, we can only recover priors over the regions of stimulus space that SliderSpace+FLUX can reach. If the latent does not admit configurations depicting, say, a particular demographic, the absence of that demographic from the recovered posterior of a target is uninformative---not evidence that the model does not associate the two.

To probe the reachable subspace in our trained sliders, we run separate chains targeting a small set of demographic descriptors---\textit{woman}, \textit{man}, \textit{black person}, \textit{white person}, \textit{asian person}, \textit{elderly person}, \textit{young person}, \textit{hispanic person}, \textit{indigenous person}, \textit{middle eastern person}---and inspect whether each chain converges to images recognizable as instances of the descriptor. We use \textit{Qwen3-VL 235B}, 1 chain per target, and 1,000 trials per chain for the experiments. Whether each chain reaches a region matching its descriptor is in turn judged automatically by \textit{Gemini 3 Flash}: each of the top-$k$ accepted images is shown to the autointerpretability model alongside the target descriptor, and the per-target match rate is the fraction of the $k$ images judged as matches. A positive outcome shows that the corresponding region is reachable, but we make no claim that all reachable regions are equally easy to reach, and demographic configurations that dominate FLUX's training data are likely more accessible than those that do not.

Eight of the ten descriptors are reached with perfect agreement; the exceptions are \textit{hispanic person} ($20\%$) and \textit{elderly person} ($0\%$). While FLUX could still generate images following those distributions, our limited number of sliders might not be able to steer the generation in that direction. This means concepts the sliders cannot reach are systematically absent from any recovered posterior regardless of whether the model holds the association. Including more sliders that provide finer control over the generation could solve this issue, but it comes at a computational cost that is prohibitive for our large-scale experiments.

\section{Top \& Bottom Sliders per Domain}
\label{app:additional_analysis}

\begin{figure*}[!htb]
    \centering
    \includegraphics[width=\textwidth]{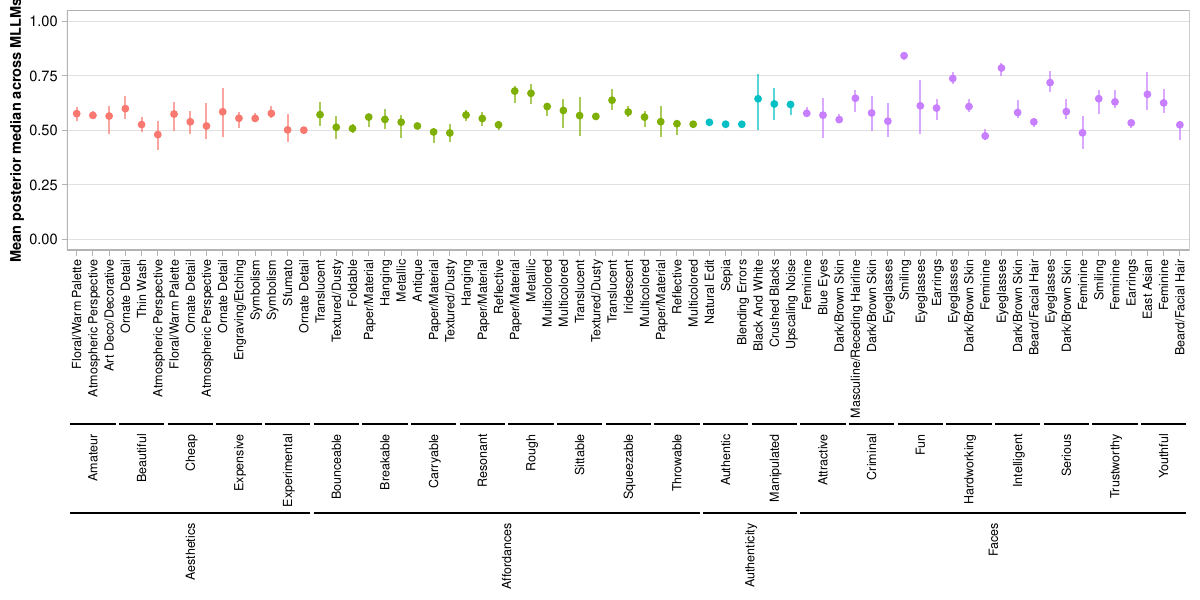}
    \caption{Top-3 (\textbf{top}) sliders per target across the four models. Points are cross-model means; bars are cross-model interquartile ranges.}
    \label{fig:headline-top-nested-axis}
\end{figure*}

\begin{figure*}[!htb]
    \centering
    \includegraphics[width=\textwidth]{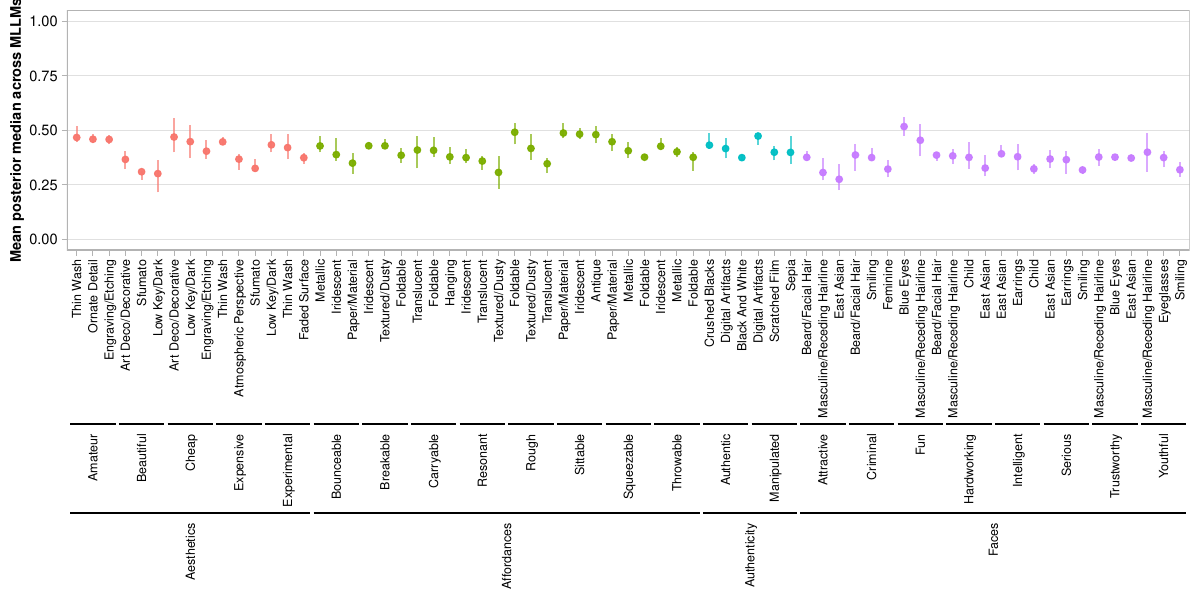}
    \caption{Bottom-3 (\textbf{bottom}) sliders per target across the four models. Points are cross-model means; bars are cross-model interquartile ranges.}
    \label{fig:headline-bottom-nested-axis}
\end{figure*}

\Cref{fig:headline-top-nested-axis,fig:headline-bottom-nested-axis} expand every target into its top-3 and bottom-3 sliders across models, giving a per-target read of the priors that the pooled view averages over. This shows: (i) Face targets often have a clear single dominant slider plus secondary modifiers (ii) Other targets show weaker single-slider dominance and more cross-cutting cues (iii) The two authenticity targets are mostly antipodal: \textit{Manipulated} loads on \textit{Black and White}/\textit{Crushed Blacks}/\textit{Upscaling Noise}, \textit{Authentic} on \textit{Natural Edit}/\textit{Sepia}/\textit{Blending Errors}, and the directions are roughly inverse.

\section{Slider Training}
\label{app:sliderspace_training}

We train SliderSpace \citep{gandikota2025sliderspace} sliders for each domain over FLUX.1-schnell \citep{labs2025flux}. For each domain, $64$ candidate sliders are trained using $50{,}000$ CLIP-encoded samples for PCA with a batch size of 32. Training runs for $3000$ iterations per slider with batch size $1$, learning rate $2 \times 10^{-3}$, LoRA rank $1$, $\alpha=1$, two denoising steps per training step, guidance scale $0$, and bfloat16 precision. At inference time, we use $4$ denoising steps, $0$ guidance scale, 512x512 resolution, and $256$ maximum sequence length.

For training, we use diverse prompts to learn diverse sliders, but prompt to isolate features that matter (e.g. face variation) while keeping other things invariant (e.g., facing camera, background).

\begin{promptbox}[Training Prompts: Faces]
\textit{id photo of a person from the shoulders up, plain white background, looking straight at camera, even lighting, centered, eye level camera angle, unobstructed face}

\textit{id photo of a man from the shoulders up, plain white background, looking straight at camera, even lighting, centered, eye level camera angle, unobstructed face}

\textit{id photo of a woman from the shoulders up, plain white background, looking straight at camera, even lighting, centered, eye level camera angle, unobstructed face}

\textit{id photo of a girl from the shoulders up, plain white background, looking straight at camera, even lighting, centered, eye level camera angle, unobstructed face}

\textit{id photo of a boy from the shoulders up, plain white background, looking straight at camera, even lighting, centered, eye level camera angle, unobstructed face}

\textit{id photo of a teenage White European man from the shoulders up, plain white background, looking straight at camera, even lighting, centered, eye level camera angle, unobstructed face}

\textit{id photo of a teenage White European woman from the shoulders up, plain white background, looking straight at camera, even lighting, centered, eye level camera angle, unobstructed face}

\textit{id photo of a young adult White European man from the shoulders up, plain white background, looking straight at camera, even lighting, centered, eye level camera angle, unobstructed face}

\textit{id photo of a young adult White European woman from the shoulders up, plain white background, looking straight at camera, even lighting, centered, eye level camera angle, unobstructed face}

\textit{id photo of a middle-aged White European man from the shoulders up, plain white background, looking straight at camera, even lighting, centered, eye level camera angle, unobstructed face}

\textit{id photo of a middle-aged White European woman from the shoulders up, plain white background, looking straight at camera, even lighting, centered, eye level camera angle, unobstructed face}

\textit{id photo of an elderly White European man from the shoulders up, plain white background, looking straight at camera, even lighting, centered, eye level camera angle, unobstructed face}

\textit{id photo of an elderly White European woman from the shoulders up, plain white background, looking straight at camera, even lighting, centered, eye level camera angle, unobstructed face}

\textit{id photo of a teenage Black African man from the shoulders up, plain white background, looking straight at camera, even lighting, centered, eye level camera angle, unobstructed face}

\textit{id photo of a teenage Black African woman from the shoulders up, plain white background, looking straight at camera, even lighting, centered, eye level camera angle, unobstructed face}

\textit{id photo of a young adult Black African man from the shoulders up, plain white background, looking straight at camera, even lighting, centered, eye level camera angle, unobstructed face}

\textit{id photo of a young adult Black African woman from the shoulders up, plain white background, looking straight at camera, even lighting, centered, eye level camera angle, unobstructed face}

\textit{id photo of a middle-aged Black African man from the shoulders up, plain white background, looking straight at camera, even lighting, centered, eye level camera angle, unobstructed face}

\textit{id photo of a middle-aged Black African woman from the shoulders up, plain white background, looking straight at camera, even lighting, centered, eye level camera angle, unobstructed face}

\textit{id photo of an elderly Black African man from the shoulders up, plain white background, looking straight at camera, even lighting, centered, eye level camera angle, unobstructed face}

\textit{id photo of an elderly Black African woman from the shoulders up, plain white background, looking straight at camera, even lighting, centered, eye level camera angle, unobstructed face}

\textit{id photo of a teenage East Asian man from the shoulders up, plain white background, looking straight at camera, even lighting, centered, eye level camera angle, unobstructed face}

\textit{id photo of a teenage East Asian woman from the shoulders up, plain white background, looking straight at camera, even lighting, centered, eye level camera angle, unobstructed face}

\textit{id photo of a young adult East Asian man from the shoulders up, plain white background, looking straight at camera, even lighting, centered, eye level camera angle, unobstructed face}

\textit{id photo of a young adult East Asian woman from the shoulders up, plain white background, looking straight at camera, even lighting, centered, eye level camera angle, unobstructed face}

\textit{id photo of a middle-aged East Asian man from the shoulders up, plain white background, looking straight at camera, even lighting, centered, eye level camera angle, unobstructed face}

\textit{id photo of a middle-aged East Asian woman from the shoulders up, plain white background, looking straight at camera, even lighting, centered, eye level camera angle, unobstructed face}

\textit{id photo of an elderly East Asian man from the shoulders up, plain white background, looking straight at camera, even lighting, centered, eye level camera angle, unobstructed face}

\textit{id photo of an elderly East Asian woman from the shoulders up, plain white background, looking straight at camera, even lighting, centered, eye level camera angle, unobstructed face}

\textit{id photo of a teenage South Asian man from the shoulders up, plain white background, looking straight at camera, even lighting, centered, eye level camera angle, unobstructed face}

\textit{id photo of a teenage South Asian woman from the shoulders up, plain white background, looking straight at camera, even lighting, centered, eye level camera angle, unobstructed face}

\textit{id photo of a young adult South Asian man from the shoulders up, plain white background, looking straight at camera, even lighting, centered, eye level camera angle, unobstructed face}

\textit{id photo of a young adult South Asian woman from the shoulders up, plain white background, looking straight at camera, even lighting, centered, eye level camera angle, unobstructed face}

\textit{id photo of a middle-aged South Asian man from the shoulders up, plain white background, looking straight at camera, even lighting, centered, eye level camera angle, unobstructed face}

\textit{id photo of a middle-aged South Asian woman from the shoulders up, plain white background, looking straight at camera, even lighting, centered, eye level camera angle, unobstructed face}

\textit{id photo of an elderly South Asian man from the shoulders up, plain white background, looking straight at camera, even lighting, centered, eye level camera angle, unobstructed face}

\textit{id photo of an elderly South Asian woman from the shoulders up, plain white background, looking straight at camera, even lighting, centered, eye level camera angle, unobstructed face}

\textit{id photo of a teenage Southeast Asian man from the shoulders up, plain white background, looking straight at camera, even lighting, centered, eye level camera angle, unobstructed face}

\textit{id photo of a teenage Southeast Asian woman from the shoulders up, plain white background, looking straight at camera, even lighting, centered, eye level camera angle, unobstructed face}

\textit{id photo of a young adult Southeast Asian man from the shoulders up, plain white background, looking straight at camera, even lighting, centered, eye level camera angle, unobstructed face}

\textit{id photo of a young adult Southeast Asian woman from the shoulders up, plain white background, looking straight at camera, even lighting, centered, eye level camera angle, unobstructed face}

\textit{id photo of a middle-aged Southeast Asian man from the shoulders up, plain white background, looking straight at camera, even lighting, centered, eye level camera angle, unobstructed face}

\textit{id photo of a middle-aged Southeast Asian woman from the shoulders up, plain white background, looking straight at camera, even lighting, centered, eye level camera angle, unobstructed face}

\textit{id photo of an elderly Southeast Asian man from the shoulders up, plain white background, looking straight at camera, even lighting, centered, eye level camera angle, unobstructed face}

\textit{id photo of an elderly Southeast Asian woman from the shoulders up, plain white background, looking straight at camera, even lighting, centered, eye level camera angle, unobstructed face}

\textit{id photo of a teenage Middle Eastern man from the shoulders up, plain white background, looking straight at camera, even lighting, centered, eye level camera angle, unobstructed face}

\textit{id photo of a teenage Middle Eastern woman from the shoulders up, plain white background, looking straight at camera, even lighting, centered, eye level camera angle, unobstructed face}

\textit{id photo of a young adult Middle Eastern man from the shoulders up, plain white background, looking straight at camera, even lighting, centered, eye level camera angle, unobstructed face}

\textit{id photo of a young adult Middle Eastern woman from the shoulders up, plain white background, looking straight at camera, even lighting, centered, eye level camera angle, unobstructed face}

\textit{id photo of a middle-aged Middle Eastern man from the shoulders up, plain white background, looking straight at camera, even lighting, centered, eye level camera angle, unobstructed face}

\textit{id photo of a middle-aged Middle Eastern woman from the shoulders up, plain white background, looking straight at camera, even lighting, centered, eye level camera angle, unobstructed face}

\textit{id photo of an elderly Middle Eastern man from the shoulders up, plain white background, looking straight at camera, even lighting, centered, eye level camera angle, unobstructed face}

\textit{id photo of an elderly Middle Eastern woman from the shoulders up, plain white background, looking straight at camera, even lighting, centered, eye level camera angle, unobstructed face}

\textit{id photo of a teenage Hispanic Latino man from the shoulders up, plain white background, looking straight at camera, even lighting, centered, eye level camera angle, unobstructed face}

\textit{id photo of a teenage Hispanic Latino woman from the shoulders up, plain white background, looking straight at camera, even lighting, centered, eye level camera angle, unobstructed face}

\textit{id photo of a young adult Hispanic Latino man from the shoulders up, plain white background, looking straight at camera, even lighting, centered, eye level camera angle, unobstructed face}

\textit{id photo of a young adult Hispanic Latino woman from the shoulders up, plain white background, looking straight at camera, even lighting, centered, eye level camera angle, unobstructed face}

\textit{id photo of a middle-aged Hispanic Latino man from the shoulders up, plain white background, looking straight at camera, even lighting, centered, eye level camera angle, unobstructed face}

\textit{id photo of a middle-aged Hispanic Latino woman from the shoulders up, plain white background, looking straight at camera, even lighting, centered, eye level camera angle, unobstructed face}

\textit{id photo of an elderly Hispanic Latino man from the shoulders up, plain white background, looking straight at camera, even lighting, centered, eye level camera angle, unobstructed face}

\textit{id photo of an elderly Hispanic Latino woman from the shoulders up, plain white background, looking straight at camera, even lighting, centered, eye level camera angle, unobstructed face}

\textit{id photo of a teenage Indigenous Native American man from the shoulders up, plain white background, looking straight at camera, even lighting, centered, eye level camera angle, unobstructed face}

\textit{id photo of a teenage Indigenous Native American woman from the shoulders up, plain white background, looking straight at camera, even lighting, centered, eye level camera angle, unobstructed face}

\textit{id photo of a young adult Indigenous Native American man from the shoulders up, plain white background, looking straight at camera, even lighting, centered, eye level camera angle, unobstructed face}

\textit{id photo of a young adult Indigenous Native American woman from the shoulders up, plain white background, looking straight at camera, even lighting, centered, eye level camera angle, unobstructed face}

\textit{id photo of a middle-aged Indigenous Native American man from the shoulders up, plain white background, looking straight at camera, even lighting, centered, eye level camera angle, unobstructed face}

\textit{id photo of a middle-aged Indigenous Native American woman from the shoulders up, plain white background, looking straight at camera, even lighting, centered, eye level camera angle, unobstructed face}

\textit{id photo of an elderly Indigenous Native American man from the shoulders up, plain white background, looking straight at camera, even lighting, centered, eye level camera angle, unobstructed face}

\textit{id photo of an elderly Indigenous Native American woman from the shoulders up, plain white background, looking straight at camera, even lighting, centered, eye level camera angle, unobstructed face}

\textit{id photo of a person with a happy expression from the shoulders up, plain white background, looking straight at camera, even lighting, centered, eye level camera angle, unobstructed face}

\textit{id photo of a person with a sad expression from the shoulders up, plain white background, looking straight at camera, even lighting, centered, eye level camera angle, unobstructed face}

\textit{id photo of a person with an angry expression from the shoulders up, plain white background, looking straight at camera, even lighting, centered, eye level camera angle, unobstructed face}

\textit{id photo of a person with a surprised expression from the shoulders up, plain white background, looking straight at camera, even lighting, centered, eye level camera angle, unobstructed face}

\textit{id photo of a person with a neutral expression from the shoulders up, plain white background, looking straight at camera, even lighting, centered, eye level camera angle, unobstructed face}

\textit{id photo of a person with a disgusted expression from the shoulders up, plain white background, looking straight at camera, even lighting, centered, eye level camera angle, unobstructed face}

\textit{id photo of a person with a fearful expression from the shoulders up, plain white background, looking straight at camera, even lighting, centered, eye level camera angle, unobstructed face}

\textit{id photo of a person with a contemptuous expression from the shoulders up, plain white background, looking straight at camera, even lighting, centered, eye level camera angle, unobstructed face}

\textit{id photo of a person laughing from the shoulders up, plain white background, looking straight at camera, even lighting, centered, eye level camera angle, unobstructed face}

\textit{id photo of a person crying from the shoulders up, plain white background, looking straight at camera, even lighting, centered, eye level camera angle, unobstructed face}

\textit{id photo of a person with very light skin from the shoulders up, plain white background, looking straight at camera, even lighting, centered, eye level camera angle, unobstructed face}

\textit{id photo of a person with light skin from the shoulders up, plain white background, looking straight at camera, even lighting, centered, eye level camera angle, unobstructed face}

\textit{id photo of a person with medium skin from the shoulders up, plain white background, looking straight at camera, even lighting, centered, eye level camera angle, unobstructed face}

\textit{id photo of a person with dark skin from the shoulders up, plain white background, looking straight at camera, even lighting, centered, eye level camera angle, unobstructed face}

\textit{id photo of a person with very dark skin from the shoulders up, plain white background, looking straight at camera, even lighting, centered, eye level camera angle, unobstructed face}

\textit{id photo of a person with short straight hair from the shoulders up, plain white background, looking straight at camera, even lighting, centered, eye level camera angle, unobstructed face}

\textit{id photo of a person with long straight hair from the shoulders up, plain white background, looking straight at camera, even lighting, centered, eye level camera angle, unobstructed face}

\textit{id photo of a person with short curly hair from the shoulders up, plain white background, looking straight at camera, even lighting, centered, eye level camera angle, unobstructed face}

\textit{id photo of a person with long curly hair from the shoulders up, plain white background, looking straight at camera, even lighting, centered, eye level camera angle, unobstructed face}

\textit{id photo of a person with coily hair from the shoulders up, plain white background, looking straight at camera, even lighting, centered, eye level camera angle, unobstructed face}

\textit{id photo of a person with wavy hair from the shoulders up, plain white background, looking straight at camera, even lighting, centered, eye level camera angle, unobstructed face}

\textit{id photo of a person with black hair from the shoulders up, plain white background, looking straight at camera, even lighting, centered, eye level camera angle, unobstructed face}

\textit{id photo of a person with brown hair from the shoulders up, plain white background, looking straight at camera, even lighting, centered, eye level camera angle, unobstructed face}

\textit{id photo of a person with blonde hair from the shoulders up, plain white background, looking straight at camera, even lighting, centered, eye level camera angle, unobstructed face}

\textit{id photo of a person with red hair from the shoulders up, plain white background, looking straight at camera, even lighting, centered, eye level camera angle, unobstructed face}

\textit{id photo of a person with grey hair from the shoulders up, plain white background, looking straight at camera, even lighting, centered, eye level camera angle, unobstructed face}

\textit{id photo of a person with white hair from the shoulders up, plain white background, looking straight at camera, even lighting, centered, eye level camera angle, unobstructed face}

\textit{id photo of a person with a beard from the shoulders up, plain white background, looking straight at camera, even lighting, centered, eye level camera angle, unobstructed face}

\textit{id photo of a person with a mustache from the shoulders up, plain white background, looking straight at camera, even lighting, centered, eye level camera angle, unobstructed face}

\textit{id photo of a person with stubble from the shoulders up, plain white background, looking straight at camera, even lighting, centered, eye level camera angle, unobstructed face}

\textit{id photo of a bald person from the shoulders up, plain white background, looking straight at camera, even lighting, centered, eye level camera angle, unobstructed face}

\textit{id photo of a person with an oval face from the shoulders up, plain white background, looking straight at camera, even lighting, centered, eye level camera angle, unobstructed face}

\textit{id photo of a person with a round face from the shoulders up, plain white background, looking straight at camera, even lighting, centered, eye level camera angle, unobstructed face}

\textit{id photo of a person with a square face from the shoulders up, plain white background, looking straight at camera, even lighting, centered, eye level camera angle, unobstructed face}

\textit{id photo of a person with a narrow face from the shoulders up, plain white background, looking straight at camera, even lighting, centered, eye level camera angle, unobstructed face}

\textit{id photo of a person wearing glasses from the shoulders up, plain white background, looking straight at camera, even lighting, centered, eye level camera angle, unobstructed face}

\textit{id photo of a person wearing sunglasses from the shoulders up, plain white background, looking straight at camera, even lighting, centered, eye level camera angle, unobstructed face}

\textit{id photo of a person wearing a hijab from the shoulders up, plain white background, looking straight at camera, even lighting, centered, eye level camera angle, unobstructed face}

\textit{id photo of a person wearing a turban from the shoulders up, plain white background, looking straight at camera, even lighting, centered, eye level camera angle, unobstructed face}

\textit{id photo of a person wearing a hat from the shoulders up, plain white background, looking straight at camera, even lighting, centered, eye level camera angle, unobstructed face}

\textit{id photo of a person with light makeup from the shoulders up, plain white background, looking straight at camera, even lighting, centered, eye level camera angle, unobstructed face}

\textit{id photo of a person with heavy makeup from the shoulders up, plain white background, looking straight at camera, even lighting, centered, eye level camera angle, unobstructed face}

\textit{id photo of a person with earrings from the shoulders up, plain white background, looking straight at camera, even lighting, centered, eye level camera angle, unobstructed face}

\textit{id photo of a person with facial piercings from the shoulders up, plain white background, looking straight at camera, even lighting, centered, eye level camera angle, unobstructed face}

\textit{id photo of a person with a subtle neck tattoo from the shoulders up, plain white background, looking straight at camera, even lighting, centered, eye level camera angle, unobstructed face}

\textit{id photo of a person with prominent facial tattoos from the shoulders up, plain white background, looking straight at camera, even lighting, centered, eye level camera angle, unobstructed face}

\textit{id photo of a person with freckles from the shoulders up, plain white background, looking straight at camera, even lighting, centered, eye level camera angle, unobstructed face}

\textit{id photo of a person with vitiligo from the shoulders up, plain white background, looking straight at camera, even lighting, centered, eye level camera angle, unobstructed face}

\textit{id photo of a person with acne from the shoulders up, plain white background, looking straight at camera, even lighting, centered, eye level camera angle, unobstructed face}

\textit{id photo of a slim person from the shoulders up, plain white background, looking straight at camera, even lighting, centered, eye level camera angle, unobstructed face}

\textit{id photo of a heavyset person from the shoulders up, plain white background, looking straight at camera, even lighting, centered, eye level camera angle, unobstructed face}

\textit{id photo of a person wearing hearing aids from the shoulders up, plain white background, looking straight at camera, even lighting, centered, eye level camera angle, unobstructed face}

\textit{id photo of a person with a facial difference from the shoulders up, plain white background, looking straight at camera, even lighting, centered, eye level camera angle, unobstructed face}

\textit{id photo of a person with Down syndrome from the shoulders up, plain white background, looking straight at camera, even lighting, centered, eye level camera angle, unobstructed face}
\end{promptbox}

\begin{promptbox}[Training Prompts: Affordances]
\textit{a photograph of a small shiny metal object, plain white background}

\textit{a photograph of a rough wooden tool, plain white background}

\textit{a photograph of a transparent glass container, plain white background}

\textit{a photograph of a colorful plastic object, plain white background}

\textit{a photograph of a soft fabric pouch or bag, plain white background}

\textit{a photograph of a rubber or silicone flexible object, plain white background}

\textit{a photograph of a ceramic or porcelain container, plain white background}

\textit{a photograph of a heavy stone or rock object, plain white background}

\textit{a photograph of a paper or cardboard object, plain white background}

\textit{a photograph of a leather object, plain white background}

\textit{a photograph of a round spherical object, plain white background}

\textit{a photograph of a flat thin disc-shaped object, plain white background}

\textit{a photograph of a long cylindrical object, plain white background}

\textit{a photograph of a sharp pointed object, plain white background}

\textit{a photograph of a cubic box-shaped object, plain white background}

\textit{a photograph of an irregular asymmetric object, plain white background}

\textit{a photograph of a ring or hoop-shaped object, plain white background}

\textit{a photograph of a curved arched object, plain white background}

\textit{a photograph of a tiny miniature object that fits on a fingertip, plain white background}

\textit{a photograph of a small pocket-sized object, plain white background}

\textit{a photograph of a large heavy bulky object, plain white background}

\textit{a photograph of an object with a long handle, plain white background}

\textit{a photograph of an object with a looped handle, plain white background}

\textit{a photograph of an object with a grip or knob, plain white background}

\textit{a photograph of a smooth object with no handle or grip, plain white background}

\textit{a photograph of a hollow container with a wide opening, plain white background}

\textit{a photograph of a hollow container with a narrow spout, plain white background}

\textit{a photograph of a solid object with no openings, plain white background}

\textit{a photograph of an object with multiple holes or perforations, plain white background}

\textit{a photograph of a stackable flat object, plain white background}

\textit{a photograph of a collapsible or foldable object, plain white background}

\textit{a photograph of a very smooth polished object, plain white background}

\textit{a photograph of a rough heavily textured object, plain white background}

\textit{a photograph of a soft padded object, plain white background}

\textit{a photograph of a sharp-edged angular object, plain white background}

\textit{a photograph of a hand tool like a wrench or hammer, plain white background}

\textit{a photograph of a kitchen utensil, plain white background}

\textit{a photograph of a container with a lid, plain white background}

\textit{a photograph of a writing instrument, plain white background}

\textit{a photograph of a fastener like a bolt, screw or clip, plain white background}

\textit{a photograph of a natural object like a shell, seed or stone, plain white background}

\textit{a photograph of an electronic device or gadget, plain white background}

\textit{a photograph of a woven or braided object, plain white background}

\textit{a photograph of a sharp cutting tool, plain white background}

\textit{a photograph of a rope or cord-like object, plain white background}

\textit{a photograph of a flat plate or tray, plain white background}

\textit{a photograph of a measuring or gripping tool, plain white background}

\textit{a photograph of a jointed or hinged object, plain white background}

\textit{a photograph of a transparent object you can see through, plain white background}

\textit{a photograph of a hollow object with a narrow neck, plain white background}

\textit{a photograph of a small metal object with a long thin handle, plain white background}

\textit{a photograph of a large wooden flat object with no openings, plain white background}

\textit{a photograph of a transparent glass object with a narrow neck and wide base, plain white background}

\textit{a photograph of a soft fabric object with a looped handle, plain white background}

\textit{a photograph of a small ceramic container with a wide opening and no lid, plain white background}

\textit{a photograph of a heavy metal object with multiple holes, plain white background}

\textit{a photograph of a flexible rubber object with a cylindrical shape, plain white background}

\textit{a photograph of a small smooth spherical metal object, plain white background}

\textit{a photograph of a large hollow plastic container with a lid, plain white background}

\textit{a photograph of a flat wooden object with a handle on one end, plain white background}

\textit{a photograph of a sharp metal object with a pointed tip and a grip, plain white background}

\textit{a photograph of a small plastic object with multiple buttons or controls, plain white background}

\textit{a photograph of a woven object with an open basket shape, plain white background}

\textit{a photograph of a long thin wooden cylindrical object, plain white background}

\textit{a photograph of a heavy stone flat rectangular object, plain white background}

\textit{a photograph of a metal coiled spiral object, plain white background}

\textit{a photograph of a small leather flat foldable object, plain white background}

\textit{a photograph of a ceramic object with a handle and a spout, plain white background}

\textit{a photograph of a metal hinged object with two arms, plain white background}

\textit{a photograph of a soft padded fabric object with a zipper, plain white background}

\textit{a photograph of a transparent plastic object with a narrow tip, plain white background}

\textit{a photograph of a rough stone object with an irregular pointed shape, plain white background}

\textit{a photograph of a large cylindrical metal container with a lid, plain white background}

\textit{a photograph of a small wooden object with an irregular carved shape, plain white background}

\textit{a photograph of a flat perforated metal object with a long handle, plain white background}

\textit{a photograph of a glass spherical object with a smooth polished surface, plain white background}

\textit{a photograph of a rubber object with a bulbous hollow squeezable shape, plain white background}

\textit{a photograph of a metal object with a threaded cylindrical shape, plain white background}

\textit{a photograph of a large fabric object that collapses flat, plain white background}

\textit{a photograph of a small electronic object with a flat rectangular shape and a screen, plain white background}
\end{promptbox}

\begin{promptbox}[Training Prompts: Aesthetics]
\textit{Renaissance oil painting, classical composition, rich colors, detailed figures}

\textit{Baroque oil painting, dramatic chiaroscuro lighting, dynamic composition}

\textit{Romantic painting, dramatic landscape, emotional intensity, stormy atmosphere}

\textit{Impressionist painting, loose brushstrokes, natural outdoor light, dappled colors}

\textit{Post-Impressionist painting, bold expressive colors, thick brushwork}

\textit{Pointillist painting, small dots of color, vibrant scene}

\textit{Cubist painting, fragmented geometric forms, multiple perspectives simultaneously}

\textit{Surrealist painting, dreamlike bizarre imagery, uncanny juxtapositions}

\textit{Abstract Expressionist painting, gestural brushstrokes, raw emotion, large canvas}

\textit{Minimalist artwork, simple geometric forms, very limited color palette}

\textit{Pop Art, bold flat colors, commercial imagery, graphic design influence}

\textit{Art Nouveau illustration, organic flowing lines, decorative natural motifs}

\textit{Expressionist painting, distorted forms, intense emotional colors}

\textit{Color field painting, large areas of flat uniform color, subtle gradients}

\textit{Contemporary abstract painting, mixed media, conceptual layered composition}

\textit{Contemporary street art, urban graffiti style, bold outlines, spray paint}

\textit{Contemporary digital art, modern aesthetic, vibrant colors, crisp edges}

\textit{Contemporary conceptual artwork, thought-provoking, unconventional materials}

\textit{Medieval illuminated manuscript style, gold leaf, flat decorative figures}

\textit{Ancient fresco painting, muted earth tones, classical figures on wall}

\textit{Japanese woodblock print, flat areas of color, decorative patterns, nature scene}

\textit{Watercolor painting, translucent washes, soft edges, delicate}

\textit{Charcoal drawing, dramatic shadows, expressive gestural lines}

\textit{Engraving or etching print, fine detailed lines, black and white}

\textit{Dark melancholic painting, somber muted tones, moody oppressive atmosphere}

\textit{Joyful colorful artwork, bright vibrant palette, uplifting cheerful scene}

\textit{Serene peaceful landscape painting, calm soft atmosphere, gentle light}

\textit{Dramatic epic historical painting, monumental composition, heroic scene}

\textit{Unsettling eerie artwork, disturbing imagery, dark surreal atmosphere}

\textit{Classical still life oil painting, detailed objects, dramatic lighting}

\textit{fine art black and white photography, dramatic contrast, artistic composition}

\textit{long exposure fine art photography, light trails, ethereal motion blur}

\textit{classical marble sculpture, carved stone, ancient Greek or Roman style}

\textit{folk art naive painting, flat simple figures, bright colors, outsider art}

\textit{intricate geometric patterns artwork, tessellations, symmetry, ornate decoration}

\textit{Gothic art, dark medieval religious imagery, pointed arches, somber figures}

\textit{Bauhaus Constructivist design, functional geometric shapes, primary colors, bold layout}

\textit{photorealistic painting indistinguishable from a photograph, hyper-detailed}

\textit{Chinese ink wash painting, sparse brushstrokes, negative space, misty mountains}

\textit{African tribal art, bold patterns, masks, ceremonial motifs, earthy colors}
\end{promptbox}

All authenticity prompts share a fixed scene placeholder, expanded to: \textit{a narrow alley flanked by Victorian red brick buildings with arched windows, wet cobblestone ground, a green wooden door on the left wall, an iron streetlamp above, a wrought iron gate at the far end with a foggy street beyond}.

\begin{promptbox}[Training Prompts: Authenticity]

\textit{CCTV security camera footage of [SCENE], grainy high-angle wide-angle shot, timestamp overlay, washed out colors}

\textit{smartphone photo of [SCENE] taken at arm's length, slight wide-angle distortion, casual framing}

\textit{smartphone snapshot of [SCENE], casual handheld shot, slightly off-center framing, mobile photography}

\textit{disposable film camera photo of [SCENE], overexposed flash, light leaks, grainy, washed out colors}

\textit{sharp professional photograph of [SCENE], shallow depth of field, blurred background bokeh, high resolution}

\textit{dashcam footage of [SCENE], road perspective through windshield, wide angle, slightly fisheye}

\textit{bodycam footage of [SCENE], chest-level perspective, slight motion blur}

\textit{raw unedited photograph of [SCENE], natural colors, no post-processing, flat exposure}

\textit{heavily retouched photograph of [SCENE], enhanced colors, airbrushed appearance, magazine style}

\textit{cinematic color graded photograph of [SCENE], teal and orange tones, film look}

\textit{vintage film filter photograph of [SCENE], faded colors, film grain, retro look}

\textit{HDR photograph of [SCENE], high dynamic range, enhanced detail in highlights and shadows}

\textit{photograph of [SCENE] saved at very low JPEG quality, blocky compression artifacts visible, degraded}

\textit{analog film grain photograph of [SCENE], textured grain, warm tones, film photography}

\textit{AI-generated image of [SCENE], uncanny realism, slight surreal quality, synthetic appearance}

\textit{natural light photograph of [SCENE], soft outdoor lighting, no flash, available light}

\textit{harsh artificial light photograph of [SCENE], strong shadows, fluorescent or neon lighting}

\textit{commercial advertising photograph of [SCENE], polished and stylized, studio quality}

\textit{photojournalism documentary photograph of [SCENE], candid moment, high contrast, reportage style}

\textit{composited photograph of [SCENE] with a fake background, cutout edges visible, inconsistent lighting}

\textit{Polaroid instant photo of [SCENE], square format, faded colors, white border, slightly overexposed}

\textit{long exposure photograph of [SCENE], motion blur on moving elements, sharp background, slow shutter speed}

\textit{black and white street photograph of [SCENE], high contrast, candid, silver gelatin film look}
\end{promptbox}

%%%%%%%%%%%%%%%%%%%%%%%%%%%%%%%%%%%%%%%%%%%%%%%%%%%%%%%%%%%%

\end{document}